\documentclass[format=manuscript, screen=true, nonacm=true, authorversion=true]{acmart}

\usepackage{booktabs} 
\usepackage{xcoffins}
\usepackage{colortbl}
\usepackage{bm}
\usepackage{subfig}
\usepackage{threeparttable}
\usepackage{multirow}
\usepackage{etoolbox}

\usepackage[ruled]{algorithm2e} 

\SetAlFnt{\small}
\SetAlCapFnt{\small}
\SetAlCapNameFnt{\small}
\SetAlCapHSkip{0pt}
\IncMargin{-\parindent}

\setcopyright{none}

\begin{document}
\title[Robots Learning to Say `No']{Robots Learning to Say `No': Prohibition and Rejective Mechanisms in Acquisition of Linguistic Negation}

\author{Frank F\"orster}
\orcid{0000-0003-1797-682X}
\affiliation{%
  \institution{University of Hertfordshire}
  \streetaddress{College Lane}
  \city{Hatfield}
  \postcode{AL10 9AB}
  \country{UK}}
\email{f.foerster@herts.ac.uk}
\author{Joe Saunders}
\affiliation{%
  \institution{University of Hertfordshire}
  \streetaddress{College Lane}
  \city{Hatfield}
  \postcode{AL10 9AB}
  \country{UK}
}
\email{j.1.saunders@herts.ac.uk}
\author{Hagen Lehmann}
\affiliation{%
  \institution{Universit\`a di Macerata}
  \streetaddress{Via Crescimbeni 30-32}
  \city{Macerata}
  \postcode{62100}
  \country{Italy}
}
\email{hagen.lehmann@gmail.com}
\author{Chrystopher L. Nehaniv}
\affiliation{%
  \institution{University of Hertfordshire}
  \streetaddress{College Lane}
  \city{Hatfield}
  \postcode{AL10 9AB}
  \country{UK}}
\email{C.L.Nehaniv@herts.ac.uk}

\begin{abstract}
  `No' belongs to the first ten words used by children and embodies the first active form of linguistic negation. Despite its early occurrence the details of its 
  acquisition process remain largely unknown. The circumstance that `no' cannot be construed as a label for perceptible objects or events puts it outside of the 
  scope of most modern accounts of language acquisition.
  Moreover, most symbol grounding architectures will struggle to ground the word due to its non-referential character.
  In an experimental study involving the child-like humanoid robot iCub that was designed to illuminate the acquisition process of negation 
  words the robot is deployed in several rounds of speech-wise unconstrained interaction with na\"{i}ve participants acting as its language teachers.
  The results corroborate the hypothesis that affect or volition plays a pivotal role in the socially distributed acquisition process. Negation words are 
  prosodically salient within prohibitive utterances and negative intent interpretations such that they can be easily isolated from the teacher's speech signal. 
  These words subsequently may be grounded in negative affective states. However, observations of the nature of prohibitive acts and the temporal relationships between 
  its linguistic and extra-linguistic components raise serious questions over the suitability of Hebbian-type algorithms for language grounding.
\end{abstract}

%
%


%
%


\maketitle


\section{Introduction}
In research on early language we often find the claim that children's early productive vocabularies were dominated by nouns referring to
concrete objects such as foods or toys. This assumption appears to have been picked up and reinforced by work in robotics on symbol grounding
(cf. \cite{Stramandinoli2017}). As a consequence there are plenty of studies that focus on the acquisition of precisely these
types of words \cite{Hollich2000}.\\
The importance of mutual or joint reference between mothers and children to perceptible objects and events is emphasized by more recent, so called
usage-based theories of language development \cite{Tomasello2003}. Mother, child, and external referent make up a triadic joint-attentional
frame which are very much the focus of these later theories. Cognitively such triadic interactional constellations are more complex than a simple
dyadic interaction.

In those areas of developmental robotics concerned with language acquisition in artificial agents the linguistic units in focus are similarly those
that can be construed as referents for concrete physical objects, object properties, or perceptible events.
Central in this area of research is the notion of symbol grounding, the construction of links between abstract symbols and signals or constructs that are
based on the embodiment of the agent \cite{Harnad1990}. The construction of such links may be regarded as a form of sense making with respect to the 
linguistic entities under consideration. The linguistic units in question are typically words or simple grammatical constructions. The agent's embodiment 
often presents itself with respect to a stream of sensorimotor data. 

In stark contrast to this focus on concrete referents, and the words that label them, is the observation that amongst the very first words produced
by a toddler are many which do not fall into this category. We find words such as `no', `hi', or 'bye' \cite{Fenson1994} which are sometimes referred
to as \emph{social words}. Negation words are thus amongst the very first words in many toddlers' active vocabularies and are used by them to
reject things or to self-prohibit \cite{Gopnik1988,Volterra1979}, the latter being a function that is rarely seen with adult speakers.
The idea to link these social words with a robot's sensorimotor data appears strangely inappropriate. The question then is how these types
of words should be handled in an embodied language acquisition framework.

\section{Symbol Grounding}
Artificial symbol grounding or perceptual symbol systems \cite{Barsalou1999} attempt to solve or break out of the symbol grounding problem, the formulation of
which is frequently attributed to Harnad \cite{Harnad1990}: If symbols are recursively defined or explained merely by the concatenation of other symbols, as is the case 
in a dictionary, how can the agent make sense of such symbols given the often circular relationship between the explanandum and explanans (see \cite{Roy2005} for 
an example)? The principle method employed in solving this problem is to connect some or all symbols of the system to sensorimotor data that originates in the agent's own 
embodiment: its visual sensors, its haptic sensors, its auditory perception, or any kind of derivative constructs that are computed from data originating from such sensory
channels. The way existing symbol grounding systems differ from each other is mainly in the method how the link between symbolic and sensor-derived data is established. 
The methods for constructing and maintaining such links may involve neural networks \cite{Sugita2005,Cangelosi2010}, symbolic artificial intelligence approaches 
\cite{Siskind2001,Dominey2005,Steels2003a}, statistical learning methods \cite{Stepanova2018}, or methods inspired by an enactive approach \cite{Nehaniv2013,Lyon2016}. 
The latter are typically data-driven and, arguably, representation-free in the sense that no models of object or event categories are constructed. Technically this 
can be made possible through the use of lazy learning algorithms \cite{Aha1997} which compute retrieval or classification requests directly on the `remembered' data.

Most of the existing work focuses and is limited to the grounding of words that may be seen to either refer to concrete objects (`concrete nouns') or to temporally unfolding 
perceptible events and processes (`concrete verbs'). More recently Stramandinoli et al. \cite{Stramandinoli2017} provided an example for grounding more abstract verbs
such as `use' or `make' on the back of already grounded concrete verbs such as `cut' or `slide'. While this is certainly an improvement in terms of the ability to ground a 
more general class of words, it is not clear how this approach would help to ground social or socio-pragmatic words such as `no', `yes', or `hi', or emotion words such
as `sad', or `upset'.

\section{Negation and Affect in Language Acquisition}
\label{sec:negation_affect}
Authors such as Pea emphasize the significance of affect in the context of acquisition of early linguistic negation \cite{Pea1980}.
Less well understood are the concrete ramifications of this primacy of affect for a cognitive architecture in terms of required
components, learning mechanisms, or the dynamics of the learning process. Hence roboticists have so far been unable to create machines
that could acquire and engage in this aspect of human speech.

Spitz \cite{Spitz1957} hypothesized that infants' major source of negation words is rooted in parental prohibitive utterances. Under this hypothesis, the 
infant's frustration, brought about by adult prohibition, leads to negative affect on the child's part who subsequently associates negative affect 
with the negative utterance. Via role reversal the child is then thought to use these negative symbols for the purpose of rejection. 

\subsection{Rejection Experiment}
In the \emph{rejection experiment} \cite{Foerster2017} we tested whether the robot's display of affect would lead to measurable changes in participants' speech
when addressing the robot. The kind of change we envisioned is best described with the notion of \emph{negative intent interpretations}: Intent interpretations
are linguistic descriptions or ascriptions of the addressee's affective or volitional state. They have been hypothesized to play a vital role for infants
to learn how to express their intent \cite{Ryan1974,Pea1980}. We use the results of the \emph{rejection experiment} within the present article as point of
reference in order to assess the comparative efficacy of the prohibition task in terms of the acquisition of negation words.\\[0.5ex]

Both the prohibition as well as the rejection experiments operationalize affect as central element of their respective hypotheses both of which describe different, yet mutually
non-exclusive mechanisms detailing the acquisition of early negation words.
The goal of both experiments is two-fold: firstly, we intend to improve our understanding of human language acquisition and, secondly, we intend to use
the so acquired knowledge to inform the design of future language grounding systems. We hope that this research will contribute to systems that transcend the
restriction of being limited to the acquisition of concrete referential words and phrases.
The acquisition system employed in the present study relies on strong interactional regularities that operate on relatively small time windows of several hundred
milliseconds to very few seconds. These statistical regularities in human conversational behavior are required for the embodied machine learning  
to work. One property of these regularities appears to be that they involve a certain asymmetry between the speakers: one party is conversationally leading
or stronger, typically the mother or the teacher-participant, while the other party is conversationally weaker or inept such as the infant or the child-humanoid.

\subsection{Overview of Prohibition Experiment}
\begin{figure}[ht]
\centerline{\includegraphics[width=\textwidth]{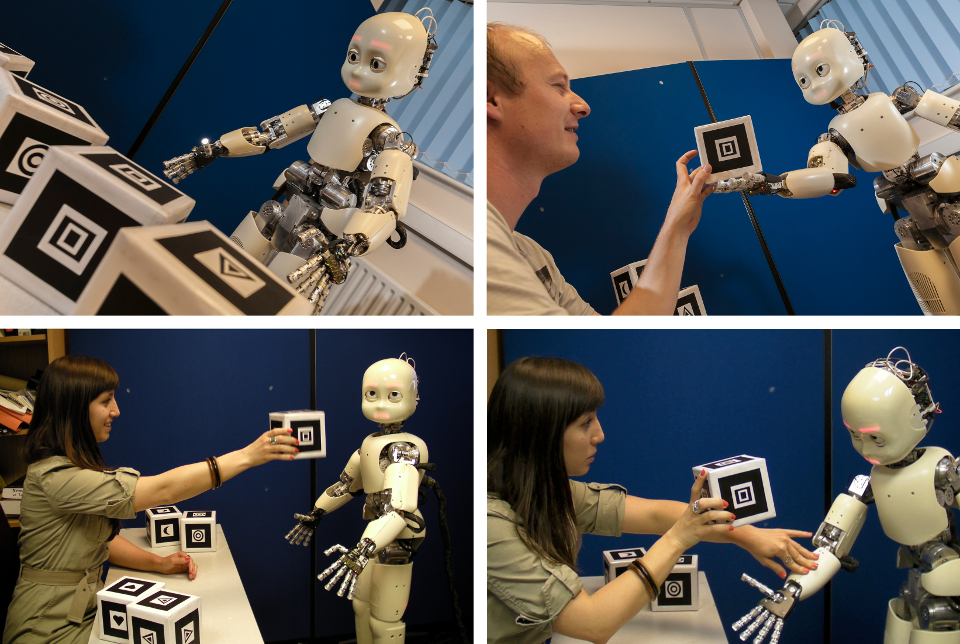}} 
\caption{Experimental setup and interactive robot behaviors as utilized in the experiment. Participants and robot face each other. The teaching objects
  are located between the two interactants on a table. Upper left: \emph{Looking around} behavior: no object is being presented to the robot.
  Upper right: \emph{Reaching} behavior: triggered by a participant's presentation of an object with positive valence. Lower left, \emph{Avoidance} behavior:
  triggered by a participant's presentation of an object with negative valence. Lower right: Modified \emph{reaching} behavior, executed in prohibition
  experiment, where participants can push back the robot's arm in order to prevent it from touching a forbidden object.}
\label{phys_setup}
\end{figure}

The participants employed were na\"{\i}ve with respect to the true purpose of the experiment, that is, they were unaware that linguistic negation was the
topic under investigation. They acted as language teachers for the iCub robot \cite{Metta2008}  and were asked to teach it the words for certain objects
on a table. The table was situated between them and the robot (see Fig. \ref{phys_setup}). In order to increase the likelihood of participants employing a
speech register akin to child-directed speech we asked them to imagine the robot as a pre-linguistic child. We also stated that the robot would like certain
objects and dislike others. We used identical instructions in both the rejection \cite{Foerster2017} as well as the present prohibition experiment both of which,
in turn, were very similar to \emph{Saunders' experiment}~\cite{Saunders2012}. Contrary to the two negation experiments, \emph{Saunders' experiment} was genuinely 
concerned with the acquisition of referentially concrete nouns, verbs, and adjectives. All three experiments utilized the same robot and took place in the same
room with the same physical setup in order to render the results comparable. \emph{Saunders' experiment} did not model affect. Its focus was sensorimotor association
of words with the robot's embodiment and we will use the results as point of reference later in this article.

The cognitive and behavioral architecture employed in all of these three experiments was largely identical with one important exception: Both the \emph{rejection}
as well as the \emph{prohibition experiment} utilized a motivation system which modulated the robot's affective expressions. These expressions were the facial expressions
smiling, frowning, and neutral, and matching body behaviors. Apart from modulating the robot's body behavior its affective state was also fed into the
symbol grounding system. As was the case in the \emph{rejection experiment}, the robot's affective states were triggered by randomly assigned valences
towards objects that were presented to it. The robot would attempt to grasp the object, in the case of a positive valence, or avoid it in the case of a negative valence.
In addition, and differing from the setup in the \emph{rejection experiment}, the robot's motivational state would turn
negative if it experienced external restriction of its arm movement.
In absence of the aforementioned motivational triggers, the robot would display a baseline behavior during which it would alternate its gaze between the
present objects and the participant's face.

\noindent The duration of the robot's gaze on a particular target were variable (cf. section \ref{sec:material}). If participants present an object with
a positive valence the robot smiles and holds out its hand towards them. It presents its palm which signals to participants that they can give it an
object if they choose to do so (\emph{reaching} behavior).

If participants present an object with a negative valence to the robot, it looks at the object briefly, starts to frown, and turns its head away.
The dynamics of the interaction can lead to several consecutive `turn away' movements which some participants interpreted as a form of head shake.
When presented with an object with neutral valence, the iCub displays a neutral facial expression. Additionally it will look in regular intervals at both
the participant's face and the presented object without approaching it (\emph{watching} behavior).
Prior to participants selecting the first object as well as between object presentations the robot displays a neutral face and switches its focus between all objects 
and the participant's face (\emph{looking around} behavior). The major driver of the robot's behaviors is thus its motivation system. The motivation system, in turn,
is modulated by the external object valences as well as physical restriction of its arm movement. The latter has an exclusively negative impact and there is no
positive counterpart.

Every instantiation of the experiment consists of five sessions and each session lasts about five minutes. The experiment was split into multiple sessions
because the word learning required some offline processing in between sessions. To this purpose both participants' speech as the recorded sensorimotor and motivation
(\emph{smm}) data were timestamped in order to allow for temporal alignment. The processing was semi-automatic in nature: the extraction of the prosodically most salient
word, one per utterance, was followed by the attaching of the \emph{relevant} \emph{smm} data to the word. This is the point where the temporal contiguity mentioned
in section \ref{sec:negation_affect} becomes crucial: Those parts of the \emph{smm} data are considered \emph{relevant} which were recorded
during the production of the respective utterance. From this point on, the salient word is grounded in the robot's embodied `experience'. After all of the participant's
utterances have been processed in this manner, the new set of so grounded words is added to the robot's embodied lexicon.
Note that a separate lexicon is created for each participant such that the robot follows an independent acquisition trajectory for every one of them.
The fact that it starts with an empty lexicon for each participant means that no designer knowledge in terms of a set of preselected words is incorporated.
Everything that the robot eventually says during the experiment originates from what the respective participant uttered in earlier sessions.

In every session but the first the lexicon is loaded into a memory-based learning system \cite{Daelemans2005}. Importantly, the robot is real-time deaf,
i.e. no real-time speech detection is in place. Rather than answering to questions, certain trigger  behaviors will make the robot query its embodied lexicon.
Trigger behaviors are behaviors which are caused by object presentations: \emph{grasping} for, \emph{rejecting}, or simply \emph{watching} a presented object.
When triggered the robot matches its \emph{smm} state against those associated to words in its current embodied lexicon and retrieves the best match. While a trigger
behaviour is active this process is continuously performed, which means at about 30 Hertz. As it is both impossible and impractical to speak at such a high rate, and
because we expect a certain level of noise within the \emph{smm} data, a thresholding mechanism is employed. This mechanism both stabilizes its linguistic output with
respect to noise as well as adjusts its speech frequency to a more plausible level.
The thresholding mechanism maintains a score for each best match word: the score of the respective word is increased whereas all other word scores are
decreased. Once the score of a particular word reaches a certain threshold, the word is sent to the speech synthesizer: the robot speaks. After it spoke,
all scores are reset to 0. The retrieval process now starts anew but on a reduced lexicon: the just-synthesized word has been removed.
The removed word is added back to the lexicon once another word has been synthesized but also if a change in the \emph{smm} state occurs (see also
section \ref{sec:material}).

\section{Materials and Methods}
\label{sec:material}

\subsection{Study Design}
 
\begin{figure}
  \begin{center}
    \includegraphics[width=.75\textwidth]{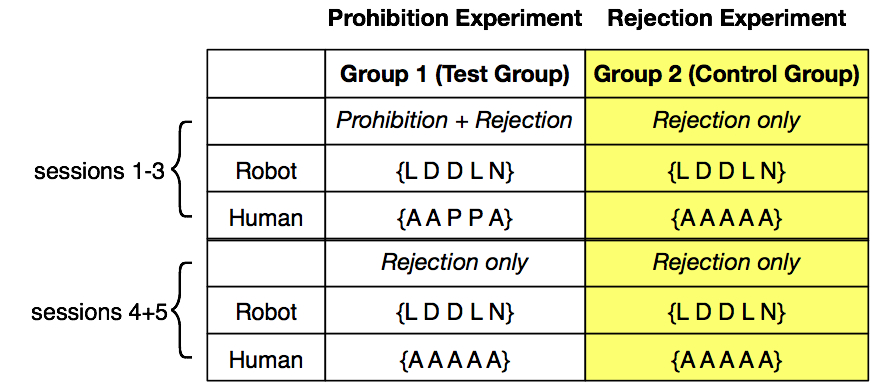}
  \end{center}
  \caption{\textbf{Alignment of Prohibition and Rejection Experiment}: \emph{Prohibition} and \emph{Rejection} in the table refer to the prohibition and 
    rejection scenarios respectively. Note that the \emph{prohibition experiment} is composed of both the \emph{prohibition} and \emph{rejection scenarios},
    whereas the \emph{rejection experiment} consists of the \emph{rejection scenario} only.
    Brackets (\textit{\{..\}}) indicate the permutations of the following values: the robot either `liked' (\textit{L}), `disliked' (\textit{D}), or `felt'
    neutral (\textit{N}) about an object. Only participants knew whether an object was allowed \textit{A} or prohibited/forbidden (\textit{P}) for the
    robot to touch. The mapping of \textit{positive}/\textit{negative}/\textit{neutral} (+/-/0) valences to objects was permuted between sessions, such that
    each object was twice \textit{liked}, twice \textit{disliked}, and once \textit{neutral} across the 5 sessions (see table \ref{tbl_mot_values}).
    The mappings were identical for every participant. The \textit{allowed}/\textit{forbidden} markers were permuted as well.}
  \label{fig_study_design}
\end{figure}

This \emph{prohibition experiment} presented in this article was designed in close alignment with the \emph{rejection experiment}, described in 
\cite{Foerster2017}, in an attempt to assess the linguistic impact of a caretaker's prohibition (Fig. \ref{fig_study_design}). 
Every instantiation of the prohibition experiment consisted of five sessions during which both the robot's behavior as well as the general experimental setup as developed for
 the \emph{rejection experiment} provide the baseline behavior and setup. However, during the first $3$ sessions, approximately half of the 
objects were additionally tagged with markers and  participants were instructed to prevent the robot from touching these - the \emph{prohibition task}. The last 
$2$ sessions of the prohibition experiment are identical to the \emph{rejection setup} such that the robot's success in acquiring negation words could
be compared between the two experiments.

\begin{table}
  \caption{Object valences per session for both prohibition and rejection experiment.}
  \label{tbl_mot_values}
  \begin{tabular}{@{\extracolsep{\fill}}lccccc}
    \hline
    & session 1 & session 2 & session 3 & session 4 & session 5\\
    \hline
    triangle & 1 & -1 & 1 & -1 & 0\\
    \hline
    moon & 0 & 1 & -1 & 1 & -1\\
    \hline
    square & -1 & 0 & 1 & -1 & 1\\
    \hline
    heart & 1 & -1 & 0 & 1 & -1\\
    \hline
    circle & -1 & 1 & -1 & 0 & 1\\
    \hline
  \end{tabular}
\end{table}
\subsubsection{Rejection Scenario} In order to increase the likelihood of the production of \emph{negative intent interpretations} we permuted the object-valence
mapping for each session (see table \ref{tbl_mot_values}). In this way it was impossible for participants to know which object the robot would dislike at the
outset of a particular session, which normally meant that they would present it at least once.

\subsubsection{Prohibition Scenario} This scenario was designed to test Spitz' hypothesis that the earliest forms of negation produced by toddlers would originate in
parental prohibitive utterances. The rejective scenario is serving as basis but is extended to include the elicitation of prohibitive utterances from the participants. 
Hence two or three of the five present objects were declared to be \textit{forbidden}, and were marked with colored dots on the side facing the participants. In
addition to the instructions from the rejection scenario participants were told that the marked objects were forbidden objects, and that the humanoid iCub robot
Deechee was not allowed to touch them. 
In order to keep Deechee from touching these forbidden boxes, participants were instructed to physically restrain the robot, in case it tried to touch them. 
Before the first session, participants were shown how to push the robot's arm back, firstly, in order to show them the ideal contact point, such that the robot's hand 
would not be damaged, and secondly, to take away their potential fear from actually touching the robot. The ideal contact point is the wrist and forearm. In this 
scenario, force control was used as the control mode for both arms, which makes it possible for participants to manipulate the arms while the robot executes a movement 
\cite{Pattacini2010}. The act of pushing the robot's arm is detected as physical resistance and registered by the perception system as \textit{resistance event}.
The occurrence of such a resistance event leads to the robot's motivation being set to negative: Deechee subsequently starts to frown. Furthermore the face-related 
gaze time is increased in order to give this emotional display a slightly higher intensity (see SI table \ref{tbl_time_constants}).

The assignment of the forbidden and allowed attributes was such that every combination of \textit{liked}/\textit{disliked} with \textit{allowed}/\textit{forbidden} 
would occur at least once within each session. This, together with the change of the valence-to-object mapping between subsequent sessions (table \ref{tbl_mot_values}) 
then led automatically to a permutation of the \textit{allowed} and \textit{disallowed} attribute-to-object mappings across sessions. In general there were either 
two or three \textit{forbidden} objects per session, and two or three \textit{allowed} ones.

In terms of the role reversal mentioned in the introduction none was needed as, in our architecture, there is neither a concept of self nor of other. For the prohibition 
experiment the behavioral trigger mechanism was modified such that negative motivation caused by participants' restriction of the robot's arm movement would not trigger 
its avoidance behavior. In these cases the robot would nevertheless frown in accordance with its motivational state.
Each participant completed five sessions of approximately five minutes each. The multi-session format facilitated linguistic development over time and was required
for the purpose of post-processing of participants' speech recordings. Participants' speech was recorded via headsets and each session was video-recorded.
Participants were not alone with the robot but either one or two more people were present. An operator was required in order to monitor the robot, and most often a
helper was present in order to put boxes back on the table that had been dropped by Deechee. In few sessions the helper was absent such that the operator
had to perform both tasks. As depicted in Fig. \ref{phys_setup} participant and robot were facing each other with a table separating the two. The teaching objects 
were 10cm long cardboard cubes which had black-and-white depictions of various shapes glued to each side of each box. The shapes were a square, a triangle, a star,
a heart, and a crescent moon and  all sides of a box showed identical shapes.
Participants first read the instructions and signed the consent form. Afterwards they took their seat opposite the iCub and were subsequently asked to count
down ``three, two, one, start''. So `start' acted as start marker for the session. After five minutes the operator would give participants a signal. 
The operator, upon hearing `start', would press a button. The button press led to a time stamp being broadcasted through the architecture which was recorded by
the robot's body memory.

\subsection{Recruiting and distribution of participants} We recruited 10 participants all of whom were native English speakers and which were gender-balanced. 
The majority of participants were students or university employees. They were remunerated with \pounds 20 once they completed all five sessions. The protocol
was approved by the ethics committee of the University of Hertfordshire under protocol number 0809/88, and approval was extended under protocol number 1112/42.

\subsection{Instructions to Participants} The majority of instructions in both negation experiments as well as \emph{Saunders' experiment} \cite{Saunders2012} were
identical: participants were told that they ought to teach the robot Deechee about the objects on the table. We tried to prime participants into adopting
a style of speech akin to child-directed speech (\textit{CDS}, \cite{Snow1977,Gallaway1994,Newport1977}) by telling them to imagine Deechee to be a two-year-old. 
In addition to what participants were told in \emph{Saunders' experiment} we also decided to mention to them that Deechee had likes and dislikes with
respect to the objects such that they would not be caught by surprise by Deechee's emotional displays. It is not clear whether the latter instruction was strictly
necessary and whether it had any impact in terms of the content and style of participants' speech. Only in the \emph{prohibition experiment} participants
were told that the marked objects on the table were forbidden for Deechee to touch. Participants were instructed to push the robot's arm away if it should
approach these objects and they were practically shown how to do this.

\subsection{Behavioral Architecture}\label{sec_bhv_arch} 
\begin{figure*}[ht]
  \begin{center}
    \includegraphics[width=\textwidth]{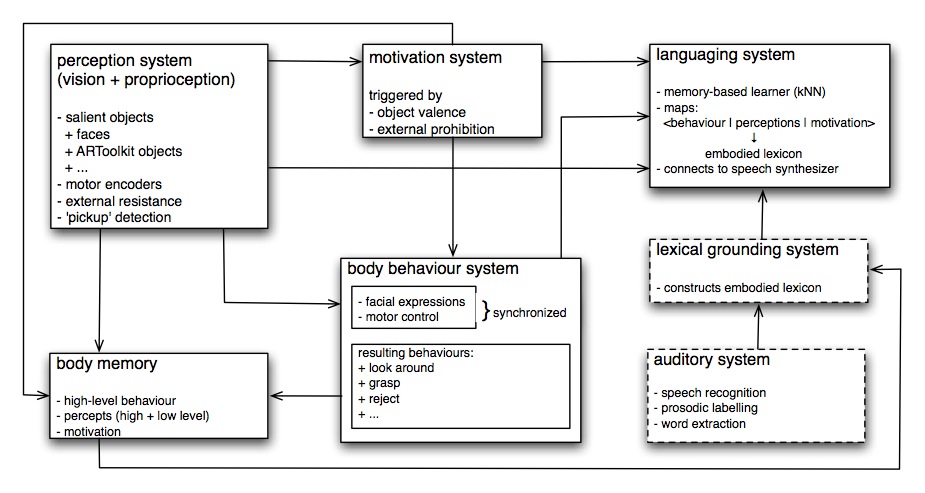}
    \caption{\textbf{Functional overview of robotic architecture for language acquisition}. Solid lines indicate components
      that are active during experimental sessions (``online''), dotted lines indicate components that work offline.}
    \label{fig_arch_overview}
  \end{center}
\end{figure*}

The behavioral architecture which generates both the humanoid's bodily and linguistic behaviour consists
of the components depicted in Fig. \ref{fig_arch_overview}.
We will sketch each component's purpose only very shortly as more elaborate descriptions have already been provided in \cite{Foerster2017} and \cite{Foerster2013}.

The \textbf{perception system} gathers and processes percepts of all modalities. Visual processing  was limited to face and object detection and based on
the system developed by R\"{u}sch et al. \cite{Ruesch2008}. We also developed a detector for object-related \emph{pick up} actions.

The \textbf{motivation system} is responsible for generating the affective-motivational state of the robot which consists of a simple scalar value between $-1$ and $1$.
$-1$ corresponds to a negative, $+1$ a positive and a small band around $0$ a neutral state (cf. \cite[Chapter 6]{Varela1991}).

The \textbf{body behavior system} generates the humanoid's robot's physical behavior which also includes its facial expressions. The behavior is generated contingent
upon the inputs from the perception and the motivation system. The behaviors are \emph{Rejecting}, \emph{Watching}, \emph{Looking around}, \emph{Reaching for object},
and \emph{Idle}. Other subsystems are informed of changes in the bodily behavior by the broadcasting of unique behavior ids.
Relevant time constants for certain parts of the behavior such as the eye gaze are listed in table \ref{tbl_time_constants} of the supporting information (SI).

The \textbf{body memory} saves high-level and low-level perceptual data as well as behavior ids and the robot's motivational state to a file.
\label{body_memory}

The \textbf{auditory system} encompasses speech recognition, word alignment, prosodic labeling, and word extraction, and are based on Saunders' system
\cite{Saunders2011}. Utterance boundaries are set based on statistics of inter-word pause durations and word durations (see \cite{Saunders2011} for details).
Important for the later analysis is the notion of \emph{prosodic saliency}. Note in this context that the first three aforementioned subsystems produce
a sequence of utterances, where each utterance consists of prosodically annotated words.
Exactly one word is extracted per utterance and that word is the prosodically most salient one. Prosodic salience is calculated
as $f_0 * energy * d_w$, where $f_0$ is the maximum fundamental frequency, $energy$ is the maximum energy, $d_w$ is the word duration, and all of these
components are normalized before said formula is applied.

The \textbf{lexical grounding system} performs the association of \emph{smm} data with the salient words originating from the auditory system
(see Fig. \ref{fig_word_grounding}). The so grounded words are subsequently added to the embodied lexicon.

\begin{figure}
  \begin{center}
    \centerline{\includegraphics[width=.75\textwidth]{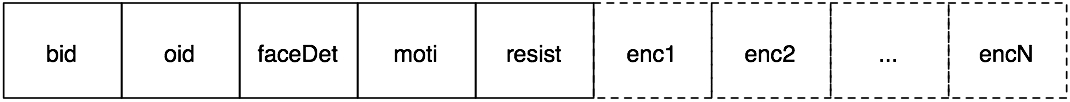}}
  \end{center}
  \caption{\textbf{Sensorimotor-motivational (\emph{smm}) vector}. Solid lines mark those data dimensions that were used for symbol grounding
    and matching; \emph{bid}: behavior id, \emph{oid}: object id, \emph{faceDet}: face detected, \emph{moti}: motivation value, \emph{resist}:
    resistance detected, \emph{encX}: encoder \#X}
  \label{fig_smm_vector}
\end{figure}

The \textbf{languaging system} generates the robot's speech. It does so based on a process that matches the robot's current \emph{smm} state (Fig.
\ref{fig_smm_vector}) against the \emph{smm} states associated to words in the embodied lexicon yielding a best-matching word in combination with a thresholding
mechanism. At the core of the matching process is the k-nearest neighbor implementation Tilburg Memory-Based Learner \cite{TiMBL2005}.
A new matching is performed whenever a new \emph{smm}-vector is available, which is the case approximately every 30ms.
The repetitive uttering of the same word is prevented by the use of a so called \emph{differential lexicon} which prevents the repeated production
of the most recently uttered word. For details of both the thresholding mechanism as the \emph{differential lexicon} see \cite{Foerster2013,Foerster2017}.

\section{Data Analysis} The following analysis is based on 5 hours of participants' speech originating from 50 sessions, 10 participants with 5 sessions
each, and was gathered as part of the prohibition experiment (cf. SI table \ref{tbl_ul_pro}). As was the case for the rejection experiment analyses were
performed on the word or corpus level, the utterance level, and the pragmatic level for negative words and utterances.
In the following the analysis will only be sketched as a more elaborate description has already been given in \cite{Foerster2017}.

The following variables were measured on the level of utterances: speech frequency in utterances per minute (\emph{u/min}), mean length of
utterance (\emph{MLU}), and the number of distinct words (cf. Fig. \ref{fig_production_rates} and SI tables \ref{tbl_ul_pro} to \ref{tbl_neg_utt_S})

Negative utterances are utterances that contain at least one negative word. Whether a word is a negative word was determined manually by examining the global
list of distinct words compiled from all participants' speech transcripts.

On the \emph{corpus-level} the prohibition corpus (\emph{PC}) is a list of all words and their frequencies that occur in participants' speech transcripts taken from the
prohibition experiment. The \emph{PCS} is a subset of the \emph{PC} containing only those words that were marked prosodically salient. Both corpora are presented
together with the corresponding corpora from the \emph{rejection} and \emph{Saunders' experiment}: \emph{RC}, \emph{SC}, \emph{RCS}, and \emph{SCS} (cf. Fig.
\ref{fig_production_rates}C).

\begin{figure}
  \begin{center}
    \centerline{\includegraphics[width=\textwidth]{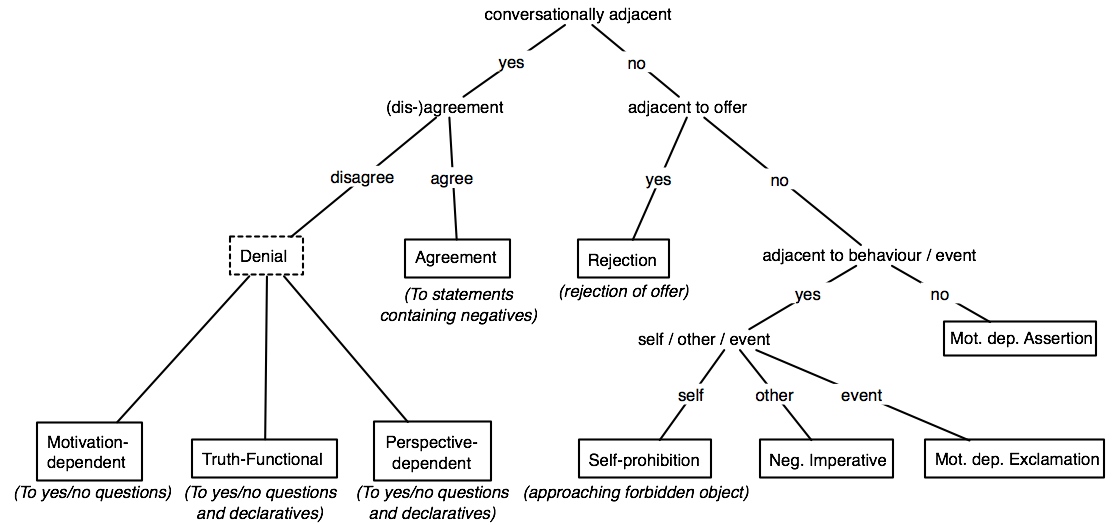}}
  \end{center}
  \caption{\textbf{Taxonomy of robot negation types}. Types of negative utterances produced by the robot and as identified by external coders.}
  \label{robot_neg_types}
\end{figure}

Only prosodically salient words enter the robot's embodied lexicon and therefore form the basis of its active vocabulary, hence our particular focus on them.
\NewCoffin \HTaxonomy
\NewCoffin \HTaxonomyPtOne
\NewCoffin \HTaxonomyPtTwo
\SetHorizontalCoffin \HTaxonomy{}
\begin{figure*}[h]
  \hspace*{-6ex}
  \SetHorizontalCoffin \HTaxonomyPtOne {
    \subfloat[Human Negation Types pt. 1]{
      \includegraphics[scale=0.75]{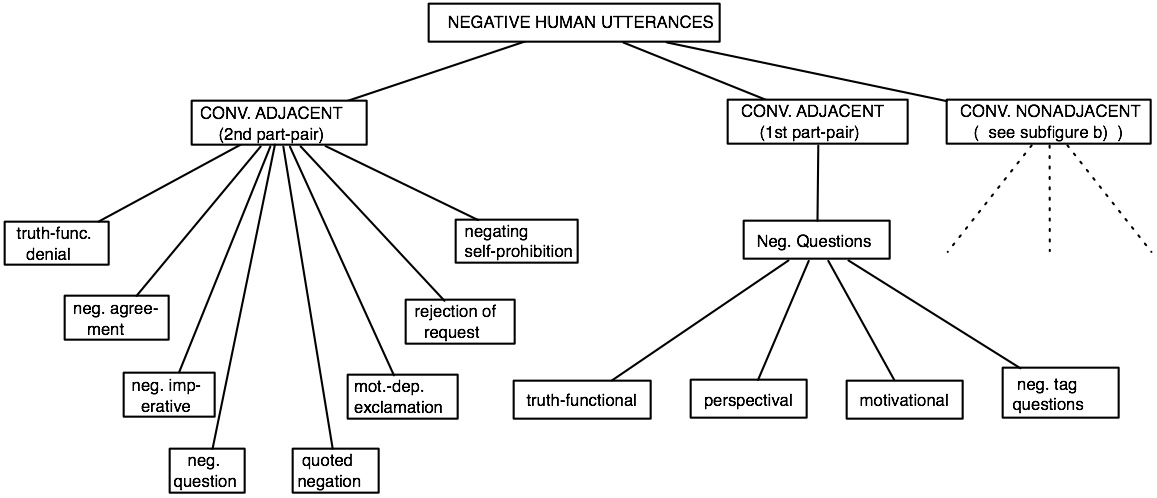}
    }
  }
  \JoinCoffins\HTaxonomy[vc,hc]\HTaxonomyPtOne[vc,hc]
  \SetHorizontalCoffin \HTaxonomyPtTwo {
    \vspace*{4ex}
    \subfloat[Human Negation Types pt. 2]{
      \includegraphics[scale=0.75]{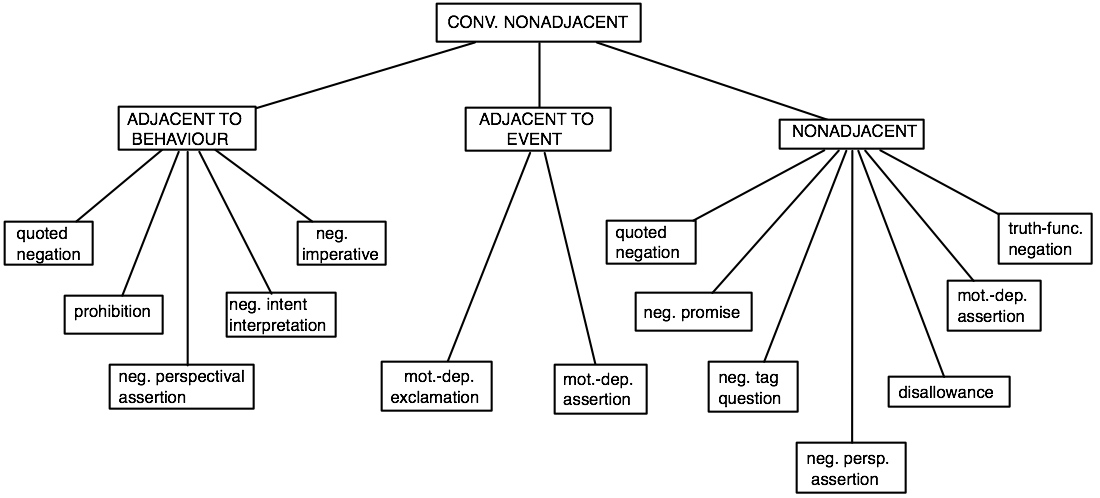}
    }
  }
  \JoinCoffins\HTaxonomy[\HTaxonomyPtOne-b,\HTaxonomyPtOne-l]\HTaxonomyPtTwo[t,l]
  \TypesetCoffin\HTaxonomy
  \caption{\textbf{Taxonomy of negation types used by participants}. Conv.: conversationally, 1st part-pair, 2nd part-pair: parts of an adjacency pair such as question 
    (1st part-pair) - answer (2nd part-pair)}
  \label{human_neg_types}
\end{figure*}
In order to be able to classify negative utterances of both participants and robot by their communicative function, their \emph{pragmatic type} of sorts, we
constructed two taxonomies, Figs. \ref{human_neg_types} and \ref{robot_neg_types}. The construction process is described in \cite{Foerster2017} and \cite{Foerster2013}
but we would like to emphasize that the resulting types can be regarded as types of speech acts in a loose sense which were enriched by the notion of
conversational adjacency. Conversational adjacency is not part of classical speech act theory (\cite{Austin1975,Searle1969}). A short sketch of the most important negation types is
given below but for a detailed description of all types we refer to the coding scheme \cite{Foerster2018}.
Upon completion of the two taxonomies two coders classified the negative utterances by type, where the first coder classified all utterances and the second coder
classified a randomly selected subset comprising $20\%$ of all utterances. This enabled us to assess the taxonomies for internal consistency
using Cohen's $\kappa$. Prior to coding the negative utterances for type, the coders coded the robots' utterances for felicity. This means, they had to make a
judgment whether they, by virtue of being fluent English speakers, perceive the negative robot utterance to be adequate or plausible in the respective
conversational context. The internal consistency of the human taxonomy in terms of Cohen's $\kappa$ was judged to be good ($\kappa=0.74$), but the consistency
of the robot taxonomy was judged to be only borderline acceptable, which triggered an automatic attempt to optimize it. Both the optimization attempt as well as our
reasons for not adopting the recommended mergers suggested by the optimization are described in \cite{Foerster2017} and \cite{Foerster2013}. Important for our current
purposes is the fact that the $\kappa$ values for the ratings of both the robot's felicity ($\kappa=0.46$) and type ($\kappa=0.41$) are at the very lower end
of what is generally regarded acceptable. This has to be kept in mind when interpreting numbers that are based on these ratings. Importantly, however, there was
no indication that any one of the two coders would have judged the the robot's negative speech systematically more favorably as compared to the other (see SI table
\ref{tbl:coder_felicity}).

Prior to describing the outcome of these attempts we need to introduce those negation types mentioned within the present article. These include the ones most frequently
produced during the experiments. A complete listing of all observed negation types can be found in the coding scheme \cite{Foerster2018}. 
In the following those types typically found in human participants' speech are qualified with `[H]'. The types typically found in the robot's speech are 
marked `[R]'. In the examples question marks indicate the intonational contour of a question, full stops the contour of an assertion.\\
\textbf{Negative Intent Interpretations (NII} [H]\textbf{)} are negative interpretations or ascriptions with regards to the addressee's motivational, emotional,
or volitional state \cite[p. 179]{Pea1980}.\\
Examples utterances falling into this category are ``No, you don't like fish'' or a simple ``No'' if it is not produced as a genuine question.\\
\textbf{Negative Motivational Questions (NMQ} [H]\textbf{)} are very similar to \emph{NII}s in that they refer to what is perceived to be the addressee's
negative motivational state. The main difference between \emph{NII}s and \emph{NMQ}s is the fact that the latter are considered genuine questions, meaning, the
speaker does expect the addressee to respond. Examples would be ``Are you not feeling well today?'' or ``You don't like apples?'' in the context of being
offered an apple, rather than the statement of a general preference.\\
\textbf{Truth-functional Denials (TFD} [H]\textbf{)} are used to deny a truth-functional assertion, with truth-functional assertions being assertions whose truth is
independent of either speaker's preferences or capabilities. Examples are ``No, it's not a hedgehog!'' in the presence of an unknown animal and counter the suggestion
of some other speaker or, again, a simple ``No.'' in reply to some positive assertion.\\
\textbf{Truth-functional Negations (TFN} [H]\textbf{)} in our taxonomy are a catch-all category for all of those kinds of truth-functional negation that
are not \emph{truth-functional denials} such as truth-functional suggestions or speculations, but also negative normative assertions such as
``In England you mustn't drive on the right-hand side.''\\
\textbf{Prohibitions (P} [H]\textbf{)} are negative utterances whose function is to prevent the addressee from doing something. Considered in isolation, such
utterances may not indicate that their function was prohibitive, as for example in the second example below. Taken out of context this utterance may be taken to
be a truth-functional negation. However, in context, when looking at a video recording of the actual interaction or when witnessing the latter `in vivo' it
becomes clear that this utterance is used as prohibition. In our experiment the prohibitive utterance can or cannot be accompanied by the participant physically
restraining the robot's arm movement.\\
Examples: ``No, you can't touch that'' or ``No you're not holding it, but you can look at them''.\\
\textbf{Disallowance (D} [H]\textbf{)} Disallowances are similar to \emph{prohibitions} but, in contrast, capture those utterances that express general negative
rules. In this sense disallowance utterances are more detached from the here and now of the interaction than \emph{prohibitive} utterances.  Whereas prohibitive
utterances are always triggered by a current action on part of the robot, \emph{disallowances} can or cannot accompany such an action.\\
Example: Speaker A takes something from the shelf and shows it to the robot saying ``You can't have this one''.\\
\textbf{Negative agreements (A} [H+R]\textbf{)} is a negative confirmation in response to a negative statement such as a ``no'', uttered by speaker $2$ in
response or addition to a ``So you don't like peanut butter, hmm?'' by speaker 1.\\
\textbf{Motivation-dependent Denials} [R] are negative answers to \emph{motivation-dependent questions} or \emph{assertions}. Their content is dependent
on the current emotional or volitional state of the addressee or her current preferences.\\
Example: ``No'' in response to ``Do you want some ice cream''.\\
\textbf{Rejections} [R] are very close to \emph{motivation-dependent denials}. The difference is that the latter is adjacent to an utterance whereas the former
is adjacent or in reaction to non-linguistic offers. For example, ``no'' in response to someone holding out an apple as a offer would fall into this category.\\
\textbf{Negative tag question (NTQ} [H]\textbf{)} are negative clauses that are attached to the end of the utterance. Semantically and pragmatically they
are probably the `least negative' of our negation types but they are easy to spot and appear to be highly frequent in British English. For example
``don't you'' as in ``You do like it, don't you?'' falls into this category. Another example would be ``weren't you'' in ``You have been to Cambridge, haven't you?''

In addition to the linguistic analyses a further analysis on the temporal relationships between participants' linguistic prohibition and their use of bodily measures
to restrain the robot was performed (cf. SI section \ref{sec:temporal_rels}).

\section{Results}
\begin{figure*}
  \SetHorizontalCoffin \ATCap {
    \begin{minipage}[t]{0.7cm}
      \hspace*{2ex}\textsf{\textbf{A}}
    \end{minipage}  
  }
  \JoinCoffins\FIGTHREE[vc,hc]\ATCap[vc,hc]
\SetHorizontalCoffin \AT {
\includegraphics[scale=0.51]{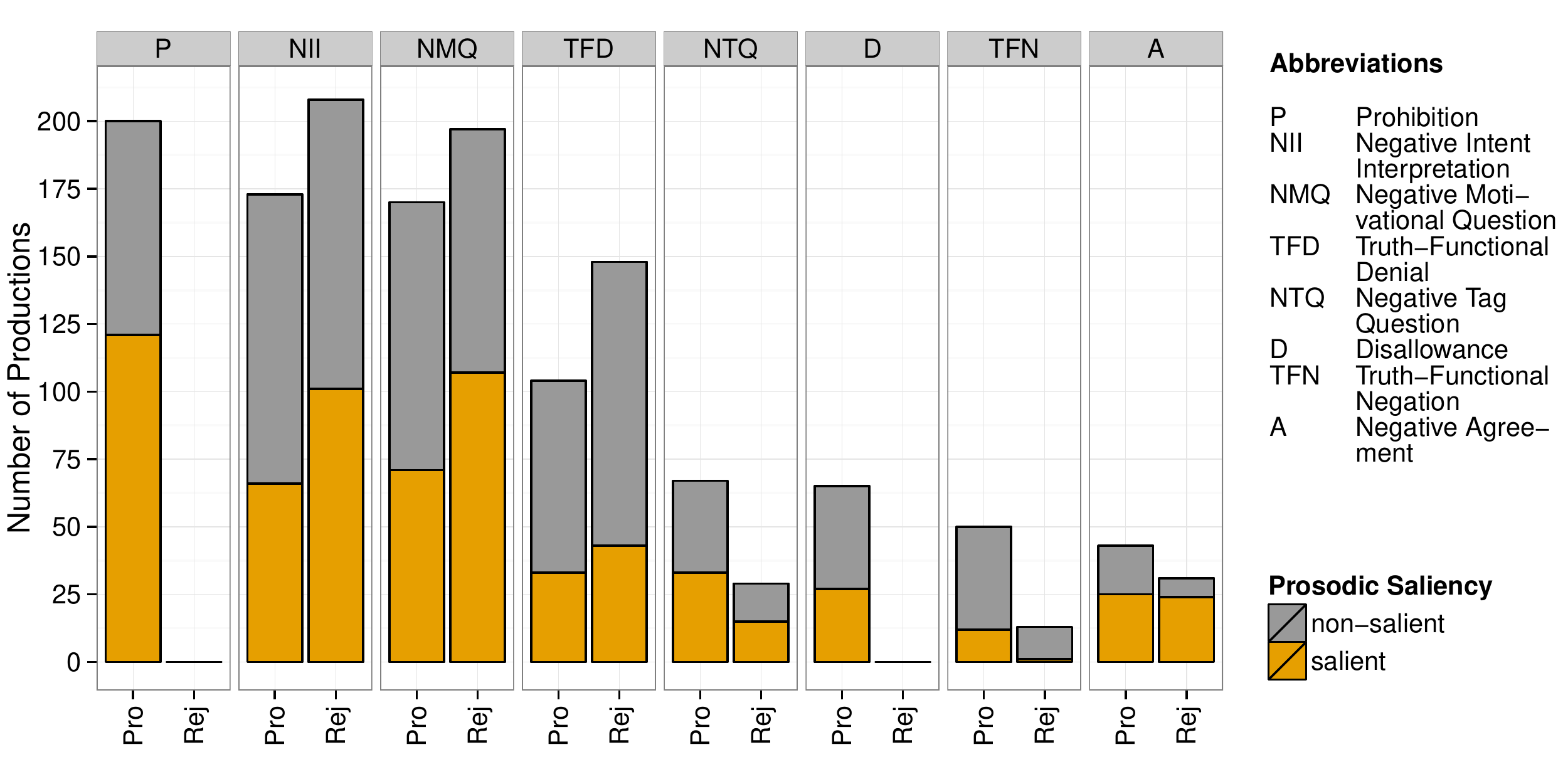}
}
\JoinCoffins\FIGTHREE[\ATCap-t, \ATCap-r]\AT[t,l]
\SetHorizontalCoffin \ABTCap {
  \begin{minipage}[t][0.7cm][c]{\textwidth}
  \end{minipage}
}
\JoinCoffins \FIGTHREE[\AT-b,\AT-l]\ABTCap[t,l]
\SetHorizontalCoffin \BTCap {
  \begin{minipage}[t]{0.7cm}
    \hspace*{2ex}\textsf{\textbf{B}}
  \end{minipage}  
}
\JoinCoffins\FIGTHREE[\ABTCap-b, \ABTCap-l]\BTCap[t,r]
\SetHorizontalCoffin \BT {
\includegraphics[scale=0.51]{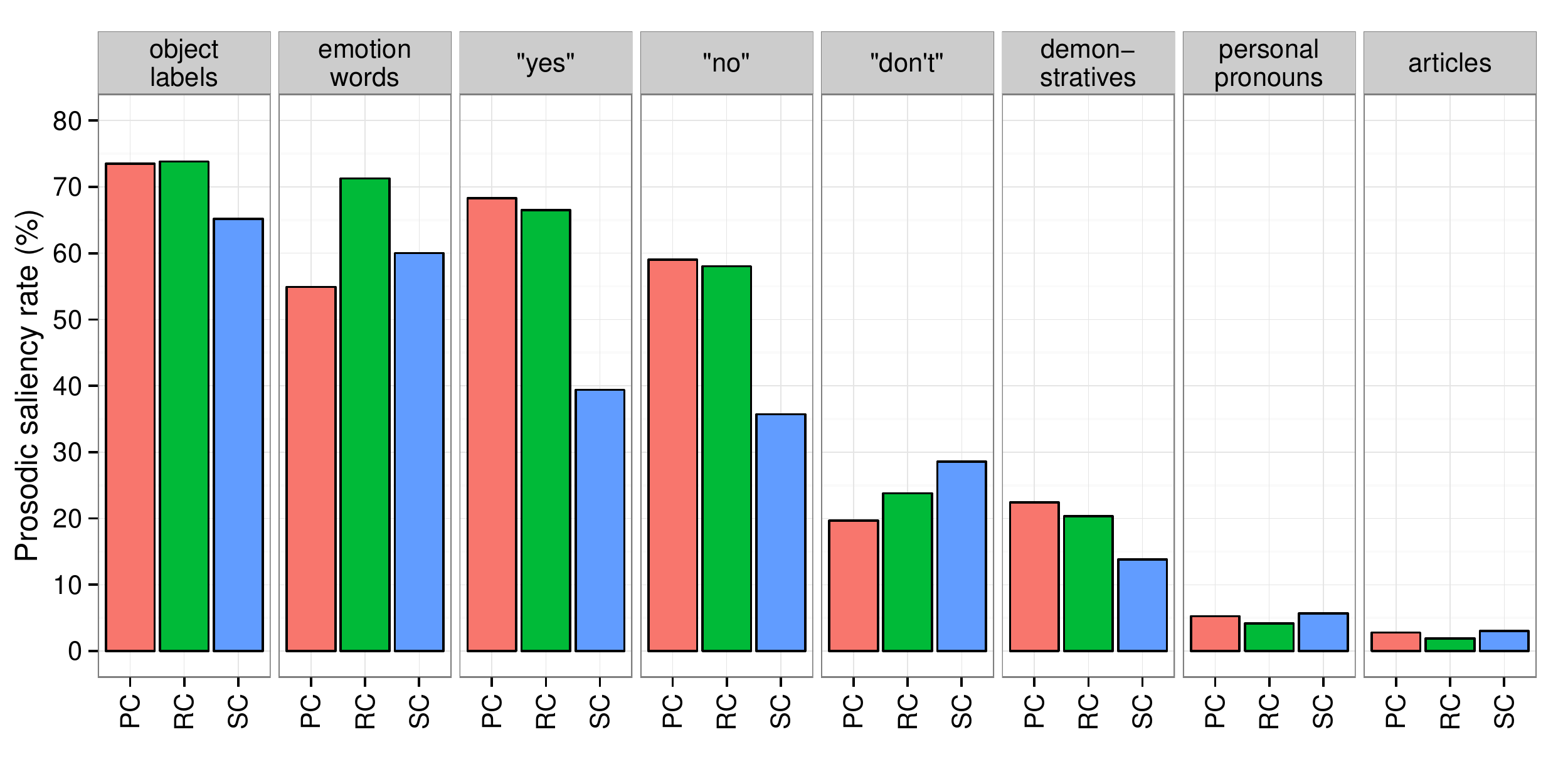}
}
\JoinCoffins \FIGTHREE[\BTCap-t,\BTCap-r]\BT[t,l]
\TypesetCoffin\FIGTHREE
\caption{(\emph{A}) Frequency of human utterances classified as being of the stated negation types (pragmatic level) and percentage of utterances falling under
  the respective type with salient negation word (only types with $> 5\%$ of total number of negative utterances, \emph{Pro}: Prohibition Experiment, \emph{Rej}:
  Rejection Experiment). (\emph{B}) Prosodic saliency rates of selected words and word groups. `No' has a considerably higher salience rate in the two negation
  experiments as compared to Saunders' experiment, \emph{PC}: Prohibition Corpus, \emph{RC}: Rejection Corpus, \emph{SC}: Saunders' et al. Corpus (see also
  SI tables S7 to S12).}
\label{fig_salience}
\end{figure*}

\NewCoffin \FIGTWO
\NewCoffin \ACap
\NewCoffin \A
\NewCoffin \BCap
\NewCoffin \B
\NewCoffin \Ce
\NewCoffin \CCap
\NewCoffin \ABCap
\SetHorizontalCoffin \FIGTWO{}
\begin{figure*}
  \SetHorizontalCoffin \ACap {
    \begin{minipage}[t]{0.5cm}
      \hspace*{2ex}\textsf{\textbf{A}}
    \end{minipage}  
  }
  \JoinCoffins\FIGTWO[vc,hc]\ACap[vc,hc]
  \SetHorizontalCoffin \A {
    \includegraphics[scale=0.297]{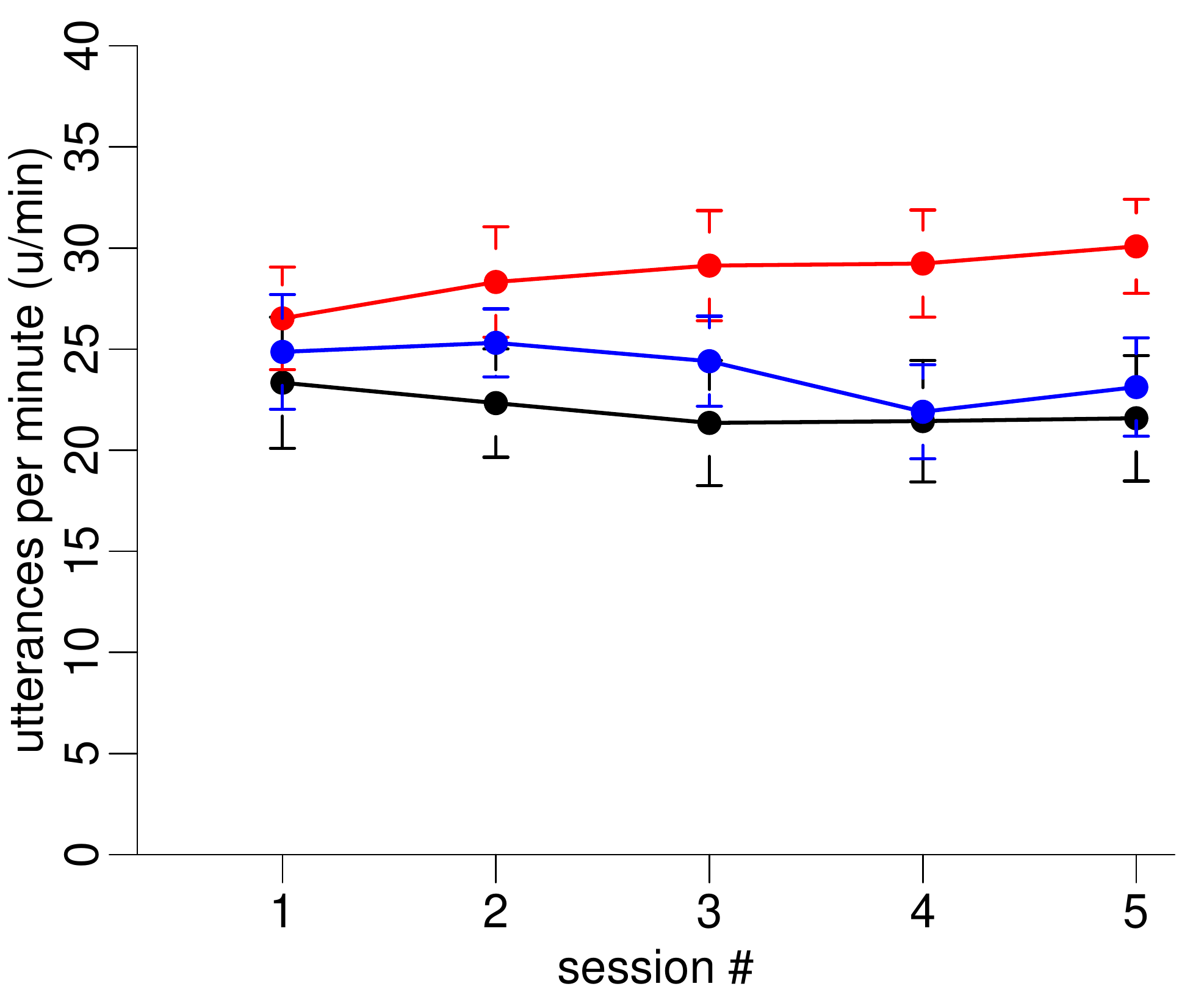}
  }
  \JoinCoffins\FIGTWO[\ACap-t, \ACap-r]\A[t,l]
  \SetHorizontalCoffin \BCap {
    \begin{minipage}[t]{0.5cm}
      \hspace*{2ex}\textsf{\textbf{B}}
    \end{minipage}  
  }
  \JoinCoffins\FIGTWO[\A-t, \A-r]\BCap[t,l]
  \SetHorizontalCoffin \B {
    \includegraphics[scale=0.297]{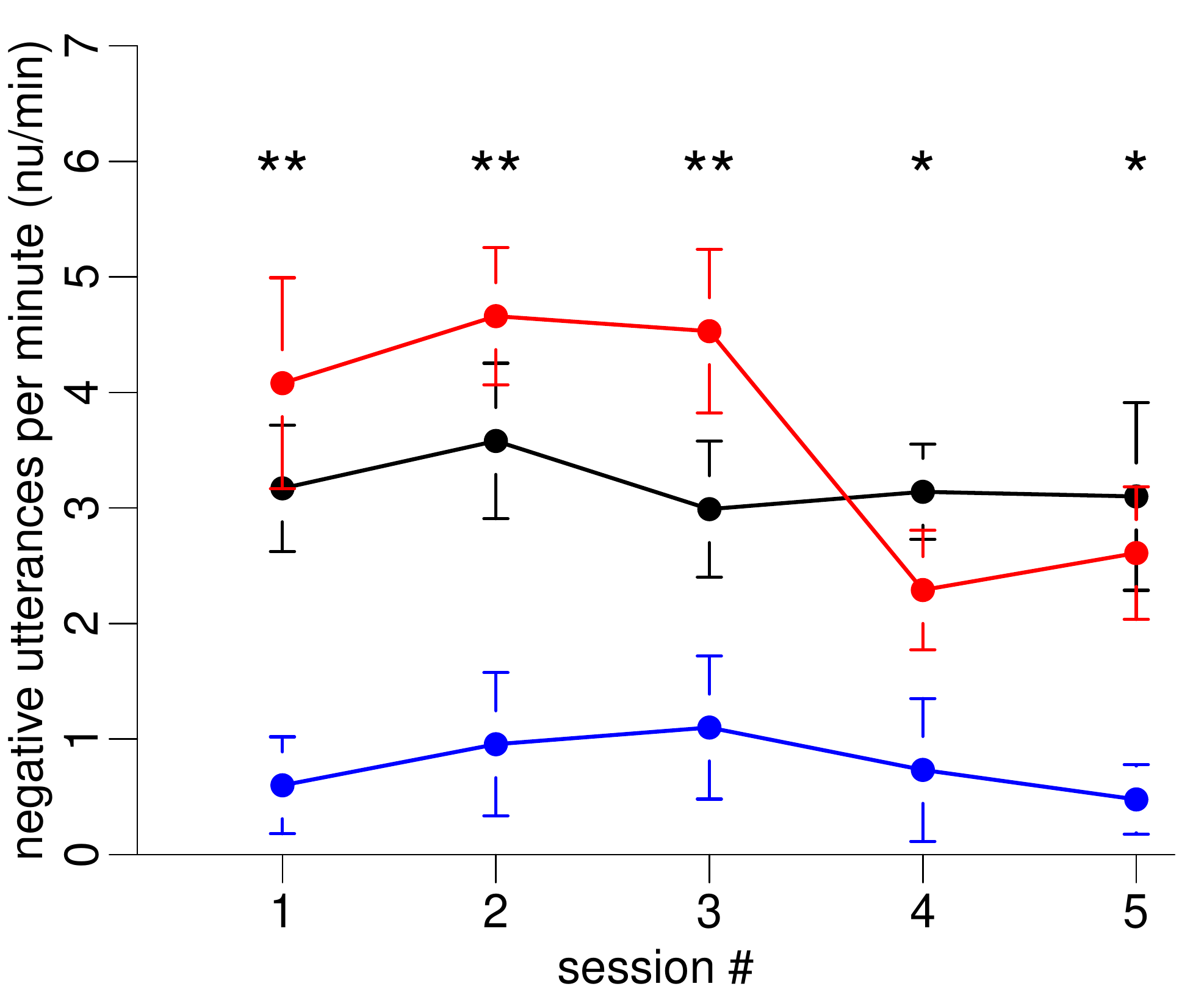}
  }
  \JoinCoffins \FIGTWO[\BCap-t,\BCap-r]\B[t,l]
  \SetHorizontalCoffin \ABCap {
    \begin{minipage}[t][0.5cm][c]{\textwidth}
    \end{minipage}
  }
  \JoinCoffins \FIGTWO[\A-b,\A-l]\ABCap[t,l]
  \SetHorizontalCoffin \CCap {
    \begin{minipage}[t]{0.5cm}
      \hspace*{2ex}\textsf{\textbf{C}}
    \end{minipage}  
  }
  \JoinCoffins \FIGTWO[\ABCap-b, \ABCap-l]\CCap[t,r]
  \SetHorizontalCoffin \Ce {
    {\small\sffamily
      \setlength{\tabcolsep}{0.4ex}
      \begin{tabular*}{0.9\textwidth}{@{\extracolsep{\fill}}lllllllllllll}
        \toprule
        & \multicolumn{6}{c}{All words} & \multicolumn{6}{c}{Salient words only}\\
        \cmidrule(lr){2-7}\cmidrule(lr){8-13}
        & \multicolumn{2}{c}{Rej. Corpus} & \multicolumn{2}{c}{Pro. Corpus} & \multicolumn{2}{c}{S. Corpus} & \multicolumn{2}{c}{Rej. Corpus} & \multicolumn{2}{c}{Pro. Corpus} & \multicolumn{2}{c}{S. Corpus} \\
        rank & \hspace{1ex}word & \% & \hspace{1ex}word & \% & \hspace{1ex}word & \% & \hspace{1ex}word & \% & \hspace{1ex}word & \% & \hspace{1ex}word & \%\\
        \cmidrule(lr){2-3}\cmidrule(lr){4-5}\cmidrule(lr){6-7}\cmidrule(lr){8-9}\cmidrule(lr){10-11}\cmidrule(lr){12-13}
        1 & \hspace{1ex}you & 7.13 & \hspace{1ex}you & 6.18 & \hspace{1ex}a & 8.71 & \hspace{1ex}square & 4.97 & \hspace{1ex}square & 4.4 & \hspace{1ex}blue & 6.91\\
        2 & \hspace{1ex}the & 5.63 & \hspace{1ex}the & 5.5 & \hspace{1ex}this & 4.55 & \cellcolor[gray]{0.85}\hspace{1ex}no & 4.64 & \hspace{1ex}triangle & 4.06 & \hspace{1ex}red & 5.54\\
        3 & \hspace{1ex}like & 3.31 & \hspace{1ex}a & 3.74 & \hspace{1ex}blue & 4.31 & \hspace{1ex}triangle & 3.95 & \hspace{1ex}circle & 3.83 & \hspace{1ex}circle & 5.15\\
        4 & \hspace{1ex}a & 2.72 & \hspace{1ex}this & 3.71 & \hspace{1ex}is & 4 & \hspace{1ex}heart & 3.8 & \cellcolor[gray]{0.85}\hspace{1ex}no & 3.66 & \hspace{1ex}heart & 4.75\\
        5 & \hspace{1ex}this & 2.7 & \hspace{1ex}one & 2.8 & \hspace{1ex}and & 3.9 & \hspace{1ex}moon & 3.53 & \hspace{1ex}one & 3.36 & \hspace{1ex}green & 4.36\\
        6 & \cellcolor[gray]{0.85}\hspace{1ex}no & 2.39 & \hspace{1ex}is & 2.45 & \hspace{1ex}red & 3.75 & \hspace{1ex}circle\hspace{1.5ex} $\bm{\uparrow}$ & 3.53 & \hspace{1ex}heart & 3.34 & \hspace{1ex}arrow & 3.56\\
        7 & \hspace{1ex}one & 2.27 & \hspace{1ex}like & 2.05 & \hspace{1ex}green & 3.55 & \hspace{1ex}like & 3.2 & \hspace{1ex}this & 3.09 & \hspace{1ex}cross & 3.48\\
        8 & \hspace{1ex}square & 1.93 & \hspace{1ex}to & 1.83 & \hspace{1ex}the & 3.29 & \hspace{1ex}circles & 2.69 & \hspace{1ex}moon & 2.8 & \hspace{1ex}side\hspace{1.6ex}$\bm{\uparrow}$ & 3.48\\
        9 & \hspace{1ex}do\hspace{3ex}$\bm{\uparrow}$ & 1.93 & \cellcolor[gray]{0.85}\hspace{1ex}no & 1.79 & \hspace{1ex}that's & 2.54 & \hspace{1ex}squares & 2.42 & \hspace{1ex}ok & 2.3 & \hspace{1ex}box & 2.82\\
        10 & \hspace{1ex}to & 1.78 & \hspace{1ex}it's & 1.66 & \hspace{1ex}you & 2.41 & \hspace{1ex}it & 2.36 & \hspace{1ex}shape & 2.14 & \hspace{1ex}shape & 2.42\\
        11 & \hspace{1ex}it\hspace{4.2ex}$\bm{\uparrow}$ & 1.78 & \hspace{1ex}heart & 1.6 & \hspace{1ex}it's & 2 & \hspace{1ex}yes & 2.28 & \hspace{1ex}yes & 2.09 & \hspace{1ex}and & 2.11\\
        12 & \hspace{1ex}that & 1.73 & \hspace{1ex}square & 1.51 & \hspace{1ex}heart & 1.99 & \hspace{1ex}one & 2.13 & \hspace{1ex}crescent & 2 & \hspace{1ex}moon\hspace{0.5ex}$\bm{\uparrow}$ & 2.11\\
        13 & \hspace{1ex}moon & 1.62 & \hspace{1ex}triangle & 1.47 & \hspace{1ex}circle & 1.85 & \hspace{1ex}right & 1.88 & \hspace{1ex}like & 1.8 & \hspace{1ex}square & 2.07\\
        14 & \hspace{1ex}heart & 1.6 & \hspace{1ex}that & 1.42 & \hspace{1ex}arrow & 1.84 & \hspace{1ex}this & 1.82 & \hspace{1ex}circles & 1.74 & \hspace{1ex}this & 2.02\\
        15 & \hspace{1ex}is & 1.47 & \hspace{1ex}it\hspace{4.5ex}$\bm{\uparrow}$ & 1.42 & \hspace{1ex}side & 1.81 & \hspace{1ex}ok & 1.73 & \hspace{1ex}again & 1.51 & \hspace{1ex}star & 1.85\\
        16 & \hspace{1ex}triangle & 1.45 & \hspace{1ex}do & 1.4 & \hspace{1ex}cross & 1.49 & \hspace{1ex}again & 1.69 & \hspace{1ex}good & 1.44 & \hspace{1ex}ok & 1.76\\
        17 & \hspace{1ex}circle & 1.32 & \hspace{1ex}moon & 1.38 & \hspace{1ex}here & 1.39 & \hspace{1ex}ok & 1.57 & \hspace{1ex}it & 1.36 & \hspace{1ex}is \hspace{3.5ex}$\bm{\uparrow}$ & 1.76\\
        18 & \hspace{1ex}it's & 1.22 & \hspace{1ex}circle & 1.29 & \hspace{1ex}we & 1.38 & \hspace{1ex}good & 1.55 & \hspace{1ex}very & 1.25 & \hspace{1ex}small & 1.54\\
        19 & \cellcolor[gray] {0.85}\hspace{1ex}don't & 1.15 & \hspace{1ex}that's & 1.28 & \hspace{1ex}on & 1.37 & \hspace{1ex}Deechee & 1.46 & \hspace{1ex}ok & 1.18 & \hspace{1ex}right\hspace{1.7ex}$\bm{\uparrow}$ & 1.54\\
        +1 & \hspace{1ex}see & 1.12 & \hspace{1ex}shape & 1.2 & \cellcolor[gray]{0.85}\hspace{1ex}no (\emph{50}) & 0.35 & \hspace{1ex}Deechee$^2$ & 1.3 & \hspace{1ex}Deechee & 1.12 
                                     & \cellcolor[gray]{0.85}\hspace{1ex}no (\emph{32})& 0.44\\ 
        \bottomrule
      \end{tabular*}
    }
  }
  \JoinCoffins\FIGTWO[\ABCap-b,\ABCap-l]\Ce[t,l]
  \TypesetCoffin\FIGTWO
  \caption{Impact of motivated behavior on linguistic production of participants. (\emph{A}) The overall production rates between prohibition (upper
    red), rejection (lower black), and Saunders' experiment (middle blue) differ only marginally (mean $\pm$ SEM). (\emph{B}) The production rate
    of negative utterances only, however, is significantly higher in the prohibition (upper red) and rejection (middle black) as compared to Saunders' experiment. See
    \textit{Supporting Information} for details. *P $<$ 0.05, **P $<$ 0.01 (\emph{C}) The large supply of negative utterances has consequences on the
    corpus level: \emph{No} is amongst the 10 most frequent words in the rejection and the prohibition corpus, whereas it is located on rank 50 in Saunders' corpus.
    In the corpora of prosodically salient words \emph{no} ranks even higher. This is due to the high saliency of the word and it subsequently enters the robot's vocabulary
    frequently. (Arrows ($\bm{\uparrow}$) indicate equality of ranks between the stated entry and the next entry above. Negation words are marked through gray
    background. The `+1' row contains the 20th most-frequent words unless a different rank is specified in brackets.)}
  \label{fig_production_rates}
\end{figure*}
In the prohibition experiment, on average, every 7th to 8th utterance of participants contains a negation word which constitutes an increase of $391\%$ compared to
the speech recorded in Saunders' scenario (cf. Fig. \ref{fig_production_rates}). This compares to a rise of $332\%$ in the rejection experiment. The frequent occurrence
of negative utterances leads to a large increase of prosodically salient negation words which subsequently enter the robot's active vocabulary. The increase is amplified
by the relatively high prosodic saliency rate of `no' in both negation experiments. The prosodic saliency of negation words, chiefly `no', is high both in relation
to Saunders' experiment as well as in relation to the average word saliency within the negation experiments (Fig. \ref{fig_salience}A).
As a consequence `no' rises to the fourth rank in the prohibition corpus of salient words \emph{PCS},  and even to second rank in the corresponding \emph{RCS}
(Fig. \ref{fig_production_rates}).

Analysing these negative utterances with respect to their communicative function reveals that, within the prohibition experiment, linguistic \emph{prohibitions},
not present within the rejection experiment, occupy the top-rank, making up $21\%$ of all negative utterances. This is remarkable as linguistic \emph{prohibitions}
were only produced when the prohibition task was given, i.e. during the first three sessions, whereas utterances of the other types were produced in all of the five
sessions. \emph{Prohibitions} are followed by \emph{negative intent interpretations} ($18\%$), \emph{negative motivational questions} ($18\%$), and
\emph{truth-functional denials} ($11\%$) (cf. Fig. \ref{fig_salience} and SI table \ref{tbl_freq_neg_types_pro}), with prosodic saliency rates of $60.5\%$, $38.2\%$, 
$41.8\%$, and $31.7\%$ respectively. This means the three motivation-dependent types provide the majority of negation words for the robot's lexicon due to two factors:
Firstly, this type of utterances are dominant in terms of the absolute numbers of productions, and, secondly, the negation words that are part of these utterances
have higher rates of prosodic saliency than the truth-functional types.

In comparison, within the rejection experiment, the most frequent negation type, with $31\%$, are \emph{negative intent interpretations} (\emph{NII}), followed by 
\emph{negative motivational questions} (\emph{NMQ}, $30\%$). Both of these types have a direct link to the robot's display of affect (see also SI table
\ref{tbl_freq_neg_types_re}). \emph{Truth-functional denials} (\emph{TFD}) rank third ($22\%$) in the \emph{RC} but have a lower
saliency rate ($29\%$), relative to the two motivational types (\emph{NII}: $54\%$, \emph{NMQ}: $49\%$). From this we can conclude that in the \emph{rejection experiment} 
the vast majority of negation words in the robot's active vocabulary originate from utterances of the two motivation-dependent types, \emph{NII}s and \emph{NMQ}s.

When comparing the two negation experiments, it becomes clear that within the prohibition experiment \emph{negative intent interpretations} and
\emph{negative motivational questions} were produced less frequently than was the case in the rejection experiment. The probable cause for this is the fact that,
in both experiments, participants had overall the same amount of time, yet participants in the prohibition experiment spent part of their time with attempts to prohibit
the robot leaving them less time to engage in \emph{NII}s and \emph{NMQ}s.

When considering the saliency rates of negation words within utterances of the aforementioned types produced within the prohibition experiment it becomes
clear that linguistic \emph{prohibitions} constitute a formidable source of negation words: within this type negation words reach the overall highest saliency rate
($61\%$), followed by \emph{negative motivational questions} ($41\%$), \emph{negative intent interpretations} ($38\%$), and
\emph{truth-functional denials} ($32\%$) (see SI table \ref{tbl_freq_sal_neg_types_pro} for the complete listing).
The combination of high production rate and high saliency rate renders them the top contributors of negation words to the robot's active vocabulary within
this experiment. Every single prohibitive utterance contained at least one negation word whereas intent interpretations and motivational questions  were sometimes
performed in a non-negative way. Some participants for example used non-negative emotion words in response to the robot's negative affective display such as `sad'
as in ``why are you sad?'' where others more commonly used the negative ``you don't like it?''. Thus, from a merely lexical perspective, \emph{prohibitions} appear to be
more reliable sources of negation words than \emph{intent interpretations} and \emph{motivational questions}.

In \emph{Saunders' experiment} in comparison `no'  is ranked 50th in the \emph{SC} and 32nd in the \emph{SCS}. There it accounts for less than $0.5\%$ of words in both
corpora. Hence both the affective or motivational displays of the robot as well as the prohibition task lead to a considerably higher rate of negative utterances when
compared to the setup used by Saunders et al. which used a near-identical setup, but without affective displays and without prohibition task.

To our surprise the robot's learning success with respect to negation was judged to be considerably lower in the prohibition ($30\%$) as compared to the
rejection experiment ($65-70\%$) (see SI tables \ref{tbl:accu_felicity} and \ref{tbl:felicity_robot_stat}). 
This result triggered an additional analysis where we aligned the signal of the robot's arm sensor for external pressure with both its affective state as
well as the timing of negative utterances of the four most frequent negation types. This analysis showed that participants from the prohibition experiment
$46\%$ of the time did not physically restrain the robot's arm when uttering \emph{prohibitions} as instructed, and in $12\%$ of all cases uttered
\emph{prohibitions} before applying restraint (see SI table \ref{tbl:prohibition_rel}). In both cases grounded words enter the robot's lexicon where the negation
word is likely to be associated with positive affect (see SI table \ref{tbl:prohibition_mot}). This may then lead to the inappropriate usage of the word. In
only $18\%$ of cases did our participants utter \emph{prohibitions} while restraining the robots arm, that is, while it was in a negative affective state.
When performing a similar analysis for \emph{negative intent interpretations} we observed that in approximately two thirds of cases (rejection experiment: $66\%$,
prohibition experiment: $63\%$) the robot is in a negative motivational state as opposed to a positive one (rejection experiment: $9\%$, prohibition experiment: $6\%$)
such that there is a high likelihood of it associating the negative word with negative affect. For \emph{negative motivational questions} the
results look similar within the prohibition experiment ($58\%$ performed while in a negative state, and only $9\%$ while in a positive state), while in the
rejection experiment the number of performances while in a negative state ($40\%$) is nearly identical to the number of performances while in a neutral state ($41\%$).
Performances of this type while the robot is in a positive state are also not very frequent within this experiment ($19\%$).
Thus, albeit lexically not being equally reliable sources of negation words as \emph{prohibitions}, \emph{negative intent interpretations} appear to be better sources
for the establishment of an association of the negative word with negative affect if word learning is mainly modeled as a process of establishing associations between
sensorimotor-affective `concepts' and linguistic items. 

\NewCoffin \FIGTHREE
\NewCoffin \ATCap
\NewCoffin \AT
\NewCoffin \BTCap
\NewCoffin \BT
\NewCoffin \ABTCap
\SetHorizontalCoffin \FIGTHREE{}

\section{Discussion}

No quantitative psycholinguistic data exists that would give us any idea about the felicity rates of infants' when it comes to the use of negation words.
We can therefore make no comparative judgment with respect to the robot's felicity rates. Young children often do make semantic mistakes during certain developmental
stages such as over-generalizing nouns \cite{Gelman1998} and grammatical constructions \cite{Bowerman1988,Brooks1999}. Unfortunately there is no such data for `pragmatic
accuracy' in terms of the (non-)successful use of negation words such as `no'.\\

\noindent From the rejection experiment we have a good indication that \emph{negative intent interpretations} on their own could theoretically be sufficient in order to
associate negation words and negative affect which may be sufficient to bootstrap linguistic negation.
 
For \emph{prohibitions} the picture is somewhat more complicated. Our participants, despite having been instructed on how to physically prevent the robot from
touching an object, often chose not to do so. Instead we observed that participants frequently held forbidden objects out of the robot's reach instead of limiting its 
arm movement. As the robot does not perform any type of goal-evaluation with respect to having reached or touched the `desired' object it does not become frustrated 
in these cases. None of the existing studies on children's
acquisition of negation provides us with detail of the interaction akin to the temporal alignment between physical restraint and linguistic prohibition in the present
study (cf. SI section \ref{sec:temporal_rels}). As a consequence we do not know if behaviour similar to the one of our participants is typical in the interaction of 
parent-child dyads, or whether we witnessed a somewhat artificial behaviour as participants may have simply been reluctant to touch the robot.

Even given the lack of quantitative data from psycholinguistics, intuitively a $30\%$ success rate appears to be very low. 
In our architecture we modeled world learning to be by and large associative or Hebbian. This decision was not made on the basis on principle, but rather by 
the application of Occam's razor: given the lack of evidence to the contrary we chose one of the simplest types of learning algorithms. Yu et al. \cite{Yu2007} provide a more 
elaborate discussion on associative learning in word acquisition. 
If we assume the core word learning mechanism to be roughly associative, and if we further assume the behavior of our participants to be sufficiently similar to that of 
caretakers' behavior when prohibiting a child, we can draw tentative conclusions for the acquisition process. Assuming that Spitz' hypothesis is correct, the child must 
already be frustrated at the time when the prohibitive \emph{no} is being uttered - at least in the majority of cases. In our experiment this was not the case due to the 
limited ways in which participants could unknowingly frustrate the robot: only the application of physical restraint to its arm would have this effect. But our participants 
appeared to use physical restraint reluctantly. Thus, if all of the above assumptions hold, there must be sources of frustration other than physical restraint - holding
an object out of a child's reach will probably quickly lead to frustration on part of the child. In this respect our robot's motivational system is most probably too limited.

However, the low pragmatic success rate also led us to search the video platform YouTube for amateur videos depicting parents that engage in prohibition. Albeit of a 
somewhat anecdotal character some videos show situations where parents clearly prohibit their children by the mere use of speech and cause the child's 
frustration as a consequence of these `touch-less' acts of prohibition (``single whammy prohibition'')~\cite{Spivey2007}. In other videos however, prohibiting utterances are
swiftly followed by a combination of corporal restraint and linguistic prohibition akin to the behaviour we expected our participants to engage in (``double whammy
prohibition'')~\cite{DeWeerd2011}. Due to the anecdotal character of this evidence it is impossible to tell which of these two variants of prohibition is more typical, whether
one is the developmental precursor of the other, or whether it is a matter of the severity of the violation rather than a matter of the developmental stage. 

It is not hard to imagine that a child, having been exposed to several instances of ``double whammy'' prohibition, would learn that `resistance is futile' and that, as a
consequence of this learning process, ``single-whammy'' prohibition suffices to stop the child from engaging in the prohibited behaviour (\emph{variant A}). 

On the other hand it seems equally plausible that prohibitive utterances may carry distinctive prosodic features that let the child infer the caretaker's negative emotional or 
volitional stance. We may assume that a child in the relevant age range can perform simple inferences within a lay theory of emotions or volition \cite{Ong2015, Ong2018}. If
we then further assume that the child experiences the caretaker's relative interactional and interventional power on a daily basis, the child should be able to infer that
resistance is futile without a need for prior exposure to ``double whammy'' prohibition (\emph{variant B}). The lack of more than anecdotal evidence prevents us from excluding
one or both of these possibilities. 

It may be of interest to observe that both \emph{variants A} and \emph{B} require a more powerful class of learning algorithms than the simple associationistic one employed
by us. Both of these variants assume that the agent learns about the efficacy of its actions with respect to some goal. In machine learning this would be typically modeled
with some type of reinforcement learning which is arguably a more powerful class of learning algorithms than simple Hebbian-type associative learning.
The replacement or supplement of our memory-based learner with some type of reinforcement learning as core learning mechanism and the explicit modeling of goals would 
therefore most certainly increase the felicity rate when using negation words. In this case the robot's frustration could be triggered whenever it can't reach an object within 
a certain time frame and when this inability is caused by another agent. This would then lead to a rise in the number of grounded negation words with negative motivation value
in the data set of the memory-based learner. Yet his extension would also weaken one of our assumptions, namely that the core learning mechanism was one of mere associationism.

Under the assumption of prohibitive utterances being the main source of children's early negation words, the only potential rescue for a purely associationistic account we can 
conceive of hinges on the notion of emotional contagion \cite{Hatfield1993} (\emph{variant C}). It shares with \emph{variant A} the idea that acoustic or prosodic qualities of 
prohibitive utterances, potentially in conjunction with corresponding facial expressions, may carry an 
emotional charge. As we have seen from our data, prohibitions are typically prosodically salient. Assuming that their acoustic properties may have the potential to negatively 
impact the affective state of the recipient, corporal restraint might not be necessary to ``turn the infant's mood sour''. Corporal restraint may indeed only be used by the care-giver 
as the very last resort. If such a mechanism of acoustic affective contagion could be implemented within our learning architecture we would arrive at a point where the non-codified 
aspects of utterances would contribute to the modulation of an interlocutor's affective state, which in turn would form part of the basis for grounding the codified units of the same, 
or adjacent utterances. 

The difference between this account of emotional contagion (\emph{variant C}) and \emph{variant A} above is that the former does not require any reasoning process operating
on actions and goals. It ascribes to the parent the power to impact the child's emotional state more or less directly by producing utterances with a certain emotional
charge. In the account under \emph{variant A} the parents power to affect the child's motivational state is more indirect: The utterance's emotional charge, rather than
impacting the child directly, is taken into account by the child's goal-oriented reasoning process. In order to be efficacious for grounding negative symbols this process
must be social: the reasoning must not only take into account the child's own abilities and goals but also the abilities and goals of the caretaker, and evaluate whether a
caretaker's intervention is likely if a certain action is chosen and given the emotional payload of the previously received utterances and other communicative emotional
signals.

As can be seen from these considerations the degrees of freedom for potential modifications of the learning architecture are many. Only studies with a high temporal resolution
in their description of parental prohibitive behavior, linguistic as extra-linguistic, appear to have the potential to create the required comparative data set. Such a data set could
then not only provide us with better means to evaluate our results but also reduce the degrees of freedom for future modifications of our learning architecture as indicated
above.

In the context of the rejection experiment \cite{Foerster2017} we determined that a lack of proper timing, that is uttering a `no' after a pause longer than the important 1 
second threshold \cite{Jefferson1989}, caused confusion in the coders when trying to determine the meaning of the word, and will presumably cause similar problems with the
interactor. 
Similarly the choice between the hypothesized learning mechanisms in the context of prohibitive utterances could be informed by detailed observations of timing. 
Assuming that emotional appraisal processes are faster than inferential processes, the temporal order between prohibitive utterances and the child's overt emotional displays 
could provide clues as to which of the two types of processes is at play.

\section{Conclusions}
Our architecture is the first to extend symbol grounding beyond the realm of sensorimotor-data to encompass affect and motivation which is in line with recent psychological
studies \cite{Kousta2011}. We have demonstrated the capacity to acquire generally felicitous non-referential linguistic behavior such as negation in a developmental scenario with a
humanoid robot developing an embodied lexicon based on its sensorimotor motivational experience in interaction with na\"ive human participants. Based on our results we
cannot exclude any of the two hypotheses on the origin of negation, due to a lack of sufficiently detailed data on the precise dynamics how prohibition in mother-child dyads
is enacted. We did show however, that at least from a lexical perspective prohibitive utterances are formidable sources of negation words.  

If more detailed data on the dynamics of prohibition were to become available the results of the presented research can give strong indications with respect to the underlying
acquisition algorithm: If the majority of prohibitive utterances are uttered at a time when the child is already frustrated, Hebbian-style methods may suffice to ground
negative words in negative affect. Yet our analysis on the temporal alignment between bodily and linguistic behavior hints towards principal limitations of Hebbian-style
learning. If prohibitive utterances typically precede or may even cause a child's frustration, Hebbian-style methods are unlikely to be a efficacious for affective grounding
because they require a certain amount of synchronicity between negative word and negative affect. In this case a more powerful type of learning algorithm would 
be required. Reinforcement learning, potentially coupled or amplified by some form of social reward signals would be a likely candidate class of learning algorithms. 
This view appears to be supported by recent work in neuroscience which posits a central role of reinforcement learning in biological decision making \cite{Niv2009} if we are
willing to assume that language learning may recruit more general learning mechanisms.


\begin{acks}
The authors would like to thank Kinga Grof for helping with the manual transcription and re-alignment of recorded speech.\\
The work was supported by the \grantsponsor{}{EU Integrated Project ITALK } ((``Integration and Transfer of Action and Language in Robots'') 
funded by the European Commission under contract number \grantnum{FP-7-214668}{FP-7-214668}.
\end{acks}

\clearpage
\appendix

\section{Supplementary Materials}

\begin{printonly}
  See the supplementary materials in the online version
\end{printonly}

\begin{screenonly}
  
\begin{table}[h]
  \caption{Constants for human-robot interaction, all values in seconds}
  \label{tbl_time_constants}
  \begin{threeparttable}
    \begin{tabular}{lcl}
      \hline
      \multicolumn{1}{c}{\textit{variable}} & \multicolumn{1}{c}{\textit{value}} & \multicolumn{1}{c}{\textit{description}}\\
      \hline
      face\_time & 0.8 & duration of iCub looking at face when \textit{pickup}\\
      & & detected and motivation $\ge$ 0\\
      \hline
      object\_time & 3 & duration of iCub looking at object when \textit{pickup}\\
      & &  detected and motivation $\ge$ 0\\
      \hline
      dwell\_time\_face & 1.2 & duration of iCub looking at face when no \textit{pickup}\\
      & &  detected\\
      \hline
      dwell\_time\_object & 2 & duration of iCub looking at \textit{obj} when no \textit{pickup}\\
      & &  detected\\
      \hline
      maxIdleTime & 3 & perceptual timeout for high level percepts: if no\\
      & & objects or faces are perceived for \textit{maxIdleTime},\\
      & &  iCub looks back at the table\\
      \hline
      grumpy\_face\_time\tnote{1} & 1.6 & duration of iCub looking at face if physical\\
      & & restraint is detected (this implies that iCub\\
      & & was reaching for an object)\\
      \hline
      grumpy\_object\_time\tnote{1} & 2 & duration of iCub looking at object if physical\\
      & & restraint is detected\\
      \hline
    \end{tabular}
    \begin{tablenotes}
    \item[1] specific to prohibition scenario
    \end{tablenotes}
  \end{threeparttable}
\end{table}

\begin{figure}[h]
  \begin{center}
    \centerline{\includegraphics[width=.75\textwidth]{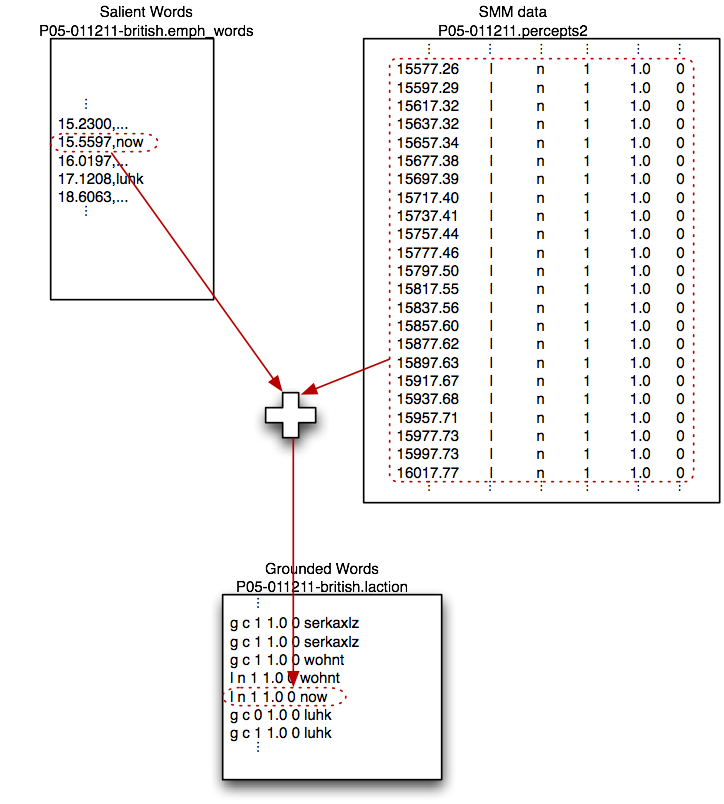}}
    \caption{\textbf{Grounding of (salient) words}. The grounding process associates lexical entries, in our case prosodically salient words,
      with the concurrently occurring sensorimotor-motivational data. In our system the salient word is propagated across the entire duration of
      the utterance, such that the time stamps, visible in the salient-words-file (top-left) mark the start and end of the respective utterance
      within which the word was produced. Time stamps for utterance boundaries are symbolized by `\emph{...}'. Also notice that we remove duplicates
      of grounded words that would ensue from the same utterance. In the given example this means that due to the lack of change within the \emph{smm}
      data during the production of the utterance the potentially ensuing 23 identical grounded words are collapsed into one (bottom).}
    \label{fig_word_grounding}
  \end{center}
\end{figure}

\clearpage
  
\subsection{Pragmatic Analysis - Details}
Figure \ref{robot_neg_types} in the main text shows the taxonomy of negation types the robot engaged in as identified by the external coders. Figure 
\ref{human_neg_types} shows the equivalent taxonomy of negation types engaged in by participants. Tables \ref{tbl_freq_neg_types_pro} and 
\ref{tbl_freq_sal_neg_types_pro} show the absolute frequencies of negative utterances categorized by negation type for the \emph{Prohibition} experiment. 
Tables \ref{tbl_freq_neg_types_re} and \ref{tbl_freq_sal_neg_types_re} give the corresponding frequencies for the \emph{Rejection} experiment.
All four tables form the basis for figure \ref{fig_salience}A in the main text. Table \ref{tbl_word_types} breaks down the uses of the occurring negation words into 
negation types as well as giving absolute and relative frequencies for their salient productions. The combination of these two types of information 
allows us to deduce which particular negation types are numerically speaking the ones most responsible for the occurrence of negation words in the 
robot's lexicon.

Both taxonomies of negation types are derived from the one developed by Roy D. Pea (9) and adapted to the speech encountered in our experiments. Pea's
taxonomy was developed for the various uses of negation of toddlers in the one-word stage and is therefore most similar to the robot's taxonomy. The
construction of the taxonomy for participants' negation types started with conversationally `symmetric' counterparts: Often times negative second
pair-parts such as `No' are preceded by negative first pair-parts such as ``Don't like the square?''.
  
Notice that, in line with Pea's taxonomy, the top-most criterion in both taxonomies is conversational adjacency. This notion is used more loosely than
is the case in the conversation analytic literature (cf. \cite{Hutchby1999}). It encompasses not only adjacency pairs in the strict sense, where the producer
of a second pair-part would be accountable for a non-production, but all utterances that appear to be sequentially linked across speakers, whether the producer
of the second pair-part would be accountable for non-production or not (cf. \cite{Foerster2013} for a more elaborate discussion).

\begin{table*}[h]
  \caption{\textbf{Frequency of participants' negation types - Prohibition Experiment}. Listed are the counts for all negation types of all
    participants and all sessions within the prohibition experiment. `?' is not a negation type but indicates that the coder could not decide
    on a type for a given utterance due to the utterance being incomplete.}
  \label{tbl_freq_neg_types_pro}
  \begin{tabular*}{\hsize}{@{\extracolsep{\fill}}llllllllllll}
    \toprule
    & P13 & P14 & P15 & P16 & P17 & P18 & P19 & P20 & P21 & P22 & total\\
    \midrule
    prohibition & 22 & 16 & 39 & 18 & 31 & 13 & 14 & 24 & 16 & 7 & 200\\
    neg. intent interpret. & 15 & 0 & 38 & 31 & 13 & 30 & 18 & 22 & 4 & 2 & 173\\
    neg. mot. question & 12 & 0 & 52 & 15 & 7 & 14 & 12 & 38 & 20 & 0 & 170\\
    truth-func. denial & 21 & 3 & 3 & 0 & 6 & 2 & 22 & 4 & 7 & 36 & 104\\
    neg. tag question & 1 & 0 & 14 & 30 & 0 & 16 & 1 & 2 & 0 & 3 & 67\\
    disallowance & 0 & 0 & 14 & 4 & 0 & 2 & 3 & 15 & 26 & 1 & 65\\
    truth-func. negation & 0 & 0 & 9 & 12 & 0 & 18 & 4 & 6 & 1 & 0 & 50\\
    neg. agreement & 0 & 0 & 15 & 3 & 10 & 0 & 7 & 3 & 5 & 0 & 43\\
    mot. dep. assertion & 0 & 0 & 3 & 12 & 1 & 1 & 0 & 5 & 1 & 0 & 23\\
    neg. persp. assertion & 1 & 1 & 0 & 4 & 0 & 5 & 1 & 2 & 0 & 1 & 15\\
    negating self-prohibition & 0 & 0 & 0 & 1 & 0 & 0 & 1 & 1 & 4 & 0 & 7\\
    apostr. negation & 0 & 0 & 1 & 1 & 0 & 1 & 0 & 0 & 3 & 0 & 6\\
    neg. imperative & 0 & 0 & 0 & 4 & 0 & 0 & 0 & 0 & 1 & 0 & 5\\
    rejection & 0 & 0 & 0 & 1 & 0 & 0 & 0 & 0 & 3 & 0 & 4\\
    neg. question & 0 & 0 & 0 & 2 & 2 & 0 & 0 & 0 & 0 & 0 & 4\\
    neg. promise & 0 & 0 & 2 & 0 & 0 & 0 & 0 & 1 & 1 & 0 & 4\\
    neg. persp. question & 1 & 1 & 0 & 0 & 0 & 0 & 1 & 0 & 0 & 0 & 3\\
    ? & 0 & 1 & 1 & 0 & 0 & 0 & 0 & 0 & 0 & 0 & 2\\
    mot. dep. exclamation & 0 & 0 & 0 & 1 & 0 & 0 & 0 & 0 & 0 & 0 & 1\\
    \midrule
    total & 73 & 22 & 191 & 139 & 70 & 102 & 84 & 123 & 92 & 50 & 946\\
    \bottomrule
  \end{tabular*}
\end{table*}

\begin{table*}[h]
  \caption{\textbf{Percentage of negation types with salient negative word - Prohibition Experiment}. Listed are the
    percentages of utterances, classified by coder 1 as being of the stated negation type, and in which at least one negation word was detected
    as being salient. All numbers are percentages relative to the total counts given in table \ref{tbl_freq_neg_types_pro}.
    `?' is not a negation type but indicates that the coder could not decide on a type for a given utterance due to the
    utterance being incomplete. The \emph{total} was calculated by weighing each utterance identically which effectively
    gives more weight to the salience rates of speakers who produced more utterances of the respective type.}
  \label{tbl_freq_sal_neg_types_pro}
  \begin{tabular*}{\hsize}{@{\extracolsep{\fill}}llllllllllll}
    \toprule
    & P13 & P14 & P15 & P16 & P17 & P18 & P19 & P20 & P21 & P22 & total\\
    \midrule
    prohibition & 50 & 87.5 & 43.6 & 88.9 & 74.2 & 38.5 & 57.1 & 54.2 & 81.3 & 14.3 & 60.5\\
    neg. mot. question & 33.3 & 0 & 48.1 & 20 & 85.7 & 28.6 & 41.7 & 39.5 & 45 & 0 & 41.8\\
    neg. intent interpret. & 20 & 0 & 44.7 & 32.3 & 46.2 & 33.3 & 44.4 & 36.4 & 50 & 100 & 38.2\\
    truth-func. denial & 19 & 33.3 & 0 & 0 & 16.7 & 0 & 31.8 & 0 & 28.6 & 50 & 31.7\\
    neg. tag question & 0 & 0 & 35.7 & 53.3 & 0 & 56.3 & 100 & 50 & 0 & 33.3 & 49.3\\
    disallowance & 0 & 0 & 28.6 & 75 & 0 & 0 & 100 & 13.3 & 57.7 & 0 & 41.5\\
    neg. agreement & 0 & 0 & 46.7 & 33.3 & 60 & 0 & 71.4 & 66.7 & 80 & 0 & 58.1\\
    truth-func. negation & 0 & 0 & 11.1 & 41.7 & 0 & 27.8 & 25 & 0 & 0 & 0 & 24\\
    mot. dep. assertion & 0 & 0 & 33.3 & 41.7 & 0 & 0 & 0 & 0 & 0 & 0 & 26.1\\
    neg. persp. assertion & 0 & 100 & 0 & 8.3 & 0 & 0 & 100 & 50 & 0 & 0 & 17.4\\
    rejection & 0 & 0 & 0 & 50 & 0 & 0 & 0 & 0 & 66.7 & 0 & 75\\
    neg. question & 0 & 0 & 0 & 100 & 50 & 0 & 0 & 0 & 0 & 0 & 75\\
    negating self-prohibition & 0 & 0 & 0 & 0 & 0 & 0 & 0 & 0 & 50 & 0 & 28.6\\
    neg. imperative & 0 & 0 & 0 & 50 & 0 & 0 & 0 & 0 & 0 & 0 & 40\\
    apostr. negation & 0 & 0 & 0 & 0 & 0 & 100 & 0 & 0 & 33.3 & 0 & 33.3\\
    neg. persp. question & 0 & 100 & 0 & 0 & 0 & 0 & 0 & 0 & 0 & 0 & 33.3\\
    ? & 0 & 100 & 0 & 0 & 0 & 0 & 0 & 0 & 0 & 0 & 50\\
    neg. promise & 0 & 0 & 0 & 0 & 0 & 0 & 0 & 100 & 0 & 0 & 25\\
    mot. dep. exclamation & 0 & 0 & 0 & 0 & 0 & 0 & 0 & 0 & 0 & 0 & 0\\
    \midrule
    total & 30.1 & 81.8 & 40.3 & 46.8 & 61.4 & 33.3 & 46.4 & 35 & 54.3 & 44 & 43.7\\
    \bottomrule
  \end{tabular*}
\end{table*}

\begin{table*}[h]
  \caption{\textbf{Frequency of participants' negation types - Rejection Experiment}. Listed are the counts for all
    negation types of all participants and all sessions within the Rejection Experiment. The last column
    lists the total count for each type across all participants minus the counts of participant P04. This participant
    had to be factored out for the subsequent consideration of salient words because a different method for detecting salient
    words was used. `?' is not a negation type but indicates that the coder could not decide on a type for a given
    utterance due to the utterance being incomplete.}
  \label{tbl_freq_neg_types_re}
  \begin{tabular*}{\hsize}{@{\extracolsep{\fill}}lllllllllllll}
    \toprule
    & P01 & P04 & P05 & P06 & P07 & P08 & P09 & P10 & P11 & P12 & total & total\\
    & & & & & & & & & & & & w/o P04\\
    \midrule
    neg. intent interpret. & 2 & 23 & 36 & 49 & 49 & 18 & 25 & 1 & 19 & 9 & 231 & 208\\
    neg. mot. question & 0 & 24 & 30 & 11 & 72 & 43 & 20 & 9 & 8 & 4 & 221 & 197\\
    truth-func. denial & 18 & 35 & 2 & 1 & 3 & 0 & 9 & 0 & 45 & 39 & 152 & 148\\
    neg. agreement & 0 & 4 & 5 & 0 & 16 & 9 & 1 & 0 & 0 & 0 & 35 & 31\\
    neg. tag question & 0 & 2 & 5 & 7 & 8 & 0 & 7 & 0 & 2 & 0 & 31 & 29\\
    neg. persp. assertion & 0 & 4 & 0 & 1 & 1 & 8 & 6 & 0 & 3 & 3 & 26 & 22\\
    mot. dep. assertion & 0 & 3 & 0 & 0 & 5 & 0 & 12 & 0 & 4 & 0 & 24 & 21\\
    truth-func. negation & 0 & 1 & 0 & 0 & 0 & 2 & 4 & 0 & 7 & 0 & 14 & 13\\
    neg. imperative & 0 & 1 & 0 & 0 & 6 & 0 & 0 & 0 & 0 & 4 & 11 & 10\\
    neg. question & 0 & 3 & 3 & 0 & 1 & 0 & 0 & 0 & 0 & 0 & 7 & 4\\
    apostr. negation & 0 & 1 & 0 & 0 & 0 & 2 & 2 & 0 & 1 & 1 & 7 & 6\\
    truth-func. question & 0 & 0 & 0 & 0 & 0 & 0 & 0 & 0 & 0 & 2 & 2 & 2\\
    neg. persp. question & 0 & 1 & 0 & 0 & 0 & 0 & 0 & 0 & 1 & 0 & 2 & 1\\
    ? & 0 & 0 & 0 & 0 & 0 & 1 & 0 & 0 & 1 & 0 & 2 & 2\\
    rejection & 0 & 0 & 0 & 0 & 0 & 0 & 0 & 0 & 1 & 0 & 1 & 1\\
    mot. dep. exclamation & 0 & 0 & 1 & 0 & 0 & 0 & 0 & 0 & 0 & 0 & 1 & 1\\
    neg. promise & 0 & 0 & 0 & 0 & 0 & 1 & 0 & 0 & 0 & 0 & 1 & 1\\
    \midrule
    total & 20 & 102 & 82 & 69 & 161 & 84 & 86 & 10 & 92 & 62 & 768 & 666\\
    \bottomrule
  \end{tabular*}
\end{table*}

\begin{table*}[h]
  \caption{\textbf{Percentage of negation types with salient negative word - Rejection Experiment}. Listed are the percentages of utterances,
    classified by coder 1 as the stated negation type, (one of) whose negation words were detected as being salient relative to the total number
    of utterances of this type. All numbers are percentages relative to the total counts given in table \ref{tbl_freq_neg_types_re}. The last column lists the average
    percentage of salient negation words across participants minus participant P04. For participant P04, one of the first participants, a different
    algorithm for detecting salient words had been used. `?' is not a negation type but indicates that the coder could not decide on a type for a
    given utterance due to the utterance being incomplete.}
  \label{tbl_freq_sal_neg_types_re}
  \begin{tabular*}{\hsize}{@{\extracolsep{\fill}}lllllllllllll}
    \toprule
    & P01 & P04 & P05 & P06 & P07 & P08 & P09 & P10 & P11 & P12 & total & total\\
    & & & & & & & & & & & & w/o P04\\
    \midrule
    neg. mot. question & 0 & 8.3 & 56.7 & 63.6 & 70.8 & 48.8 & 30 & 22.2 & 12.5 & 50 & 49.3 & 54.3\\
    neg. intent interpret. & 0 & 4.3 & 69.4 & 51 & 42.9 & 44.4 & 48 & 0 & 36.8 & 33.3 & 44.2 & 48.6\\
    truth-func. denial & 44.4 & 0 & 0 & 0 & 66.7 & 0 & 55.6 & 0 & 48.9 & 15.4 & 28.3 & 29.1\\
    neg. agreement & 0 & 0 & 100 & 0 & 87.5 & 44.4 & 100 & 0 & 0 & 0 & 68.6 & 77.4\\
    neg. tag question & 0 & 0 & 80 & 28.6 & 50 & 0 & 42.9 & 0 & 100 & 0 & 48.4 & 51.7\\
    neg. persp. assertion & 0 & 0 & 0 & 100 & 100 & 50 & 33.3 & 0 & 100 & 33.3 & 46.2 & 54.5\\
    neg. question & 0 & 0 & 100 & 0 & 100 & 0 & 0 & 0 & 0 & 0 & 57.1 & 100\\
    neg. imperative & 0 & 0 & 0 & 0 & 0 & 0 & 0 & 0 & 0 & 75 & 27.3 & 30\\
    mot. dep. assertion & 0 & 0 & 0 & 0 & 0 & 0 & 16.7 & 0 & 0 & 0 & 8.3 & 9.5\\
    truth-func. question & 0 & 0 & 0 & 0 & 0 & 0 & 0 & 0 & 0 & 100 & 100 & 100\\
    rejection & 0 & 0 & 0 & 0 & 0 & 0 & 0 & 0 & 100 & 0 & 100 & 100\\
    neg. persp. question & 0 & 100 & 0 & 0 & 0 & 0 & 0 & 0 & 0 & 0 & 50 & 0\\
    ? & 0 & 0 & 0 & 0 & 0 & 100 & 0 & 0 & 0 & 0 & 50 & 50\\
    apostr. negation & 0 & 0 & 0 & 0 & 0 & 0 & 0 & 0 & 0 & 100 & 14.3 & 16.7\\
    truth-func. negation & 0 & 0 & 0 & 0 & 0 & 50 & 0 & 0 & 0 & 0 & 7.1 & 7.7\\
    mot. dep. exclamation & 0 & 0 & 0 & 0 & 0 & 0 & 0 & 0 & 0 & 0 & 0 & 0\\
    neg. promise & 0 & 0 & 0 & 0 & 0 & 0 & 0 & 0 & 0 & 0 & 0 & 0\\
    \midrule
    total & 40 & 3.9 & 65.9 & 50.7 & 58.4 & 46.4 & 36 & 20 & 39.1 & 29 & 41.8 & 47.6\\
    \bottomrule
  \end{tabular*}
\end{table*}

\NewCoffin \TC
\NewCoffin \TCTable
\NewCoffin \TCBlank
\NewCoffin \TCSal
\SetHorizontalCoffin \TC{}
\begin{table*}[h]
  \caption{\textbf{Negation words within most frequent negation types}. \emph{(Top)} Listed are the absolute frequencies of negation words
    grouped by negation types as produced by all participants in all sessions within both experiments. The percentages in brackets give the share
    of the respective word relative to all negative words produced within the respective type. \emph{(Bottom)} Listed are the number of salient
    words for each combination of negation word and type for the most frequently produced types and words. The percentages in brackets give the
    share of salient productions relative to the total number of productions of the respective word-type combination.}
  \label{tbl_word_types}
  \SetHorizontalCoffin \TCTable{
    \begin{tabular*}{\hsize}{@{\extracolsep{\fill}}llllllll}
      \toprule
      \textbf{Type} & neg. intent & neg. mot. & truth-func. & prohi- & disallow- & truth-func. & neg. tag\\
                    & interpret. & question & denial & bition & ance & negation & question\\
      \midrule
      \textbf{Word} & & & & & & &\\
      no & 174  (\small 39.2\%) & 191 (\small 46.2\%) & 212 (\small 67.9\%) & 129 (\small 52.9\%) & 39 (\small 45.3\%) & 14 (\small 21.2\%) & 1 \hspace{1ex}(\small 1\%)\\
      not & 59 \hspace{1ex}{\small (13.3\%)} & 47 \hspace{1ex}{\small (11.4\%)} & 93 \hspace{1ex}{\small (29.8\%)} & 40 \hspace{0.5ex}{\small (16.4\%)} & 27 {\small (31.4\%)}& 30 {\small (45.5\%)} & 0\\
      don't & 201 {\small (45.3\%)} & 164 {\small (39.7\%)} & 1 \hspace{2ex}{\small (0.3\%)} & 2 \hspace{2ex}{\small (0.8\%)} & 2 \hspace{1ex}{\small (2.3\%)} & 2 \hspace{1ex}{\small (3\%)} & 58 {\small (58.6\%)}\\
      isn't & 0 & 9 \hspace{3ex}{\small(2.2\%)} & 4 \hspace{2ex}{\small (1.3\%)} & 0 & 0 & 0 & 18 {\small (18.2\%)}\\
      can't & 0 & 0 & 0 & 68 \hspace{1ex}{\small (27.9\%)} & 16 {\small (18.6\%)} & 1 \hspace{1ex}{\small (1.5\%)}& 3 \hspace{1ex}{\small (3\%)}\\
      haven't & 0 & 0 & 1 \hspace{2ex}{\small (0.3\%)} & 0 & 0 & 7 \hspace{1ex}{\small (10.6\%)} & 1 \hspace{1ex}{\small (1\%)}\\
      wasn't & 0 & 0 & 1 \hspace{2ex}{\small (0.3\%)} & 0 & 0 & 0 & 0\\
      cannot & 0 & 0 & 0 & 1 \hspace{2ex}{\small (0.4\%)} & 1 \hspace{1ex}{\small (1.2\%)} & 0 & 0\\
      neither & 0 & 0 & 0 & 0 & 1 \hspace{1ex}{\small (1.2\%)} & 0 & 0\\
      didn't & 7 \hspace{2ex}{\small (1.6\%)} & 1 \hspace{2ex}{\small (0.2\%)} & 0 & 0 & 0 & 5 \hspace{1ex}{\small (7.6\%)} & 12 {\small (12.1\%)}\\
      doesn't & 1 \hspace{2ex}{\small (0.2\%)} & 0 & 0 & 0 & 0 & 5 \hspace{1ex}{\small (7.6\%)} & 3 \hspace{1ex}{\small (3\%)}\\
      hasn't & 0 & 0 & 0 & 0 & 0 & 1 \hspace{1ex}{\small (1.5\%)}& 2 \hspace{1ex}{\small (2\%)}\\
      weren't & 0 & 0 & 0 & 0 & 0 & 0 & 1 \hspace{1ex}{\small (1\%)}\\
      won't & 2 \hspace{2ex}{\small (0.5\%)} & 1 \hspace{2ex}{\small (0.2\%)} & 0 & 0 & 0 & 0 & 0\\
      mustn't & 0 & 0 & 0 & 4 \hspace{2ex}{\small (1.6\%)} & 0 & 1 \hspace{1ex}{\small (1.5\%)} & 0\\
      \bottomrule
    \end{tabular*}
  }
  \JoinCoffins \TC[vc,hc]\TCTable[vc,hc]
  \SetHorizontalCoffin \TCBlank{
    \begin{minipage}[c][0.5cm][c]{16cm}
    \end{minipage}
  }
  \JoinCoffins \TC[\TCTable-b,\TCTable-l]\TCBlank[t,l]
  \SetHorizontalCoffin \TCSal{
    \begin{tabular*}{.78\hsize}{llllll}
      \toprule
      \textbf{Type} & neg. intent & neg. mot. & truth-func. & prohibition & disallowance\\
                    & interpret. & question & denial & & \\
      \midrule
      \textbf{Word} & & & & &\\
      no & 110 {\small (62.9\%)} & 152 {\small (79.6\%)} & 73 {\small (34.6\%)} & 85 {\small (65.9\%)} & 12 {\small (30.8\%)} \\
      not & 17 \hspace{1ex}{\small (28.8\%)} & 8 \hspace{2ex}{\small (17.0\%)} & 7 \hspace{1ex}{\small (7.5\%)} & 7 \hspace{1ex}{\small (17.5\%)} & 4 \hspace{1ex}{\small (14.8\%)} \\
      don't & 39 \hspace{1ex}{\small (19.3\%)} & 24 \hspace{1ex}{\small (14.7\%)} & 0 & 1 \hspace{1ex}{\small (50.0\%)} & 1 \hspace{1ex}{\small (50.0\%)} \\
      can't & 0 & 0 & 0 & 27 {\small (39.7\%)} & 8 \hspace{1ex}{\small (50.0\%)} \\
      \bottomrule
    \end{tabular*}
  }
  \JoinCoffins \TC[\TCBlank-b,\TCBlank-l]\TCSal[t,l]
  \TypesetCoffin \TC
\end{table*}

\subsection{Corpus / Word Level Analysis - Details}
In this section tables of the word corpora underlying Fig. \ref{fig_production_rates}C are given. Tables \ref{tbl_wf_p} and \ref{tbl_wf_r} contain the word frequencies of most frequent
and other words of interest. Tables \ref{tbl_wf_ps} and \ref{tbl_wf_rs} tabulate the same words but with frequencies for only prosodically salient productions
of these words. Both tables were compiled from speech recorded during the rejection and prohibition experiments. Tables \ref{tbl_wf_s} and \ref{tbl_wf_ss}
contain the equivalent listings for Saunders' et al. experiment.
  
\begin{table*}[h]
  \caption{\textbf{Word-frequencies in Prohibition Experiment}. Listed are the ten most frequent words
    within said experiment across all participants and sessions. Given are the rank, the word count (\emph{cnt}) and the
    percentage relative to the total number of words in the experiment. Apart from the highest-ranking words the same
    statistics are given for object labels, negation words, and words linked to the motivational state of the robot.
    See \cite{Foerster2013} for the complete listing of all words.}
  \label{tbl_wf_p}
  \begin{tabular*}{\hsize}{@{\extracolsep{\fill}}llllllllllll}
    \textsl{rank} & word & cnt & \% & \hspace*{2ex}\textsl{rank} & word & cnt & \% & \hspace*{2ex}\textsl{rank} & word & cnt & \%\\
    \toprule
    \textsl{(1)} & you & 1591 & 6.18 & \hspace*{2ex}\textsl{(27)} & yes & 227 & 0.88 & \hspace*{2ex}\textsl{(114)} & sad & 8 & 0.03\\
    \textsl{(2)} & the & 1416 & 5.5 & \hspace*{2ex}\textsl{(31)} & not & 198 & 0.77 & \hspace*{2ex}\textsl{(116)} & smiling & 6 & 0.02\\
    \textsl{(3)} & a & 962 & 3.74 & \hspace*{2ex}\textsl{(32)} & crescent & 184 & 0.72 & \hspace*{2ex}\textsl{(116)} & haven't & 6 & 0.02\\
    \textsl{(4)} & this & 956 & 3.71 & \hspace*{2ex}\textsl{(33)} & circles & 168 & 0.65 & \hspace*{2ex}\textsl{(117)} & mustn't & 5 & 0.02\\
    \textsl{(5)} & one & 722 & 2.8 & \hspace*{2ex}\textsl{(43)} & Deechee (2) & 132 & 0.51 & \hspace*{2ex}\textsl{(118)} & won't & 4 & 0.02\\
    \textsl{(6)} & is & 632 & 2.45 & \hspace*{2ex}\textsl{(47)} & box & 123 & 0.48 & \hspace*{2ex}\textsl{(119)} & target & 3 & 0.01\\
    \textsl{(7)} & like & 527 & 2.05 & \hspace*{2ex}\textsl{(56)} & Deechee & 103 & 0.4 & \hspace*{2ex}\textsl{(119)} & hasn't & 3 & 0.01\\
    \textsl{(8)} & to & 471 & 1.83 & \hspace*{2ex}\textsl{(58)} & can't (2) & 98 & 0.38 & \hspace*{2ex}\textsl{(120)} & crescents & 2 & 0.01\\
    \textsl{(9)} & no & 461 & 1.79 & \hspace*{2ex}\textsl{(63)} & squares & 86 & 0.33 & \hspace*{2ex}\textsl{(120)} & cannot & 2 & 0.01\\
    \textsl{(10)} & it's & 428 & 1.66 & \hspace*{2ex}\textsl{(66)} & hearts & 79 & 0.31 & \hspace*{2ex}\textsl{(120)} & pyramids & 2 & 0.01\\
    \textsl{(11)} & heart & 411 & 1.6 & \hspace*{2ex}\textsl{(69)} & triangles & 72 & 0.28 & \hspace*{2ex}\textsl{(121)} & shouldn't & 1 & 0\\
    \textsl{(12)} & square & 389 & 1.51 & \hspace*{2ex}\textsl{(75)} & nice & 59 & 0.23 & \hspace*{2ex}\textsl{(121)} & moons & 1 & 0\\
    \textsl{(13)} & triangle & 377 & 1.47 & \hspace*{2ex}\textsl{(81)} & know & 51 & 0.20 & \hspace*{2ex}\textsl{(121)} & nono & 1 & 0\\
    \textsl{(15)} & do & 360 & 1.40 & \hspace*{2ex}\textsl{(88)} & favorite & 34 & 0.13 & \hspace*{2ex}\textsl{(121)} & wouldn't & 1 & 0\\
    \textsl{(16)} & moon & 356 & 1.38 & \hspace*{2ex}\textsl{(100)} & happy & 22 & 0.09 & \hspace*{2ex}\textsl{(121)} & neither & 1 & 0\\
    \textsl{(17)} & circle & 332 & 1.29 & \hspace*{2ex}\textsl{(101)} & smile & 21 & 0.08 & \hspace*{2ex}\textsl{(121)} & pyramid & 1 & 0\\
    \textsl{(19)} & shape & 310 & 1.2 & \hspace*{2ex}\textsl{(102)} & isn't & 20 & 0.08 & \hspace*{2ex}\textsl{(121)} & weren't & 1 & 0\\
    \textsl{(26)} & play & 229 & 0.89 & \hspace*{2ex}\textsl{(103)} & didn't & 19 & 0.07 & & & & \\
    \textsl{(26)} & don't & 229 & 0.89 & \hspace*{2ex}\textsl{(113)} & doesn't & 9 & 0.04 & & & & \\
    \bottomrule
  \end{tabular*}
\end{table*}
\begin{table*}[h]
  \caption{\textbf{Word-frequencies of prosodically salient words in Prohibition Experiment}. Listed are the ten most frequent salient words
    within said experiment across all participants and sessions. Given are the rank, the word count (\emph{cnt}) and the
    percentage relative to the total number of words in the experiment. Apart from the highest-ranking words the same
    statistics are given for object labels, negation words, and words linked to the motivational state of the robot.
    See \cite{Foerster2013} for the complete listing of all words.}
  \label{tbl_wf_ps}
  \begin{tabular*}{\hsize}{@{\extracolsep{\fill}}llllllllllll}
    \textsl{rank} & word & cnt & \% & \hspace*{2ex}\textsl{rank} & word & cnt & \% & \hspace*{2ex}\textsl{rank} & word & cnt & \%\\
    \toprule
    \textsl{(1)} & square & 327 & 4.4 & \hspace*{2ex}\textsl{(25)} & Deechee & 67 & 0.9 & \hspace*{2ex}\textsl{(65)} & know & 9 & 0.12\\
    \textsl{(2)} & triangle & 302 & 4.06 & \hspace*{2ex}\textsl{(27)} & squares & 62 & 0.83 & \hspace*{2ex}\textsl{(69)} & smiling & 5 & 0.07\\
    \textsl{(3)} & circle & 285 & 3.83 & \hspace*{2ex}\textsl{(27)} & is & 62 & 0.83 & \hspace*{2ex}\textsl{(71)} & target & 3 & 0.04\\
    \textsl{(4)} & no & 272 & 3.66 & \hspace*{2ex}\textsl{(32)} & triangles & 50 & 0.67 & \hspace*{2ex}\textsl{(71)} & sad & 3 & 0.04\\
    \textsl{(5)} & one & 250 & 3.36 & \hspace*{2ex}\textsl{(35)} & the & 45 & 0.61 & \hspace*{2ex}\textsl{(72)} & mustn't & 2 & 0.03\\
    \textsl{(6)} & heart & 248 & 3.34 & \hspace*{2ex}\textsl{(35)} & don't & 45 & 0.61 & \hspace*{2ex}\textsl{(72)} & crescents & 2 & 0.03\\
    \textsl{(7)} & this & 230 & 3.09 & \hspace*{2ex}\textsl{(37)} & hearts & 43 & 0.58 & \hspace*{2ex}\textsl{(72)} & cannot & 2 & 0.03\\
    \textsl{(8)} & moon & 208 & 2.8 & \hspace*{2ex}\textsl{(40)} & can't (2) & 39 & 0.53 & \hspace*{2ex}\textsl{(72)} & doesn't & 2 & 0.03\\
    \textsl{(9)} & ok & 171 & 2.3 & \hspace*{2ex}\textsl{(45)} & not & 31 & 0.42 & \hspace*{2ex}\textsl{(72)} & haven't & 2 & 0.03\\
    \textsl{(10)} & shape & 159 & 2.14 & \hspace*{2ex}\textsl{(47)} & do & 29 & 0.39 & \hspace*{2ex}\textsl{(73)} & nono & 1 & 0.01\\
    \textsl{(11)} & yes & 155 & 2.09 & \hspace*{2ex}\textsl{(49)} & favorite & 27 & 0.36 & \hspace*{2ex}\textsl{(73)} & won't & 1 & 0.01\\
    \textsl{(12)} & crescent & 149 & 2 & \hspace*{2ex}\textsl{(53)} & a & 21 & 0.28 & \hspace*{2ex}\textsl{(73)} & hasn't & 1 & 0.01\\
    \textsl{(13)} & like & 134 & 1.8 & \hspace*{2ex}\textsl{(54)} & to & 20 & 0.27 & \hspace*{2ex}\textsl{(73)} & pyramids & 1 & 0.01\\
    \textsl{(14)} & circles & 129 & 1.74 & \hspace*{2ex}\textsl{(54)} & nice & 20 & 0.27 & \hspace*{2ex}\textsl{(73)} & neither & 1 & 0.01\\
    \textsl{(20)} & Deechee (2) & 83 & 1.12 & \hspace*{2ex}\textsl{(57)} & it's & 17 & 0.23 & \hspace*{2ex}\textsl{(73)} & pyramid & 1 & 0.01\\
    \textsl{(21)} & you & 79 & 1.06 & \hspace*{2ex}\textsl{(59)} & smile & 15 & 0.2 & \hspace*{2ex}\textsl{(73)} & weren't & 1 & 0.01\\
    \textsl{(23)} & play & 75 & 1.01 & \hspace*{2ex}\textsl{(64)} & happy & 10 & 0.13 & & & & \\
    \textsl{(24)} & box & 72 & 0.97 & \hspace*{2ex}\textsl{(65)} & isn't & 9 & 0.12 & & & & \\
    \bottomrule
  \end{tabular*}
\end{table*}
\begin{table*}[h]
  \caption{\textbf{Word-frequencies in Rejection Experiment}. Listed are the ten most frequent words
    within said experiment across all participants and sessions. Given are the rank, the word count (\emph{cnt}) and the
    percentage relative to the total number of words in the experiment. Apart from the highest-ranking words the same
    statistics are given for object labels, negation words, and words linked to the motivational state of the robot.
    See \cite{Foerster2013} for the complete listing of all words.}
  \label{tbl_wf_r}
  \begin{tabular*}{\hsize}{@{\extracolsep{\fill}}llllllllllll}
    \textsl{rank} & word & cnt & \% & \hspace*{2ex}\textsl{rank} & word & cnt & \% & \hspace*{2ex}rank & word & cnt & \%\\
    \toprule
    \textsl{(1)} & you & 1245 & 7.13 & \hspace*{2ex}\textsl{(35)} & Deechee & 127 & 0.73 & \hspace*{2ex}\textsl{(93)} & didn't & 11 & 0.06\\
    \textsl{(2)} & the & 983 & 5.63 & \hspace*{2ex}\textsl{(38)} & not & 118 & 0.68 & \hspace*{2ex}\textsl{(94)} & didn't (2) & 10 & 0.06\\
    \textsl{(3)} & like & 579 & 3.31 & \hspace*{2ex}\textsl{(41)} & box & 110 & 0.63 & \hspace*{2ex}\textsl{(95} & pyramid & 9 & 0.05\\
    \textsl{(4)} & a & 475 & 2.72 & \hspace*{2ex}\textsl{(42)} & Deechee (2) & 103 & 0.59 & \hspace*{2ex}\textsl{(93)} & isn't & 11 & 0.06\\
    \textsl{(5)} & this & 471 & 2.7 & \hspace*{2ex}\textsl{(44)} & triangles & 99 & 0.57 & \hspace*{2ex}\textsl{(97)} & moons & 7 & 0.04\\
    \textsl{(6)} & no & 417 & 2.39 & \hspace*{2ex}\textsl{(54)} & don't (2) & 69 & 0.4 & \hspace*{2ex}\textsl{(100)} & rectangle & 4 & 0.02\\
    \textsl{(7)} & one & 396 & 2.27 & \hspace*{2ex}\textsl{(58)} & crescent & 58 & 0.33 & \hspace*{2ex}\textsl{(101)} & won't & 3 & 0.02\\
    \textsl{(8)} & square & 337 & 1.93 & \hspace*{2ex}\textsl{(65)} & sad & 43 & 0.25 & \hspace*{2ex}\textsl{(100)} & smiling & 4 & 0.02\\
    \textsl{(8)} & do & 337 & 1.93 & \hspace*{2ex}\textsl{(68)} & shape & 39 & 0.22 & \hspace*{2ex}\textsl{(101)} & can't & 3 & 0.02\\
    \textsl{(9)} & to & 311 & 1.93 & \hspace*{2ex}\textsl{(69)} & happy & 38 & 0.22 & \hspace*{2ex}\textsl{(101)} & pyramids & 3 & 0.02\\
    \textsl{(11)} & moon & 283 & 1.62 & \hspace*{2ex}\textsl{(69)} & nice & 38 & 0.22 & \hspace*{2ex}\textsl{(103)} & wouldn't & 1 & 0.01\\
    \textsl{(12)} & heart & 279 & 1.6 & \hspace*{2ex}\textsl{(70)} & favorite & 37 & 0.21 & \hspace*{2ex}\textsl{(102)} & doesn't & 2 & 0.01\\
    \textsl{(14)} & triangle & 254 & 1.45 & \hspace*{2ex}\textsl{(71)} & target & 36 & 0.21 & \hspace*{2ex}\textsl{(102)} & doesn't (2) & 2 & 0.01\\
    \textsl{(15)} & circle & 231 & 1.32 & \hspace*{2ex}\textsl{(75)} & hearts & 30 & 0.17 & \hspace*{2ex}\textsl{(103)} & couldn't & 1 & 0.01\\
    \textsl{(17)} & don't & 200 & 1.15 & \hspace*{2ex}\textsl{(78)} & arteen & 27 & 0.15 & \hspace*{2ex}\textsl{(103)} & wasn't & 1 & 0.01\\
    \textsl{(21)} & circles & 190 & 1.09 & \hspace*{2ex}\textsl{(86)} & know & 18 & 0.1 & \hspace*{2ex}\textsl{(103)} & weren't & 1 & 0.01\\
    \textsl{(23)} & squares & 180 & 1.03 & \hspace*{2ex}\textsl{(91)} & smile & 13 & 0.07 & \hspace*{2ex}\textsl{(103)} & can't (2) & 1 & 0.01\\
    \textsl{(24)} & yes & 179 & 1.02 & \hspace*{2ex}\textsl{(92)} & rectangles & 12 & 0.07 & \hspace*{2ex}\textsl{(100)} & haven't & 4 & 0.02\\
    \bottomrule
  \end{tabular*}
\end{table*}
\begin{table*}[h]
  \caption{\textbf{Word-frequencies of prosodically salient words in Rejection Experiment}. Listed are the ten most frequent salient words
    within said experiment across all participants and sessions. Given are the rank, the word count (\emph{cnt}) and the percentage relative
    to the total number of words in the experiment. Apart from the highest-ranking words the same statistics are given for object labels, negation
    words, and words linked to the motivational state of the robot. See \cite{Foerster2013} for the complete listing of all words.}
  \label{tbl_wf_rs}
  \begin{tabular*}{\hsize}{@{\extracolsep{\fill}}llllllllllll}
    \textsl{rank} & word & cnt & \% & \hspace*{2ex}\textsl{rank} & word & cnt & \% & \hspace*{2ex}\textsl{rank} & word & cnt & \%\\
    \toprule
    \textsl{(1)} & square & 259 & 4.97 & \hspace*{2ex}\textsl{(24)} & don't & 52 & 1 & \hspace*{2ex}\textsl{(46)} & do & 12 & 0.23\\
    \textsl{(2)} & no & 242 & 4.64 & \hspace*{2ex}\textsl{(27)} & box & 45 & 0.86 & \hspace*{2ex}\textsl{(47)} & rectangles & 11 & 0.21\\
    \textsl{(3)} & triangle & 206 & 3.95 & \hspace*{2ex}\textsl{(29)} & crescent & 39 & 0.75 & \hspace*{2ex}\textsl{(47)} & a & 11 & 0.21\\
    \textsl{(4)} & heart & 198 & 3.8 & \hspace*{2ex}\textsl{(31)} & are & 36 & 0.69 & \hspace*{2ex}\textsl{(50)} & pyramid & 8 & 0.15\\
    \textsl{(5)} & moon & 184 & 3.53 & \hspace*{2ex}\textsl{(32)} & sad & 35 & 0.67 & \hspace*{2ex}\textsl{(53)} & isn't & 5 & 0.1\\
    \textsl{(5)} & circle & 184 & 3.53 & \hspace*{2ex}\textsl{(33)} & target & 31 & 0.59 & \hspace*{2ex}\textsl{(54)} & rectangle & 4 & 0.08\\
    \textsl{(6)} & like & 167 & 3.2 & \hspace*{2ex}\textsl{(34)} & favorite & 27 & 0.52 & \hspace*{2ex}\textsl{(55)} & pyramids & 3 & 0.06\\
    \textsl{(7)} & circles & 140 & 2.69 & \hspace*{2ex}\textsl{(35)} & arteen & 25 & 0.48 & \hspace*{2ex}\textsl{(55)} & moons & 3 & 0.06\\
    \textsl{(8)} & squares & 126 & 2.42 & \hspace*{2ex}\textsl{(36)} & not & 24 & 0.46 & \hspace*{2ex}\textsl{(55)} & smiling & 3 & 0.06\\
    \textsl{(9)} & it & 123 & 2.36 & \hspace*{2ex}\textsl{(37)} & hearts & 22 & 0.42 & \hspace*{2ex}\textsl{(56)} & can't & 2 & 0.04\\
    \textsl{(10)} & yes & 119 & 2.28 & \hspace*{2ex}\textsl{(38)} & to & 20 & 0.38 & \hspace*{2ex}\textsl{(56)} & won't & 2 & 0.04\\
    \textsl{(11)} & one & 111 & 2.13 & \hspace*{2ex}\textsl{(38)} & happy & 20 & 0.38 & \hspace*{2ex}\textsl{(57)} & didn't (2) & 1 & 0.02\\
    \textsl{(13)} & this & 95 & 1.82 & \hspace*{2ex}\textsl{(39)} & shape & 19 & 0.36 & \hspace*{2ex}\textsl{(57)} & couldn't & 1 & 0.02\\
    \textsl{(18)} & Deechee & 76 & 1.46 & \hspace*{2ex}\textsl{(39)} & nice & 19 & 0.36 & \hspace*{2ex}\textsl{(57)} & doesn't & 1 & 0.02\\
    \textsl{(19)} & Deechee (2) & 68 & 1.3 & \hspace*{2ex}\textsl{(41)} & the & 17 & 0.33 & \hspace*{2ex}\textsl{(57)} & didn't & 1 & 0.02\\
    \textsl{(21)} & triangles & 62 & 1.19 & \hspace*{2ex}\textsl{(46)} & smile & 12 & 0.23 & \hspace*{2ex}\textsl{(57)} & haven't & 1 & 0.02\\
    \textsl{(23)} & you & 53 & 1.02 & \hspace*{2ex}\textsl{(46)} & don't (2) & 12 & 0.23 & \\
    \bottomrule
  \end{tabular*}
\end{table*}
\begin{table*}[h]
  \caption{\textbf{Word-frequencies in the experiment of Saunders et al. \cite{Saunders2012}}. Listed are the ten most frequent words
    within said experiment across all participants and sessions. Given are the rank, the word count (\emph{cnt}) and the
    percentage relative to the total number of words uttered during the entire experiment. Apart from the highest-ranking words the same
    statistics are given for object labels, object properties, negation words, and words linked to the motivational state of the robot.
    See \cite{Foerster2013} for the complete listing of all words.}
  \label{tbl_wf_s}
  \begin{tabular*}{\hsize}{@{\extracolsep{\fill}}llllllllllll}
    \textsl{rank} & word & cnt & \% & \hspace*{2ex}\textsl{rank} & word & cnt & \% & \hspace*{2ex}\textsl{rank} & word & cnt & \% \\
    \toprule
    \textsl{(1)} & a & 702 & 8.71 & \hspace*{2ex}\textsl{(23)} & shape & 88 & 1.09 & \hspace*{2ex}\textsl{(73)} & done & 22 & 0.27\\
    \textsl{(2)} & this & 367 & 4.55 & \hspace*{2ex}\textsl{(24)} & right & 87 & 1.08 & \hspace*{2ex}\textsl{(75)} & don't & 21 & 0.26\\
    \textsl{(3)} & blue & 347 & 4.31 & \hspace*{2ex}\textsl{(25)} & box & 87 & 1.08 & \hspace*{2ex}\textsl{(80)} & crescent & 17 & 0.21\\
    \textsl{(4)} & is & 322 & 4 & \hspace*{2ex}\textsl{(28)} & small & 76 & 0.94 & \hspace*{2ex}\textsl{(101)} & not & 14 & 0.17\\
    \textsl{(5)} & and & 314 & 3.9 & \hspace*{2ex}\textsl{(30)} & square & 69 & 0.86 & \hspace*{2ex}\textsl{(107)} & colors & 11 & 0.14\\
    \textsl{(6)} & red & 302 & 3.75 & \hspace*{2ex}\textsl{(31)} & like & 65 & 0.81 & \hspace*{2ex}\textsl{(109)} & isn't & 11 & 0.14\\
    \textsl{(7)} & green & 286 & 3.55 & \hspace*{2ex}\textsl{(32)} & star & 61 & 0.76 & \hspace*{2ex}\textsl{(137)} & nice & 6 & 0.07\\
    \textsl{(8)} & the & 265 & 3.29 & \hspace*{2ex}\textsl{(40)} & bigger & 41 & 0.51 & \hspace*{2ex}\textsl{(150)} & can't & 4 & 0.05\\
    \textsl{(9)} & that's & 237 & 2.94 & \hspace*{2ex}\textsl{(41)} & white & 41 & 0.51 & \hspace*{2ex}\textsl{(162)} & didn't & 3 & 0.04\\
    \textsl{(10)} & you & 194 & 2.41 & \hspace*{2ex}\textsl{(42)} & large & 40 & 0.5 & \hspace*{2ex}\textsl{(165)} & favorite & 3 & 0.04\\
    \textsl{(11)} & it's & 161 & 2 & \hspace*{2ex}\textsl{(44)} & colour & 37 & 0.46 & \hspace*{2ex}\textsl{(176)} & aren't & 3 & 0.04\\
    \textsl{(12)} & heart & 160 & 1.99 & \hspace*{2ex}\textsl{(53)} & Deechee & 33 & 0.41 & \hspace*{2ex}\textsl{(177)} & squares & 3 & 0.04\\
    \textsl{(13)} & circle & 149 & 1.85 & \hspace*{2ex}\textsl{(54)} & yes & 33 & 0.41 & \hspace*{2ex}\textsl{(178)} & circles & 3 & 0.04\\
    \textsl{(14)} & arrow & 148 & 1.84 & \hspace*{2ex}\textsl{(64)} & smile & 28 & 0.35 & \hspace*{2ex}\textsl{(229)} & triangle & 1 & 0.01\\
    \textsl{(15)} & side & 146 & 1.81 & \hspace*{2ex}\textsl{(65)} & no & 28 & 0.35 & \hspace*{2ex}\textsl{(260)} & never & 1 & 0.01\\
    \textsl{(16)} & cross & 120 & 1.49 & \hspace*{2ex}\textsl{(68)} & shapes & 25 & 0.31 & \hspace*{2ex}\textsl{(264)} & happy & 1 & 0.01\\
    \textsl{(20)} & moon & 95 & 1.18 & \hspace*{2ex}\textsl{(72)} & big & 22 & 0.27 & \hspace*{2ex}\textsl{(273)} & excited & 1 & 0.01\\
    \bottomrule
  \end{tabular*}
\end{table*}
\begin{table*}[h]
  \caption{\textbf{Word-frequencies of prosodically salient words in the experiment of Saunders et al. \cite{Saunders2012}}. Listed are the ten most salient
    words within said experiment across all participants and sessions. Given are the rank, the word count (\emph{cnt}) and the
    percentage relative to the total number of words uttered during the entire experiment. Apart from the highest-ranking words the same
    statistics are given for object labels, negation words, and words linked to the motivational state of the robot.
    See \cite{Foerster2013} for the complete listing of all words.}
  \label{tbl_wf_ss}
  \begin{tabular*}{\hsize}{@{\extracolsep{\fill}}llllllllllll}
    \textsl{rank} & word & cnt & \% & \hspace*{2ex}\textsl{rank} & word & cnt & \% & \hspace*{2ex}\textsl{rank} & word & cnt & \% \\
    \toprule
    \textsl{(1)} & blue & 157 & 6.91 & \hspace*{2ex}\textsl{(19)} & right & 35 & 1.54 & \hspace*{2ex}\textsl{(49)} & no & 10 & 0.44 \\
    \textsl{(2)} & red & 126 & 5.54 & \hspace*{2ex}\textsl{(23)} & colour & 24 & 1.06 & \hspace*{2ex}\textsl{(50)} & it's & 10 & 0.44 \\
    \textsl{(3)} & circle & 117 & 5.15 & \hspace*{2ex}\textsl{(24)} & good & 22 & 0.97 & \hspace*{2ex}\textsl{(52)} & done & 10 & 0.44 \\
    \textsl{(4)} & heart & 108 & 4.75 & \hspace*{2ex}\textsl{(25)} & bigger & 22 & 0.97 & \hspace*{2ex}\textsl{(54)} & the & 8 & 0.35 \\
    \textsl{(5)} & green & 99 & 4.36 & \hspace*{2ex}\textsl{(26)} & a & 21 & 0.92 & \hspace*{2ex}\textsl{(67)} & don't & 6 & 0.26 \\
    \textsl{(6)} & arrow & 81 & 3.56 & \hspace*{2ex}\textsl{(27)} & that's & 20 & 0.88 & \hspace*{2ex}\textsl{(71)} & isn't & 6 & 0.26 \\
    \textsl{(7)} & cross & 79 & 3.48 & \hspace*{2ex}\textsl{(28)} & Deechee & 19 & 0.84 & \hspace*{2ex}\textsl{(72)} & big & 6 & 0.26 \\
    \textsl{(8)} & side & 79 & 3.48 & \hspace*{2ex}\textsl{(29)} & shapes & 18 & 0.79 & \hspace*{2ex}\textsl{(78)} & colors & 5 & 0.22 \\
    \textsl{(9)} & box & 64 & 2.82 & \hspace*{2ex}\textsl{(30)} & it & 18 & 0.79 & \hspace*{2ex}\textsl{(91)} & didn't & 3 & 0.13 \\
    \textsl{(10)} & shape & 55 & 2.42 & \hspace*{2ex}\textsl{(32)} & you & 18 & 0.79 & \hspace*{2ex}\textsl{(93)} & not & 3 & 0.13 \\
    \textsl{(11)} & and & 48 & 2.11 & \hspace*{2ex}\textsl{(33)} & smile & 17 & 0.75 & \hspace*{2ex}\textsl{(97)} & favorite & 3 & 0.13 \\
    \textsl{(12)} & moon & 48 & 2.11 & \hspace*{2ex}\textsl{(36)} & white & 17 & 0.75 & \hspace*{2ex}\textsl{(100)} & circles & 3 & 0.13 \\
    \textsl{(13)} & square & 47 & 2.07 & \hspace*{2ex}\textsl{(38)} & large & 16 & 0.7 & \hspace*{2ex}\textsl{(107)} & nice & 3 & 0.13 \\
    \textsl{(14)} & this & 46 & 2.02 & \hspace*{2ex}\textsl{(41)} & yes & 13 & 0.57 & \hspace*{2ex}\textsl{(119)} & squares & 2 & 0.09 \\
    \textsl{(15)} & star & 42 & 1.85 & \hspace*{2ex}\textsl{(43)} & crescent & 13 & 0.57 & \hspace*{2ex}\textsl{(146)} & can't & 1 & 0.04 \\
    \textsl{(17)} & is & 40 & 1.76 & \hspace*{2ex}\textsl{(46)} & like & 11 & 0.48 & \hspace*{2ex}\textsl{(156)} & aren't & 1 & 0.04 \\
    \textsl{(18)} & small & 35 & 1.54 & \hspace*{2ex}\textsl{(47)} & yea & 11 & 0.48 \\
    \bottomrule
  \end{tabular*}
\end{table*}
\begin{table*}[h]
  \caption{\textbf{Adjusted accumulated word rankings in both negation experiments}. Listed are the 25 top-ranking words of all
    words and of the subset of salient words only within each experiment. The ranking within each list results from conceiving of the
    frequency-ordered word lists for each participant as voting ballots that are ordered descendingly with regard to
    word-frequency. The frequency itself is subsequently ignored. This approach eliminates the greater influence
    of very talkative participants on the accumulated rankings as opposed to the lesser influence of rather taciturn participants.
    The voting ballots were processed by the ranked-pair algorithm which determines the ordered list of winners of this ``voting
    process''. A quote ('"') entry in the rank column indicates a tie: the corresponding word has the same rank as the previous word
    in the column.}
  \label{tbl_adj_rank}
  \begin{center}
    \begin{tabular*}{\hsize}{@{\extracolsep{\fill}}llllllllllll}
      \toprule
      \multicolumn{4}{c}{Rejection Experiment} &\hspace*{5ex} & \multicolumn{4}{c}{Prohibition Experiment}\\
      \cmidrule(r){1-4} \cmidrule(l){6-9}
      \multicolumn{2}{c}{All Words} & \multicolumn{2}{c}{Salient Words} & \hspace*{5ex} & \multicolumn{2}{c}{All Words} &
                                                                                                                          \multicolumn{2}{c}{Salient Words}\\
      \cmidrule(r){1-2}\cmidrule(lr){3-4}\cmidrule(lr){6-7}\cmidrule(l){8-9}
      Rank & Word & Rank & Word & \hspace*{5ex} & Rank & Word & Rank & Word\\
      \midrule
      1 & you & 1 & triangle & \hspace*{5ex} & 1 & you & 1 & square\\
      2 & the & 2 & no & \hspace*{5ex} & 2 & the & 2 & no\\
      3 & a & " & square & \hspace*{5ex}& 3 & this & 3 & circle\\
      4 & like & 3 & heart & \hspace*{5ex} &4 & a & 4 & moon\\
      " & no & 4 & moon & \hspace*{5ex} & 5 & is & 5 & triangle\\
      5 & this & 5 & circle & \hspace*{5ex} & 6 & heart & 6 & heart\\
      6 & square & 6 & yes & \hspace*{5ex} & 7 & one & " & one\\
      7 & one & 7 & like & \hspace*{5ex} & 8 & to & 7 & this\\
      8 & that & 8 & it & \hspace*{5ex} & 9 & no & 8 & yes\\
      9 & do & 9 & squares & \hspace*{5ex} & 10 & like & 9 & like\\
      10 & moon & 10 & this & \hspace*{5ex} & 11 & triangle & " & ok\\
      11 & it & 11 & one & \hspace*{5ex} & " & square & 10 & again\\
      12 & triangle & 12 & ok & \hspace*{5ex} & 12 & it & 11 & shape\\
      13 & circle & " & again & \hspace*{5ex} & 13 & it's & 12 & it\\
      14 & it's & 13 & circles & \hspace*{5ex} & 14 & that & 13 & crescent\\
      15 & heart & 14 & good & \hspace*{5ex} & " & and & 14 & you\\
      16 & to & 15 & oh & \hspace*{5ex} & 15 & circle & 15 & good\\
      17 & yes & 16 & right & \hspace*{5ex} & 16 & very & 16 & very\\
      18 & is & 17 & triangles & \hspace*{5ex} & 17 & do & 17 & ok(2)\\
      19 & ok & 18 & that & \hspace*{5ex} & 18 & moon & 18 & right\\
      20 & oh & 19 & Deechee(2) & \hspace*{5ex} & 19 & that's & 19 & circles\\
      21 & want & 20 & you & \hspace*{5ex} & 20 & shape & 20 & round\\
      22 & don't & 21 & about & \hspace*{5ex} & 21 & can & 21 & Deechee(2)\\
      23 & well & 22 & done & \hspace*{5ex} & 22 & we & 22 & done\\
      24 & circles & 23 & ah & \hspace*{5ex} & 23 & at & 23 & today\\
      \bottomrule
    \end{tabular*}
  \end{center}
\end{table*}

\subsection{Utterance Level Analysis - Details}
In this section the complete measurements for the utterance level are listed as well as the tables relating to the cross-experimental statistical comparison of 
the \emph{utterances per minute} measure. For additional comparisons involving measurements such as the \emph{number of distinct words} or 
\emph{mean length of utterance (MLU)} see \cite{Foerster2013}.
\begin{table*}[h]
  \caption{\textbf{Utterance-level measures for Prohibition Experiment}. All participants and all sessions. Any given number
    refers to the participant with participant id noted on top the corresponding column and the session number in the corresponding
    first column. Abbreviations: \textsl{sX}: session nr. X, \textsl{\# w/\# u}: total number of words/utterances uttered by participant,
    \textsl{\# dw}: number of distinct words, \textsl{MLU}: mean length of utterance, \textsl{w/min / u/min}: words/utterances per minute.}
  \label{tbl_ul_pro}
  \begin{tabular*}{\hsize}{@{\extracolsep{\fill}}clcccccccccc}
    \toprule
    & & P13 & P14 & P15 & P16 & P17 & P18 & P19 & P20 & P21 & P22\\
    \midrule
    \multirow{7}{*}{s1} & d (s) & 301.8 & 317.8 & 311.8 & 332.3 & 303.8 & 301.1 & 318.9 & 300.4 & 359.3 & 319.3\\
    & \# w & 535 & 279 & 611 & 704 & 200 & 691 & 644 & 508 & 332 & 404\\
    & \# u & 134 & 92 & 165 & 194 & 66 & 185 & 179 & 138 & 119 & 124\\
    & \# dw & 110 & 76 & 134 & 195 & 63 & 170 & 178 & 114 & 74 & 99\\
    & MLU & 4 & 3 & 3.7 & 3.6 & 3 & 3.7 & 3.6 & 3.7 & 2.8 & 3.3\\
    & w/min & 106.4 & 52.7 & 117.6 & 127.1 & 39.5 & 137.7 & 121.2 & 101.5 & 55.4 & 75.9\\
    & u/min & 26.6 & 17.4 & 31.8 & 35 & 13 & 36.9 & 33.7 & 27.6 & 19.9 & 23.3\\
    \cmidrule{2-12}
    \multirow{7}{*}{s2} & d (s) & 324.2 & 305.7 & 301.4 & 307 & 310.7 & 296 & 308.4 & 308.3 & 317.6 & 312.1\\
    & \# w & 653 & 307 & 715 & 830 & 215 & 702 & 535 & 610 & 363 & 332\\
    & \# u & 178 & 93 & 187 & 204 & 78 & 188 & 147 & 138 & 133 & 110\\
    & \# dw & 100 & 73 & 126 & 187 & 53 & 188 & 117 & 110 & 75 & 77\\
    & MLU & 3.7 & 3.3 & 3.8 & 4.1 & 2.8 & 3.7 & 3.6 & 4.4 & 2.7 & 3\\
    & w/min & 120.9 & 60.3 & 142.3 & 162.2 & 41.5 & 142.3 & 104.1 & 118.7 & 68.6 & 63.8\\
    & u/min & 32.9 & 18.3 & 37.2 & 39.9 & 15.1 & 38.1 & 28.6 & 26.9 & 25.1 & 21.1\\
    \cmidrule{2-12}
    \multirow{7}{*}{s3} & d (s) & 297.4 & 332.5 & 294.6 & 326.8 & 302.3 & 309.1 & 316.2 & 302.6 & 306.1 & 315.8\\
    & \# w & 424 & 343 & 717 & 774 & 184 & 702 & 610 & 625 & 329 & 477\\
    & \# u & 133 & 95 & 180 & 221 & 71 & 195 & 170 & 137 & 141 & 162\\
    & \# dw & 70 & 70 & 152 & 205 & 52 & 178 & 130 & 87 & 82 & 100\\
    & MLU & 3.2 & 3.6 & 4 & 3.5 & 2.6 & 3.6 & 3.6 & 4.6 & 2.3 & 2.9\\
    & w/min & 85.5 & 61.9 & 146 & 142.1 & 36.5 & 136.3 & 115.8 & 123.9 & 64.5 & 90.6\\
    & u/min & 26.8 & 17.1 & 36.7 & 40.6 & 14.1 & 37.9 & 32.3 & 27.2 & 27.6 & 30.8\\
    \cmidrule{2-12}
    \multirow{7}{*}{s4} & d (s) & 307.6 & 319.9 & 308.5 & 316.1 & 314.7 & 314.8 & 301.7 & 298.6 & 301.1 & 316.2\\
    & \# w & 501 & 298 & 698 & 714 & 259 & 750 & 536 & 490 & 332 & 589\\
    & \# u & 154 & 89 & 181 & 192 & 83 & 198 & 160 & 131 & 119 & 204\\
    & \# dw & 69 & 55 & 132 & 195 & 40 & 194 & 117 & 93 & 59 & 127\\
    & MLU & 3.3 & 3.3 & 3.9 & 3.7 & 3.1 & 3.8 & 3.4 & 3.7 & 2.8 & 2.9\\
    & w/min & 97.7 & 55.9 & 135.7 & 135.5 & 49.4 & 143 & 106.6 & 98.5 & 66.2 & 111.7\\
    & u/min & 30 & 16.7 & 35.2 & 36.4 & 15.8 & 37.7 & 31.8 & 26.3 & 23.7 & 38.7\\
    \cmidrule{2-12}
    \multirow{7}{*}{s5} & d (s) & 306.7 & 380.3 & 312.4 & 320.1 & 293.4 & 302.1 & 311.2 & 303.2 & 306.6 & 317.4\\
    & \# w & 476 & 340 & 656 & 728 & 310 & 716 & 591 & 577 & 333 & 493\\
    & \# u & 160 & 100 & 186 & 188 & 102 & 199 & 178 & 157 & 127 & 170\\
    & \# dw & 64 & 52 & 131 & 198 & 42 & 176 & 109 & 105 & 69 & 106\\
    & MLU & 3 & 3.4 & 3.5 & 3.9 & 3 & 3.6 & 3.3 & 3.7 & 2.6 & 2.9\\
    & w/min & 93.1 & 53.6 & 126 & 136.5 & 63.4 & 142.2 & 113.9 & 114.2 & 65.2 & 93.2\\
    & u/min & 31.3 & 15.8 & 35.7 & 35.2 & 20.9 & 39.5 & 34.3 & 31.1 & 24.9 & 32.1\\
    \bottomrule
  \end{tabular*}
\end{table*}
\begin{table*}[h]
  \caption{\textbf{Utterance-level measures for Rejection Experiment}. All participants and all sessions. Any given number refers
    to the participant with participant id noted on top the corresponding column and the session number in the corresponding
    first column. Abbreviations: \textsl{sX}: session nr. X, \textsl{\# w/\# u}: total number of words/utterances uttered by participant,
    \textsl{\# dw}: number of distinct words, \textsl{MLU}: mean length of utterance, \textsl{w/min / u/min}: average number of words /
    utterances per minute}
  \label{tbl_ul_rej}
  \begin{tabular*}{\hsize}{@{\extracolsep{\fill}}clcccccccccc}
    \toprule
    & & P01 & P04 & P05 & P06 & P07 & P08 & P09 & P10 & P11 & P12\\
    \midrule
    \multirow{7}{*}{s1} & d (s) & 272 & 168.4 & 308.6 & 178.3 & 380.9 & 290.4 & 301.9 & 275.5 & 298.3 & 300.5\\
    & \# w & 111 & 314 & 370 & 392 & 729 & 505 & 474 & 26 & 458 & 353\\
    & \# u & 50 & 67 & 133 & 104 & 225 & 157 & 126 & 16 & 114 & 92\\
    & \# dw & 21 & 101 & 96 & 100 & 140 & 145 & 133 & 6 & 103 & 114\\
    & MLU & 2.2 & 4.7 & 2.8 & 3.8 & 3.2 & 3.2 & 3.8 & 1.6 & 4 & 3.8\\
    & w/min & 24.5 & 111.9 & 71.9 & 131.9 & 114.8 & 104.3 & 94.2 & 5.7 & 92.1 & 70.5\\
    & u/min & 11 & 23.9 & 25.9 & 35 & 35.4 & 32.4 & 25 & 3.5 & 22.9 & 18.4\\
    \cmidrule{2-12}
    \multirow{7}{*}{s2} & d (s) & 285.4 & 196 & 305.2 & 293.2 & 303.4 & 259.8 & 306.6 & 287.6 & 298.8 & 312.7\\
    & \# w & 138 & 346 & 323 & 580 & 666 & 414 & 437 & 70 & 338 & 230\\
    & \# u & 66 & 83 & 110 & 156 & 183 & 112 & 122 & 36 & 106 & 82\\
    & \# dw & 26 & 77 & 71 & 86 & 107 & 89 & 138 & 27 & 90 & 76\\
    & MLU & 2.1 & 4.2 & 2.9 & 3.7 & 3.6 & 3.7 & 3.6 & 1.9 & 3.2 & 2.8\\
    & w/min & 29 & 105.9 & 63.5 & 118.7 & 131.7 & 95.6 & 85.5 & 14.6 & 67.9 & 44.1\\
    & u/min & 13.9 & 25.4 & 21.6 & 31.9 & 36.2 & 25.9 & 23.9 & 7.5 & 21.3 & 15.7\\
    \cmidrule{2-12}
    \multirow{7}{*}{s3} & d (s) & 307.9 & 297.1 & 249.9 & 318 & 307.8 & 296.8 & 299.2 & 290.8 & 306.6 & 302.5\\
    & \# w & 159 & 468 & 302 & 662 & 569 & 431 & 513 & 62 & 242 & 107\\
    & \# u & 66 & 111 & 102 & 168 & 180 & 128 & 139 & 31 & 98 & 38\\
    & \# dw & 29 & 107 & 73 & 97 & 100 & 96 & 155 & 27 & 56 & 22\\
    & MLU & 2.4 & 4.2 & 3 & 3.9 & 3.1 & 3.4 & 3.7 & 2 & 2.5 & 2.8\\
    & w/min & 31 & 94.5 & 72.5 & 124.9 & 110.3 & 87.1 & 102.9 & 12.8 & 47.4 & 21.2\\
    & u/min & 12.9 & 22.4 & 24.5 & 31.7 & 35.1 & 25.9 & 27.9 & 6.4 & 19.2 & 7.5\\
    \cmidrule{2-12}
    \multirow{7}{*}{s4} & d (s) & 319.2 & 265.5 & 213.2 & 329.9 & 307.8 & 301.5 & 300.4 & 289.4 & 300.2 & 303.5\\
    & \# w & 204 & 393 & 253 & 685 & 541 & 364 & 495 & 84 & 187 & 152\\
    & \# u & 80 & 93 & 89 & 187 & 184 & 105 & 143 & 33 & 70 & 67\\
    & \# dw & 30 & 104 & 67 & 95 & 108 & 88 & 152 & 23 & 58 & 27\\
    & MLU & 2.5 & 4.2 & 2.8 & 3.7 & 2.9 & 3.5 & 3.5 & 2.5 & 2.7 & 2.3\\
    & w/min & 38.3 & 88.8 & 71.2 & 124.6 & 105.5 & 72.4 & 99.1 & 17.4 & 37.4 & 30.1\\
    & u/min & 15 & 21 & 25 & 34 & 35.9 & 20.9 & 28.6 & 6.8 & 14 & 13.2\\
    \cmidrule{2-12}
    \multirow{7}{*}{s5} & d (s) & 305.9 & 220.4 & 269.1 & 307.1 & 314.5 & 324.5 & 301.3 & 290.2 & 319.5 & 299\\
    & \# w & 160 & 370 & 356 & 633 & 582 & 406 & 402 & 66 & 211 & 134\\
    & \# u & 61 & 92 & 135 & 157 & 188 & 128 & 132 & 36 & 74 & 54\\
    & \# dw & 23 & 104 & 81 & 89 & 105 & 127 & 116 & 20 & 74 & 36\\
    & MLU & 2.6 & 4 & 2.6 & 4 & 3.1 & 3.2 & 3 & 1.8 & 2.9 & 2.5\\
    & w/min & 31.4 & 100.7 & 79.4 & 123.7 & 111 & 75.1 & 80.1 & 13.6 & 39.6 & 26.9\\
    & u/min & 12 & 25 & 30.1 & 30.7 & 35.9 & 23.7 & 26.3 & 7.4 & 13.9 & 10.8\\
    \bottomrule
  \end{tabular*}
\end{table*}
\begin{table*}[h]
  \caption{\textbf{Utterance-level measures for participants speech from Saunders et al. \cite{Saunders2012}}. Any given number refers to the participant
    with participant id noted on top the corresponding column and the session number in the corresponding first column. Abbreviations: \textsl{sX}:
    session nr. X, \textsl{\# w/\# u}: total number of words/utterances uttered by participant, \textsl{\# dw}: number of distinct words,
    \textsl{MLU}: mean length of utterance, \textsl{w/min / u/min}: average number of words / utterances per minute, \textsl{n/a}: data for
    corresponding session was not available.}
  \begin{tabular*}{\hsize}{@{\extracolsep{\fill}}lllllllllll}
    \toprule
    & & M02 & F05 & M03 & F01 & F02 & M01 & F03 & F06 & F04 \\
    \midrule
    \multirow{7}{*}{s1} & d (s) & 172.7 & 185.5 & 170.1 & 107.1 & 115.2 & 167.9 & n/a & 177.3 & 120.2 \\
    & \# w & 156 & 268 & 120 & 130 & 267 & 210 & n/a & 371 & 290 \\
    & \# u & 51 & 80 & 41 & 34 & 55 & 62 & n/a & 99 & 75 \\
    & \# dw & 29 & 70 & 34 & 42 & 74 & 58 & n/a & 103 & 85 \\
    & MLU & 3.1 & 3.4 & 2.9 & 3.8 & 4.9 & 3.4 & n/a & 3.7 & 3.9 \\
    & w/min & 54.2 & 86.7 & 42.3 & 72.8 & 139.1 & 75.1 & n/a & 125.5 & 144.8 \\
    & u/min & 17.7 & 25.9 & 14.5 & 19 & 28.6 & 22.2 & n/a & 33.5 & 37.5 \\
    \cmidrule{2-11}
    \multirow{7}{*}{s2} & d (s) & 102.4 & 130.6 & 118.9 & 117.9 & 125.5 & 136.9 & 130.7 & 138.7 & 119.7 \\
    & \# w & 105 & 205 & 142 & 145 & 249 & 219 & 214 & 264 & 178 \\
    & \# u & 34 & 48 & 45 & 35 & 55 & 63 & 61 & 77 & 60 \\
    & \# dw & 24 & 77 & 37 & 37 & 65 & 44 & 56 & 83 & 41 \\
    & MLU & 3.1 & 4.3 & 3.2 & 4.1 & 4.5 & 3.5 & 3.5 & 3.4 & 3 \\
    & w/min & 61.5 & 94.2 & 71.7 & 73.8 & 119 & 95.9 & 98.3 & 114.2 & 89.2 \\
    & u/min & 19.9 & 22.1 & 22.7 & 17.8 & 26.3 & 27.6 & 28 & 33.3 & 30.1 \\
    \cmidrule{2-11}
    \multirow{7}{*}{s3} & d (s) & 119.3 & 129.7 & 115.5 & 114 & 122.7 & 129 & 123.3 & 133.7 & 128.4 \\
    & \# w & 99 & 215 & 97 & 123 & 236 & 162 & 200 & 278 & 220 \\
    & \# u & 31 & 57 & 36 & 34 & 52 & 42 & 64 & 73 & 69 \\
    & \# dw & 21 & 57 & 39 & 27 & 57 & 36 & 67 & 79 & 61 \\
    & MLU & 3.2 & 3.8 & 2.7 & 3.6 & 4.5 & 3.9 & 3.1 & 3.8 & 3.2 \\
    & w/min & 49.8 & 99.4 & 50.4 & 64.8 & 115.4 & 75.3 & 97.3 & 124.7 & 102.8 \\
    & u/min & 15.6 & 26.4 & 18.7 & 17.9 & 25.4 & 19.5 & 31.1 & 32.8 & 32.2 \\
    \cmidrule{2-11}
    \multirow{7}{*}{s4} & d (s) & 125.9 & 118.3 & 116.4 & 115.8 & 122.5 & 126.8 & 116.7 & 128.5 & 126.5 \\
    & \# w & 90 & 174 & 82 & 107 & 172 & 127 & 200 & 273 & 192 \\
    & \# u & 35 & 40 & 31 & 28 & 39 & 38 & 61 & 70 & 60 \\
    & \# dw & 18 & 44 & 29 & 29 & 43 & 25 & 66 & 86 & 52 \\
    & MLU & 2.6 & 4.3 & 2.6 & 3.8 & 4.4 & 3.3 & 3.3 & 3.9 & 3.2 \\
    & w/min & 42.9 & 88.2 & 42.3 & 55.4 & 84.3 & 60.1 & 102.8 & 127.5 & 91 \\
    & u/min & 16.7 & 20.3 & 16 & 14.5 & 19.1 & 18 & 31.4 & 32.7 & 28.4 \\
    \cmidrule{2-11}
    \multirow{7}{*}{s5} & d (s) & 205.7 & 107.4 & 125.2 & 113.8 & 117.7 & 102.5 & 130.7 & 128.6 & 126.5 \\
    & \# w & 160 & 182 & 99 & 122 & 200 & 93 & 233 & 234 & 155 \\
    & \# u & 53 & 47 & 35 & 28 & 43 & 32 & 76 & 67 & 59 \\
    & \# dw & 24 & 48 & 37 & 29 & 45 & 23 & 57 & 83 & 46 \\
    & MLU & 3 & 3.9 & 2.8 & 4.4 & 4.7 & 2.9 & 3.1 & 3.5 & 2.6 \\
    & w/min & 46.7 & 101.6 & 47.4 & 64.3 & 102 & 54.5 & 106.9 & 109.2 & 73.5 \\
    & u/min & 15.5 & 26.2 & 16.8 & 14.8 & 21.9 & 18.7 & 34.9 & 31.3 & 28 \\
    \bottomrule
  \end{tabular*}
\end{table*}
\begin{table*}[h]
  \caption{\textbf{Utterance-level measures for negative utterances in Prohibition Experiment}. Numbers refer to the participant with
    the id noted in the top row and session number in the first column. Abbreviations: \textsl{sX}: session nr. X, \textsl{\# nw/\# nu}: total
    number of negation words/negative utterances uttered by participant, \textsl{\# dnw}: number of unique negation words, \textsl{MLU}: mean
    length of utterance, \textsl{nw/min / nu/min}: negation words / negative utterances per minute}
  \begin{tabular*}{\hsize}{@{\extracolsep{\fill}}clcccccccccc}
    \toprule
    & & P13 & P14 & P15 & P16 & P17 & P18 & P19 & P20 & P21 & P22\\
    \midrule
    \multirow{7}{*}{s1} & d (s) & 301.8 & 317.8 & 311.8 & 332.3 & 303.8 & 301.1 & 318.9 & 300.4 & 359.3 & 319.3\\
    & \# nw & 22 & 0 & 61 & 29 & 16 & 36 & 22 & 24 & 33 & 5\\
    & \# nu & 19 & 0 & 52 & 25 & 14 & 36 & 20 & 23 & 20 & 4\\
    & \# dnw & 7 & 0 & 8 & 5 & 4 & 7 & 5 & 6 & 3 & 4\\
    & MLU & 4.8 & 0 & 4.5 & 5.6 & 2.7 & 4.4 & 3.8 & 3.9 & 3.6 & 4.8\\
    & nw/min & 4.4 & 0 & 11.7 & 5.2 & 3.2 & 7.2 & 4.1 & 4.8 & 5.5 & 0.9\\
    & nu/min & 3.8 & 0 & 10 & 4.5 & 2.8 & 7.2 & 3.8 & 4.6 & 3.3 & 0.8\\
    \cmidrule{2-12}
    \multirow{7}{*}{s2} & d (s) & 324.2 & 305.7 & 301.4 & 307 & 310.7 & 296 & 308.4 & 308.3 & 317.6 & 312.1\\
    & \# nw & 43 & 12 & 45 & 34 & 18 & 19 & 29 & 34 & 36 & 15\\
    & \# nu & 36 & 8 & 37 & 32 & 16 & 18 & 24 & 30 & 24 & 15\\
    & \# dnw & 5 & 3 & 6 & 6 & 3 & 6 & 4 & 5 & 4 & 2\\
    & MLU & 4.4 & 2.4 & 5.2 & 4.8 & 2.1 & 4.7 & 4.4 & 5.7 & 3.1 & 3.1\\
    & nw/min & 8 & 2.4 & 9.0 & 6.6 & 3.5 & 3.9 & 5.6 & 6.6 & 6.8 & 2.9\\
    & nu/min & 6.7 & 1.6 & 7.4 & 6.3 & 3.1 & 3.6 & 4.7 & 5.8 & 4.5 & 2.9\\
    \cmidrule{2-12}
    \multirow{7}{*}{s3} & d (s) & 297.4 & 332.5 & 294.6 & 326.8 & 302.3 & 309.1 & 316.2 & 302.6 & 306.1 & 315.8\\
    & \# nw & 7 & 9 & 49 & 34 & 26 & 25 & 29 & 33 & 25 & 19\\
    & \# nu & 7 & 8 & 45 & 28 & 25 & 23 & 25 & 30 & 24 & 18\\
    & \# dnw & 3 & 3 & 7 & 6 & 5 & 7 & 4 & 5 & 4 & 3\\
    & MLU & 4.4 & 3 & 5.0 & 5.0 & 1.9 & 4.9 & 4.7 & 5.3 & 2.4 & 2.7\\
    & nw/min & 1.4 & 1.6 & 10 & 6.2 & 5.2 & 4.9 & 5.5 & 6.5 & 4.9 & 3.6\\
    & nu/min & 1.4 & 1.4 & 9.2 & 5.1 & 5 & 4.5 & 4.7 & 5.9 & 4.7 & 3.4\\
    \cmidrule{2-12}
    \multirow{7}{*}{s4} & d (s) & 307.6 & 319.9 & 308.5 & 316.1 & 314.7 & 314.8 & 301.7 & 298.6 & 301.1 & 316.2\\
    & \# nw & 5 & 1 & 30 & 28 & 8 & 15 & 6 & 21 & 9 & 7\\
    & \# nu & 5 & 1 & 24 & 25 & 8 & 14 & 6 & 20 & 8 & 7\\
    & \# dnw & 3 & 1 & 5 & 6 & 1 & 6 & 3 & 6 & 2 & 3\\
    & MLU & 3.8 & 1 & 5.0 & 5.5 & 1.9 & 5.1 & 4.7 & 4.3 & 4 & 3.9\\
    & nw/min & 1 & 0.2 & 5.8 & 5.3 & 1.5 & 2.9 & 1.2 & 4.2 & 1.8 & 1.3\\
    & nu/min & 1 & 0.2 & 4.7 & 4.7 & 1.5 & 2.7 & 1.2 & 4.0 & 1.6 & 1.3\\
    \cmidrule{2-12}
    \multirow{7}{*}{s5} & d (s) & 306.7 & 380.3 & 312.4 & 320.1 & 293.4 & 302.1 & 311.2 & 303.2 & 306.6 & 317.4\\
    & \# nw & 6 & 3 & 32 & 28 & 7 & 11 & 10 & 20 & 16 & 6\\
    & \# nu & 6 & 3 & 30 & 27 & 7 & 11 & 9 & 20 & 16 & 6\\
    & \# dnw & 3 & 2 & 3 & 6 & 1 & 4 & 3 & 5 & 2 & 3\\
    & MLU & 5.3 & 1.3 & 2.8 & 5.3 & 1.3 & 4.6 & 4.3 & 4.5 & 2.5 & 2.8\\
    & nw/min & 1.2 & 0.5 & 6.1 & 5.2 & 1.4 & 2.2 & 1.9 & 4 & 3.1 & 1.1\\
    & nu/min & 1.2 & 0.5 & 5.8 & 5.1 & 1.4 & 2.2 & 1.7 & 4 & 3.1 & 1.1\\
    \bottomrule
  \end{tabular*}
\end{table*}
\begin{table*}[h]
  \caption{\textbf{Utterance-level measures for negative utterances in Rejection Experiment}. All numbers refer to the participant with
    the id noted in the top row and session number in the first column. Abbreviations: \textsl{sX}: session nr. X, \textsl{\# nw/\# nu}: total
    number of negation words/negative utterances uttered by participant, \textsl{\# dnw}: number of unique negation words, \textsl{MLU}: mean
    length of utterance , \textsl{nw/min / nu/min}: negation words / negative utterances per minute}
  \begin{tabular*}{\hsize}{@{\extracolsep{\fill}}clcccccccccc}
    \toprule
    & & P01 & P04 & P05 & P06 & P07 & P08 & P09 & P10 & P11 & P12\\
    \midrule
    \multirow{7}{*}{s1} & d (s) & 272 & 168.4 & 308.6 & 178.3 & 380.9 & 290.4 & 301.9 & 275.5 & 298.3 & 300.5\\
    & \# nw & 1 & 15 & 25 & 11 & 30 & 22 & 18 & 0 & 25 & 14\\
    & \# nu & 1 & 12 & 24 & 10 & 28 & 21 & 18 & 0 & 21 & 13\\
    & \# dnw & 1 & 5 & 4 & 3 & 4 & 3 & 5 & 0 & 5 & 3\\
    & MLU & 2 & 4.4 & 1.8 & 3.7 & 3.2 & 3.5 & 4.5 & 0 & 5.6 & 3.8\\
    & nw/min & 0.2 & 5.3 & 4.9 & 3.7 & 4.7 & 4.5 & 3.6 & 0 & 5 & 2.8\\
    & nu/min & 0.2 & 4.3 & 4.7 & 3.4 & 4.4 & 4.3 & 3.6 & 0 & 4.2 & 2.6\\
    \cmidrule{2-12}
    \multirow{7}{*}{s2} & d (s) & 285.4 & 196 & 305.2 & 293.2 & 303.4 & 259.8 & 306.6 & 287.6 & 298.8 & 312.7\\
    & \# nw & 3 & 23 & 21 & 13 & 35 & 18 & 21 & 2 & 41 & 17\\
    & \# nu & 3 & 19 & 20 & 12 & 34 & 17 & 18 & 2 & 29 & 14\\
    & \# dnw & 2 & 4 & 3 & 2 & 4 & 3 & 4 & 2 & 4 & 3\\
    & MLU & 2.7 & 4.9 & 3 & 3.5 & 4.1 & 4.5 & 6.2 & 2.5 & 4.2 & 2.8\\
    & nw/min & 0.6 & 7 & 4.1 & 2.7 & 6.9 & 4.2 & 4.1 & 0.4 & 8.2 & 3.3\\
    & nu/min & 0.6 & 5.8 & 3.9 & 2.5 & 6.7 & 3.9 & 3.5 & 0.4 & 5.8 & 2.7\\
    \cmidrule{2-12}
    \multirow{7}{*}{s3} & d (s) & 307.9 & 297.1 & 249.9 & 318 & 307.8 & 296.8 & 299.2 & 290.8 & 306.6 & 302.5\\
    & \# nw & 5 & 32 & 10 & 18 & 37 & 15 & 23 & 2 & 18 & 11\\
    & \# nu & 5 & 23 & 10 & 17 & 35 & 15 & 20 & 2 & 15 & 8\\
    & \# dnw & 2 & 4 & 3 & 4 & 5 & 4 & 6 & 2 & 3 & 1\\
    & MLU & 3.4 & 5.3 & 1.2 & 4.6 & 3 & 2.3 & 6.2 & 3 & 4.4 & 4\\
    & nw/min & 1 & 6.5 & 2.4 & 3.4 & 7.2 & 3.0 & 4.6 & 0.4 & 3.5 & 2.2\\
    & nu/min & 1 & 4.6 & 2.4 & 3.2 & 6.8 & 3.0 & 4 & 0.4 & 2.9 & 1.6\\
    \cmidrule{2-12}
    \multirow{7}{*}{s4} & d (s) & 319.2 & 265.5 & 213.2 & 329.9 & 307.8 & 301.5 & 300.4 & 289.4 & 300.2 & 303.5\\
    & \# nw & 8 & 26 & 12 & 17 & 25 & 21 & 14 & 3 & 21 & 18\\
    & \# nu & 8 & 20 & 12 & 17 & 24 & 20 & 14 & 3 & 20 & 14\\
    & \# dnw & 2 & 2 & 5 & 4 & 5 & 2 & 6 & 1 & 5 & 2\\
    & MLU & 1.8 & 4.6 & 3.2 & 3.9 & 3.2 & 3.2 & 5.1 & 4.7 & 3.4 & 2.6\\
    & nw/min & 1.5 & 5.9 & 3.4 & 3.1 & 4.9 & 4.2 & 2.8 & 0.6 & 4.2 & 3.6\\
    & nu/min & 1.5 & 4.5 & 3.4 & 3.1 & 4.7 & 4 & 2.8 & 0.6 & 4 & 2.8\\
    \cmidrule{2-12}
    \multirow{7}{*}{s5} & d (s) & 305.9 & 220.4 & 269.1 & 307.1 & 314.5 & 324.5 & 301.3 & 290.2 & 319.5 & 299\\
    & \# nw & 3 & 36 & 19 & 12 & 40 & 11 & 17 & 3 & 9 & 14\\
    & \# nu & 3 & 28 & 16 & 12 & 40 & 11 & 15 & 3 & 6 & 13\\
    & \# dnw & 2 & 5 & 3 & 3 & 3 & 2 & 3 & 1 & 3 & 2\\
    & MLU & 3.7 & 4.7 & 2.9 & 4.4 & 2.7 & 3.9 & 3.4 & 2 & 3.7 & 3.3\\
    & nw/min & 0.6 & 9.8 & 4.2 & 2.3 & 7.6 & 2 & 3.4 & 0.6 & 1.7 & 2.8\\
    & nu/min & 0.6 & 7.6 & 3.6 & 2.3 & 7.6 & 2 & 3.0 & 0.6 & 1.1 & 2.6\\
    \bottomrule
  \end{tabular*}
\end{table*}
\begin{table*}[h]
  \begin{center}
    \caption{\textbf{Utterance-level measures for negative utterances of participants speech from Saunders et al. \cite{Saunders2012}}.
      Any given number refers to the participant with participant id noted on top the corresponding column and the session number in the
      corresponding first column. Abbreviations: \textsl{sX}: session nr. X, \textsl{\# nw/\# nu}: total number of negative words/utterances uttered
      by participant, \textsl{\# dnw}: number of distinct negative words, \textsl{MLU}: mean length of negative utterances, \textsl{nw/min / nu/min}:
      average number of neg. words / neg. utterances per minute, \textsl{n/a}: data for corresponding session was not available.}
    \label{tbl_neg_utt_S}
    \begin{tabular*}{\hsize}{@{\extracolsep{\fill}}lllllllllll}
      \toprule
      & & M02 & F05 & M03 & F01 & F02 & M01 & F03 & F06 & F04\\
      \midrule
      \multirow{7}{*}{s1} & d (s) & 172.7 & 185.5 & 170.1 & 107.1 & 115.2 & 167.9 & n/a & 177.3 & 120.2\\
      & \# nw & 0 & 0 & 2 & 0 & 0 & 0 & n/a & 11 & 2\\
      & \# nu & 0 & 0 & 1 & 0 & 0 & 0 & n/a & 10 & 2\\
      & \# dnw & 0 & 0 & 2 & 0 & 0 & 0 & n/a & 5 & 1\\
      & MLU & 0 & 0 & 19 & 0 & 0 & 0 & n/a & 6.4 & 10\\
      & nw/min & 0 & 0 & 0.7 & 0 & 0 & 0 & n/a & 3.7 & 1\\
      & nu/min & 0 & 0 & 0.4 & 0 & 0 & 0 & n/a & 3.4 & 1\\
      \cmidrule{2-11}
      \multirow{7}{*}{s2} & d (s) & 102.4 & 130.6 & 118.9 & 117.9 & 125.5 & 136.9 & 130.7 & 138.7 & 119.7\\
      & \# nw & 0 & 0 & 1 & 5 & 0 & 0 & 0 & 14 & 1\\
      & \# nu & 0 & 0 & 1 & 4 & 0 & 0 & 0 & 13 & 1\\
      & \# dnw & 0 & 0 & 1 & 2 & 0 & 0 & 0 & 6 & 1\\
      & MLU & 0 & 0 & 3 & 5.8 & 0 & 0 & 0 & 5.6 & 6\\
      & nw/min & 0 & 0 & 0.5 & 2.5 & 0 & 0 & 0 & 6.1 & 0.5\\
      & nu/min & 0 & 0 & 0.5 & 2 & 0 & 0 & 0 & 5.6 & 0.5\\
      \cmidrule{2-11}
      \multirow{7}{*}{s3} & d (s) & 119.3 & 129.7 & 115.5 & 114 & 122.7 & 129 & 123.3 & 133.7 & 128.4\\
      & \# nw & 0 & 1 & 3 & 2 & 0 & 0 & 0 & 14 & 3\\
      & \# nu & 0 & 1 & 3 & 2 & 0 & 0 & 0 & 13 & 2\\
      & \# dnw & 0 & 1 & 2 & 1 & 0 & 0 & 0 & 5 & 1\\
      & MLU & 0 & 8 & 3.7 & 5 & 0 & 0 & 0 & 3 & 2\\
      & nw/min & 0 & 0.5 & 1.6 & 1.1 & 0 & 0 & 0 & 6.3 & 1.4\\
      & nu/min & 0 & 0.5 & 1.6 & 1.1 & 0 & 0 & 0 & 5.8 & 0.9\\
      \cmidrule{2-11}
      \multirow{7}{*}{s4} & d (s) & 125.9 & 118.3 & 116.4 & 115.8 & 122.5 & 126.8 & 116.7 & 128.5 & 126.5\\
      & \# nw & 0 & 0 & 2 & 0 & 0 & 0 & 0 & 14 & 0\\
      & \# nu & 0 & 0 & 2 & 0 & 0 & 0 & 0 & 12 & 0\\
      & \# dnw & 0 & 0 & 1 & 0 & 0 & 0 & 0 & 6 & 0\\
      & MLU & 0 & 0 & 2.5 & 0 & 0 & 0 & 0 & 6.6 & 0\\
      & nw/min & 0 & 0 & 1 & 0 & 0 & 0 & 0 & 6.5 & 0\\
      & nu/min & 0 & 0 & 1 & 0 & 0 & 0 & 0 & 5.6 & 0\\
      \cmidrule{2-11}
      \multirow{7}{*}{s5} & d (s) & 205.7 & 107.4 & 125.2 & 113.8 & 117.7 & 102.5 & 130.7 & 128.6 & 126.5\\
      & \# nw & 0 & 0 & 1 & 1 & 0 & 0 & 1 & 7 & 0\\
      & \# nu & 0 & 0 & 1 & 1 & 0 & 0 & 1 & 6 & 0\\
      & \# dnw & 0 & 0 & 1 & 1 & 0 & 0 & 1 & 4 & 0\\
      & MLU & 0 & 0 & 4 & 4 & 0 & 0 & 5 & 6.5 & 0\\
      & nw/min & 0 & 0 & 0.5 & 0.5 & 0 & 0 & 0.5 & 3.3 & 0\\
      & nu/min & 0 & 0 & 0.5 & 0.5 & 0 & 0 & 0.5 & 2.8 & 0\\
      \bottomrule
    \end{tabular*}
  \end{center}
\end{table*}
\clearpage
\begin{table}[h]
  \caption{Comparison of \emph{u/min} between negation experiments and
    Saunders' et al. experiment.}
  \begin{minipage}[t][0.7cm][t]{9.5cm}
    {\small\textbf{(a)} All Utterances}\\
  \end{minipage}
  \setlength{\tabcolsep}{0.9ex}
  \begin{tabular}{llllll}
    \toprule
    & Saunders & Rejection & Prohibition \\
    & mean (sd) & mean (sd) & mean(sd) & F(2,26)$^*$ & p\\
    \midrule
    s1 & 24.86 (8.01) & 23.34 (10.2) & 26.52 (8.03) & 0.32 & 0.7290\\
    s2 & 25.31 (5.04) & 22.33 (8.47) & 28.32 (8.62) & 1.54 & 0.2340\\
    s3 & 24.4 (6.68) & 21.35 (9.79) & 29.11 (8.57) & 2.11 & 0.1410\\
    s4 & 21.9 (6.99) & 21.44 (9.49) & 29.23 (8.38) & 2.67 & 0.0882\\
    s5 & 23.12 (7.31) & 21.58 (9.82) & 30.08 (7.35) & 2.97 & 0.0689\\
    \bottomrule
  \end{tabular}
  \begin{minipage}[t][1cm][t]{9.5cm}
    \vspace*{2ex}
    {\small\textbf{(a)} Negative Utterances Only}\\
  \end{minipage}
  \begin{tabular}{llllll}
    \toprule
    & Saunders & Rejection & Prohibition \\
    & mean (sd) & mean (sd) & mean(sd) & F(2,26)$^*$ & p \\
    \midrule
    s1 & 0.6 (1.19) & 3.17 (1.73) & 4.08 (2.89) & 6.31 & 0.0060\\
    s2 & 0.96 (1.86) & 3.58 (2.13) & 4.66 (1.88) & 8.83 & 0.0012\\
    s3 & 1.1 (1.86) & 2.99 (1.86) & 4.53 (2.24) & 6.98 & 0.0037\\
    s4 & 0.73 (1.85) & 3.14 (1.3) & 2.29 (1.63) & 5.45 & 0.0105\\
    s5 & 0.48 (0.9) & 3.1 (2.57) & 2.61 (1.82) & 4.93 & 0.0153\\
    \bottomrule
  \end{tabular}
\end{table}

\subsection{Robot: Evaluation of Acquisition}
Table \ref{tbl:coder_felicity} compares the relative felicity rates of the robot's negative utterances as judged by each coder. Initially we also
tentatively included so called \emph{pragmatic negatives} in the coding set which are words which in a certain context may fulfill the same communicative function
as a lexical negative. `Go' for example, if expressed with a certain assertive prosodic contour, may do the same communicative work as `no' in a situation where
the participant asks whether Deechee wants to play with a particular box or not. We excluded these `negatives' in the end because they constituted a source of
disagreement amongst coders. Participant P12 receives this special treatment for the following reason: upon completion of his five sessions, we were made aware of
his conscious decision to adopt an unnatural speech register in order to `win the teaching game', based on some hypothesis of his regarding the underlying acquisition
algorithm. He thereby explicitly ignored our instruction to speak to the robot as if it was a small child.

Table \ref{tbl:accu_felicity} depicts the outcome of the evaluation of the robot's negative linguistic productions for felicity, that is whether
its utterances would be considered adequate in the respective situations as judged by an external observer. In line with our study design the impact
of the prohibition task is measured by comparing the robot's `negative' linguistic performance of the last two sessions (table \ref{tbl:accu_felicity_s45}).
Table S\ref{tbl:felicity_robot_stat} shows the result of a t-test which was performed in order to ascertain that the results of table \ref{tbl:accu_felicity_s45}
were not skewed by those sessions were Deechee was more talkative, i.e. where it uttered more negative utterances. The test is based on the felicity rates of the robot on a
per-participant basis where the rates of all sessions are accumulated (row \emph{all} in table \ref{tbl:accu_felicity_s45}) and where two different exclusion criteria
were applied: Under criterion 1 all available data is used but from those set of sessions where the robot did not produce any negative words. 
Under criterion 2 those participants' felicity rates are excluded which are based on a total count lower than 10.

\begin{table}[h]
  \caption{\textbf{Felicity rate judgements by coder}: \emph{full}: full coding set, no exclusions, \emph{-prag.}: coding set where utterances containing pragmatic
    negatives are removed, \emph{-P12}: coding set where \emph{P12}'s utterances are excluded, \emph{-P12 \&\& -prag.}: coding set where both \emph{P12}'s and
    pragmatic negatives are excluded.}
  \label{tbl:coder_felicity}
  \setlength{\tabcolsep}{2.325ex}
  \begin{tabular}{ccccc}
    \toprule
    coder & full & - prag. & - P12 & - P12 \&\& -prag. \\
    \midrule
    c1 & 53.91 & 46.94 & 54.88 & 48.57 \\
    c2 & 51.94 & 48.98 & 54.35 & 55.71 \\
    \bottomrule
  \end{tabular}
\end{table}
\NewCoffin \AccuFelTable
\NewCoffin \AccuFelTableR
\NewCoffin \AccuFelTableP
\begin{table*}[h] 
  \setlength{\tabcolsep}{0.35ex}
  \caption{\textbf{Accumulated frequencies for robot negation types and their felicity}. Displayed are the accumulated frequencies of the various negation
    types the robot engaged in across sessions and their felicities. The following abbreviations are used for the negation types: \emph{TD}: truth-func. denial,
    \emph{MD}: mot.-dep. denial, \emph{A}: neg. agreement, \emph{R}: rejection, \emph{I}: neg. imperative, \emph{E}: mot. dep. exclamation, \emph{SP}:
    self-prohibition, \emph{PD}: perspective-dependent denial. See the SI coding scheme for a description of each type including examples.}
  \label{tbl:accu_felicity}
  \SetHorizontalCoffin \AccuFelTableR {
    \subfloat[Rejection Experiment]{
      \begin{small}
        \hspace*{-3ex}
        \begin{tabular*}{\hsize}{@{\extracolsep{\fill}}lcccccccccccccccccccccc}
          \toprule
          & \multicolumn{2}{c}{P01} & \multicolumn{2}{c}{P04} & \multicolumn{2}{c}{P05} & \multicolumn{2}{c}{P06} & \multicolumn{2}{c}{P07} & \multicolumn{2}{c}{P08} & 
            \multicolumn{2}{c}{P09} & \multicolumn{2}{c}{P10} & \multicolumn{2}{c}{P11} & \multicolumn{2}{c}{P12} &  \multicolumn{2}{c}{\emph{total}}\\
          \textbf{type} & cnt & \%fel & cnt & \%fel & cnt & \%fel & cnt & \%fel & cnt & \%fel & cnt & \%fel & cnt & \%fel & cnt & \%fel & cnt & \%fel & cnt & \%fel & \emph{cnt} & \emph{\%fel}\\
          \cmidrule(lr){2-3}\cmidrule(lr){4-5}\cmidrule(lr){6-7}\cmidrule(lr){8-9}\cmidrule(lr){10-11}\cmidrule(lr){12-13}\cmidrule(lr){14-15}
          \cmidrule(lr){16-17}\cmidrule(lr){18-19}\cmidrule(lr){20-21}\cmidrule(lr){22-23}
          TD & 0 & n/a & 5 & 100 & 0 & n/a & 0 & n/a & 0 & n/a & 0 & n/a & 0 & n/a & 0 & n/a & 1 & 0 & 13 & 53.85 & \emph{19} & \emph{63.16} \\
          E & 0 & n/a & 0 & n/a & 0 & n/a & 0 & n/a & 1 & 100 & 1 & 100 & 0 & n/a & 0 & n/a & 0 & n/a & 0 & n/a & \emph{2} & \emph{100} \\
          A & 0 & n/a & 4 & 100 & 8 & 100 & 0 & n/a & 7 & 85.71 & 1 & 100 & 4 & 100 & 0 & n/a & 2 & 100 & 1 & 100 & \emph{27} & \emph{96.3} \\
          R & 0 & n/a & 3 & 100 & 17 & 29.41 & 0 & n/a & 4 & 75 & 11 & 36.36 & 8 & 100 & 0 & n/a & 16 & 68.75 & 16 & 62.5 & \emph{75} & \emph{58.67} \\
          MD & 0 & n/a & 0 & n/a & 30 & 40 & 0 & n/a & 15 & 73.33 & 13 & 84.62 & 7 & 85.71 & 0 & n/a & 10 & 40 & 6 & 33.33 & \emph{81} & \emph{56.79} \\
          I & 0 & n/a & 0 & n/a & 0 & n/a & 0 & n/a & 0 & n/a & 0 & n/a & 20 & 90 & 0 & n/a & 0 & n/a & 0 & n/a & \emph{20} & \emph{90} \\
          \midrule
          all & 0 & n/a & 12 & 100 & 55 & 45.45 & 0 & n/a & 27 & 77.78 & 26 & 65.38 & 39 & 92.31 & 0 & n/a & 29 & 58.62 & 36 & 55.56 & \emph{224} & \emph{66.07} \\
          \bottomrule
        \end{tabular*}
      \end{small}
    }
  }
  \JoinCoffins\AccuFelTable[vc,hc]\AccuFelTableR[vc,hc]
  \SetHorizontalCoffin \AccuFelTableP {
    \vspace*{1ex}
    \subfloat[Prohibition Experiment]{
      \begin{small}
        \hspace*{-3ex}
        \begin{tabular*}{\hsize}{@{\extracolsep{\fill}}lcccccccccccccccccccccc}
          \toprule
          & \multicolumn{2}{c}{P13} & \multicolumn{2}{c}{P14} & \multicolumn{2}{c}{P15} & \multicolumn{2}{c}{P16} & \multicolumn{2}{c}{P17} & \multicolumn{2}{c}{P18} & \multicolumn{2}{c}{P19} & 
            \multicolumn{2}{c}{P20} & \multicolumn{2}{c}{P21} & \multicolumn{2}{c}{P22} & \multicolumn{2}{c}{\emph{total}}\\ 
          \textbf{type} & cnt & \%fel & cnt & \%fel & cnt & \%fel & cnt & \%fel & cnt & \%fel & cnt & \%fel & cnt & \%fel & cnt & \%fel & cnt & \%fel & cnt & \%fel & \emph{cnt} & \emph{\%fel}\\
          \cmidrule(lr){2-3}\cmidrule(lr){4-5}\cmidrule(lr){6-7}\cmidrule(lr){8-9}\cmidrule(lr){10-11}\cmidrule(lr){12-13}\cmidrule(lr){14-15}
          \cmidrule(lr){16-17}\cmidrule(lr){18-19}\cmidrule(lr){20-21}\cmidrule(lr){22-23}
          TD & 5 & 0 & 10 & 0 & 3 & 0 & 1 & 0 & 8 & 12.5 & 0 & n/a & 0 & n/a & 2 & 0 & 10 & 30 & 3 & 33.33 & \emph{42} & \emph{11.9}\\
          E & 0 & n/a & 0 & n/a & 0 & n/a & 0 & n/a & 0 & n/a & 0 & n/a & 0 & n/a & 1 & 0 & 0 & n/a & 2 & 100 & \emph{3} &\emph{66.66} \\
          A & 0 & n/a & 0 & n/a & 12 & 83.33 & 4 & 100 & 0 & n/a & 0 & n/a & 0 & n/a & 6 & 83.33 & 8 & 75 & 3 & 100 & \emph{33} & \emph{84.84}\\
          R & 0 & n/a & 3 & 33.33 & 13 & 46.15 & 1 & 0 & 15 & 73.33 & 2 & 100 & 0 & n/a & 4 & 75 & 10 & 70 & 1 & 0 & \emph{49} & \emph{61.22}\\
          MD & 0 & n/a & 1 & 0 & 38 & 39.47 & 7 & 14.29 & 28 & 42.86 & 1 & 100 & 0 & n/a & 12 & 25 & 12 & 58.33 & 0 & n/a & \emph{99} & \emph{39.39}\\
          I & 0 & n/a & 0 & n/a & 0 & n/a & 4 & 25 & 0 & n/a & 0 & n/a & 0 & n/a & 0 & n/a & 0 & n/a & 0 & n/a & \emph{4} & \emph{25} \\
          PD & 0 & n/a & 1 & 100 & 0 & n/a & 0 & n/a & 1 & 0 & 0 & n/a & 4 & 100 & 4 & 25 & 0 & n/a & 0 & n/a & \emph{10} & \emph{60}\\
          SP & 0 & n/a & 0 & n/a & 2 & 0 & 1 & 0 & 0 & n/a & 0 & n/a & 16 & 100 & 7 & 42.86 & 16 & 68.75 & 0 & n/a & \emph{42} & \emph{71.43} \\
          \midrule
          all & 5 & 0 & 15 & 13.33 & 68 & 45.59 & 18 & 33.33 & 52 & 46.15 & 3 & 100 & 20 & 100 & 36 & 41.67 & 56 & 60.71 & 9 & 66.67 & \emph{282} & \emph{50} \\
          \bottomrule
        \end{tabular*}
      \end{small}
    }
  }
  \JoinCoffins\AccuFelTable[\AccuFelTableR-b,\AccuFelTableR-l]\AccuFelTableP[t,l]
  \TypesetCoffin\AccuFelTable
\end{table*}
\NewCoffin \AccuFelTableLast
\NewCoffin \AccuFelTableRLast
\NewCoffin \AccuFelTablePLast
\begin{table*}[h]
  \setlength{\tabcolsep}{0.35ex}
  \caption{\textbf{Accumulated frequencies for sessions 4+5 for robot negation types and their felicity}. Displayed are the accumulated frequencies
    of the various negation types the robot engaged in during the last two sessions and their felicities. The following abbreviations are used for 
    the negation types: \emph{TD}: truth-func. denial, \emph{MD}: mot.-dep. denial, \emph{A}: neg. agreement, \emph{R}: rejection, \emph{I}: neg. imperative,
    \emph{E}: mot. dep. exclamation, \emph{SP}: self-prohibition, \emph{PD}: perspective-dependent denial. See SI coding scheme for a description of each type
    including examples.}
  \label{tbl:accu_felicity_s45}
  \SetHorizontalCoffin \AccuFelTableRLast {
    \subfloat[Rejection Experiment]{
      \begin{small}
        \hspace*{-3ex}
        \begin{tabular*}{\hsize}{@{\extracolsep{\fill}}lcccccccccccccccccccccc}
          \toprule
          & \multicolumn{2}{c}{P01} & \multicolumn{2}{c}{P04} & \multicolumn{2}{c}{P05} & \multicolumn{2}{c}{P06} & \multicolumn{2}{c}{P07} & \multicolumn{2}{c}{P08} & 
            \multicolumn{2}{c}{P09} & \multicolumn{2}{c}{P10} & \multicolumn{2}{c}{P11} & \multicolumn{2}{c}{P12} & \multicolumn{2}{c}{\emph{total}}\\
          \textbf{type} & cnt & \%fel & cnt & \%fel & cnt & \%fel & cnt & \%fel & cnt & \%fel & cnt & \%fel & cnt & \%fel & cnt & \%fel & cnt & \%fel & cnt & \%fel & cnt & \%fel\\
          \cmidrule(lr){2-3}\cmidrule(lr){4-5}\cmidrule(lr){6-7}\cmidrule(lr){8-9}\cmidrule(lr){10-11}\cmidrule(lr){12-13}\cmidrule(lr){14-15}
          \cmidrule(lr){16-17}\cmidrule(lr){18-19}\cmidrule(lr){20-21}\cmidrule(lr){22-23}
          E & 0 & n/a & 0 & n/a & 0 & n/a & 0 & n/a & 0 & n/a & 1 & 100 & 0 & n/a & 0 & n/a & 0 & n/a & 0 & n/a & \emph{1} & \emph{0} \\
          TD & 0 & n/a & 1 & 100 & 0 & n/a & 0 & n/a & 0 & n/a & 0 & n/a & 0 & n/a & 0 & n/a & 1 & 0 & 4 & 100 & \emph{6} & \emph{83.33} \\
          A & 0 & n/a & 4 & 100 & 1 & 100 & 0 & n/a & 4 & 100 & 1 & 100 & 0 & n/a & 0 & n/a & 0 & n/a & 1 & 100 & \emph{11} & \emph{100} \\
          R & 0 & n/a & 2 & 100 & 6 & 50 & 0 & n/a & 3 & 100 & 3 & 66.67 & 1 & 100 & 0 & n/a & 10 & 50 & 3 & 100 & \emph{28} & \emph{67.86} \\
          MD & 0 & n/a & 0 & n/a & 7 & 42.86 & 0 & n/a & 9 & 66.67 & 5 & 80 & 1 & 100 & 0 & n/a & 8 & 25 & 3 & 0 & \emph{33} & \emph{48.48} \\
          I & 0 & n/a & 0 & n/a & 0 & n/a & 0 & n/a & 0 & n/a & 0 & n/a & 5 & 100 & 0 & n/a & 0 & n/a & 0 & n/a & \emph{5} & \emph{100} \\
          \midrule
          all & 0 & n/a & 7 & 100 & 14 & 50 & 0 & n/a & 16 & 81.25 & 10 & 80 & 7 & 100 & 0 & n/a & 19 & 36.84 & 11 & 72.73 & \emph{84} & \emph{67.86} \\
          \bottomrule
        \end{tabular*}
      \end{small}
    }
  }
  \JoinCoffins\AccuFelTableLast[vc,hc]\AccuFelTableRLast[vc,hc]
  \SetHorizontalCoffin \AccuFelTablePLast {
    \vspace*{1ex}
    \subfloat[Prohibition Experiment]{
      \begin{small}
        \hspace*{-3ex}
        \begin{tabular*}{\hsize}{@{\extracolsep{\fill}}lcccccccccccccccccccccc}
          \toprule
          & \multicolumn{2}{c}{P13} & \multicolumn{2}{c}{P14} & \multicolumn{2}{c}{P15} & \multicolumn{2}{c}{P16} & \multicolumn{2}{c}{P17} & \multicolumn{2}{c}{P18} & 
            \multicolumn{2}{c}{P19} & \multicolumn{2}{c}{P20} & \multicolumn{2}{c}{P21} & \multicolumn{2}{c}{P22} & \multicolumn{2}{c}{\emph{total}}\\
          \textbf{type} & cnt & \%fel & cnt & \%fel & cnt & \%fel & cnt & \%fel & cnt & \%fel & cnt & \%fel & cnt & \%fel & cnt & \%fel & cnt & \%fel & cnt & \%fel & cnt & \%fel \\
          \cmidrule(lr){2-3}\cmidrule(lr){4-5}\cmidrule(lr){6-7}\cmidrule(lr){8-9}\cmidrule(lr){10-11}\cmidrule(lr){12-13}\cmidrule(lr){14-15}
          \cmidrule(lr){16-17}\cmidrule(lr){18-19}\cmidrule(lr){20-21}\cmidrule(lr){22-23}
          TD & 5 & 0 & 10 & 0 & 3 & 0 & 1 & 0 & 6 & 0 & 0 & n/a & 0 & n/a & 1 & 0 & 0 & n/a & 2 & 0 & \emph{28} & \emph{0} \\
          MD & 0 & 0 & 1 & 0 & 17 & 29.41 & 6 & 0 & 25 & 36 & 0 & n/a & 0 & n/a & 8 & 12.5 & 9 & 77.78 & 0 & n/a & \emph{66} & \emph{33.33} \\
          PD & 0 & n/a & 1 & 100 & 0 & n/a & 0 & n/a & 1 & 0 & 0 & n/a & 0 & n/a & 3 & 33.33 & 0 & n/a & 0 & n/a & \emph{5} & \emph{40} \\
          R & 0 & n/a & 3 & 33.33 & 3 & 33.33 & 1 & 0 & 5 & 20 & 0 & n/a & 0 & n/a & 1 & 0 & 4 & 100 & 1 & 0 & \emph{18} & \emph{38.89} \\
          A & 0 & n/a & 0 & n/a & 1 & 100 & 3 & 100 & 0 & n/a & 0 & n/a & 0 & n/a & 0 & n/a & 0 & n/a & 0 & n/a & \emph{4} & \emph{100} \\
          E & 0 & n/a & 0 & n/a & 0 & n/a & 0 & n/a & 0 & n/a & 0 & n/a & 0 & n/a & 1 & 0 & 0 & n/a & 1 & 100 & \emph{2} & \emph{50} \\
          SP & 0 & n/a & 0 & n/a & 0 & n/a & 1 & 0 & 0 & n/a & 0 & n/a & 5 & 100 & 2 & 0 & 5 & 0 & 0 & n/a & \emph{13} & \emph{38.46} \\
          \midrule
          all & 5 & 0 & 15 & 13.33 & 24 & 29.17 & 12 & 25 & 37 & 27.03 & 0 & n/a & 5 & 100 & 16 & 12.5 & 18 & 61.11 & 4 & 25 & \emph{136} & \emph{30.15} \\
          \bottomrule
        \end{tabular*}
      \end{small}
    }
  }
  \JoinCoffins\AccuFelTableLast[\AccuFelTableRLast-b,\AccuFelTableRLast-l]\AccuFelTablePLast[t,l]
  \TypesetCoffin\AccuFelTableLast
\end{table*}
\begin{table}[h]
  \caption{\textbf{Statistical comparison of felicity rates between both experiments}: Given are the mean and standard deviation for the felicities of the robot's
    production of negation during the sessions 4 and 5 under two criteria: \textbf{Crit. 1}: data basis = felicity values of all participants but \emph{P01},
    \emph{P06}, \emph{P10}, and \emph{P18}. \textbf{Crit. 2}: data basis as in crit. 1 plus additional exclusion of \emph{P04}, \emph{P09}, \emph{P13}, \emph{P19},
    and \emph{P22}. \ensuremath{\star: p < 0.01}, \ensuremath{\dagger: p < 0.02}.}
  \label{tbl:felicity_robot_stat}
  \setlength{\tabcolsep}{2.15ex}
  \begin{tabular}{cccc}
    \toprule
    criterion & experiment & mean \% felicity (std) & T \\
    \midrule
    \multirow{2}{*}{crit. 1} & R  & 74.4 (23.8) &  \multirow{2}{*}{$3.0^\star$}\\
              & P  & 32.57 (30.31) & \\
    \cmidrule{2-4}
    \multirow{2}{*}{crit. 2} & R  & 64.16 (19.8) & \multirow{2}{*}{$3.2^\dagger$} \\
              & P  & 28.02 (17.08) & \\
    \bottomrule
  \end{tabular}
\end{table}

\subsubsection{Temporal Relationships between Prohibitive Linguistic and Corporal Action}
\label{sec:temporal_rels}
The following analysis correlates participants' linguistic prohibition and disallowances, in the following called $prohibition^+$ (see main text and coding scheme
(ref\_to\_coding\_scheme) for the distinction of the two types), with participants' bodily application of restraint, and the robot's motivational state. Whether participants
restrained the humanoid's arm or not was determined via recorded measurements of its sensor detecting external force and which forms the basis of
its compliant behaviour. In order to perform this correlation we fused three different data sources: the pragmatic codes as determined within the pragmatic
analysis, the timed transcriptions of participants' speech, and the timed recording of the said force sensor from the robot's log files which also
contains a record of its motivational state. Figure \ref{fig:push_mot_align} gives a visual depiction of such an alignment.
\begin{figure*}
  \includegraphics[scale=0.535]{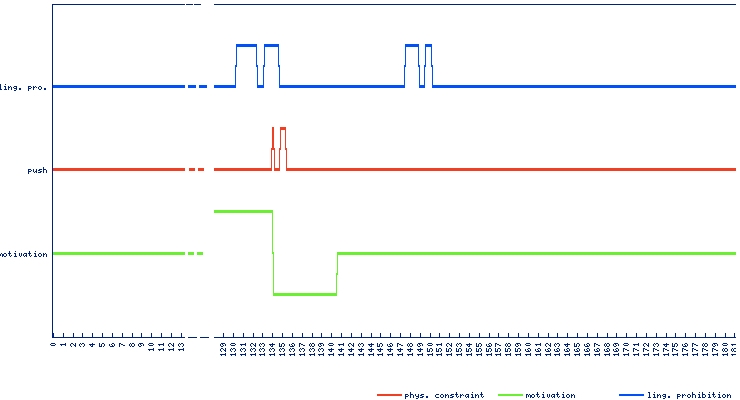}
  \caption{\textbf{Excerpt of reconstructed temporal profile of human-robot interaction}: the given excerpt, taken from the reconstructed profile of P14's
    3rd session, displays  the temporal relation between \emph{prohibitive} utterances and utterances of \emph{disallowance} (top blue line), the robot's 
    sensing of pressure being applied to its arm (middle red line), and the robot's internal motivation (bottom green line).}
  \label{fig:push_mot_align}
\end{figure*}
\noindent Upon having performed said temporal reconstruction for all participants of the prohibition experiment we determined the prevailing types of temporal
relationships between $prohibition^+$ and participants' application of corporal restraint (in the following just called \emph{push}). These are depicted
in figure \ref{fig:push_rels}. Additionally we observed two relations which can be decomposed into the basic relations depicted in said figure:
$$\text{between\_pushes} \Leftrightarrow \text{after\_push}(i) \wedge \text{before\_push}(j)\text{ with }j > i$$
$$\text{during\_several\_pushes} \Leftrightarrow \text{overlap\_after}(i) \wedge \text{overlap\_before}(j) \text{ with }j > i.$$
Based on observations of the experimental video recordings we imposed a time constraint of 4 seconds as maximum gap size between
linguistic and corporal action for any two instances of \emph{prohibition}$^+$ and \emph{push} to be considered in any of the stated temporal relations.
Table \ref{tbl:prohibition_rel} shows the resulting counts for each participant and session in which the prohibitive task was active. As can be seen from
there, participants often did not restrain the robot's arm movement or restrained it only after uttering a \emph{prohibition}$^+$ (``before push'', see also main
text). This is opposed to how we imagined them to act and leads to a violation of the simultaneity constraint in our learning architecture.

\subsubsection{How temporal relations of prohibitive action impact on the acquisition of negation}\hfill \break
In the following we give some examples in order for the reader to better understand why our participants' unexpected behavior, that is participants
either not restraining the robot's arm as advised or them restraining the robot's arm after uttering prohibitions, is detrimental for the acquisition of
negation.
Let us assume \emph{no} as default negative word and let us further assume that \emph{no} is the salient word of the $prohibitions^+$-type utterances
in question.

Similar to other mechanisms, which establish an association between object labels and perceptual features in other symbol-grounding architectures we may regard our
memory-based learner as a roughly associative learner. Associations between labels and other sensorimotor-motivational (\emph{smm}) data come to be by virtue of there
being a majority of exemplars in the memory where such an association is established. For our purposes this means that, all other things, i.e. sensorimotor-states,
being equal, such an association is established as soon as the majority of \emph{no}'s are attached to sensorimotor-motivational data with a negative motivational
entry. This means that any temporal relation leading to a \emph{no} with negative motivational value attached being added to the lexicon is \textbf{beneficial}
for our learning target. By contrast any temporal relation which leads to a \emph{no} with positive motivational value being added is \textbf{detrimental} to
this purpose.

Our version of symbol grounding is implemented such that the salient word is associated with all variants of the sensorimotor-motivational (\emph{smm}) 
vector that co-occurred during the time when the utterance was produced. During short time frames of a few seconds, the typical length of an utterance, 
in most cases nothing in this vector changes: The robot's behaviour does not change, the presented object stays the same, and, importantly, 
the object is recognized by the object detection to be the same. If this is the case while participants produce an utterance, the outcome will be one 
additional exemplar or lexical entry that is added to the robot's embodied lexicon. Yet, if one \emph{smm} change occurs during this production, two 
lexical entries for the same word will be created, one for each variant of the \emph{smm} vector.
Changes  in the \emph{smm} data are caused through changes of the robot's behaviour, which for their part are caused by either timeouts in the body behavior
system or changes in the object recognition. Also changes in the object id itself are forms of sensorimotor changes and so is the change of the robot's
motivational state. We will in the following make the simplifying assumption that the robot's object recognition works perfectly.

The humanoids behaviour was implemented such that it would only grasp for objects that it likes, i.e. objects that cause its motivational state to be 
positive. Under perfect object recognition the robot's motivational state will be positive before the participant restrains its arm movement (push action).
Restraints of the robot's agency lead immediately to Deechee becoming `grumpy', i.e. a negative motivational state.
\begin{figure*}[h]
  \hspace*{-2ex}
  \includegraphics[scale=0.55]{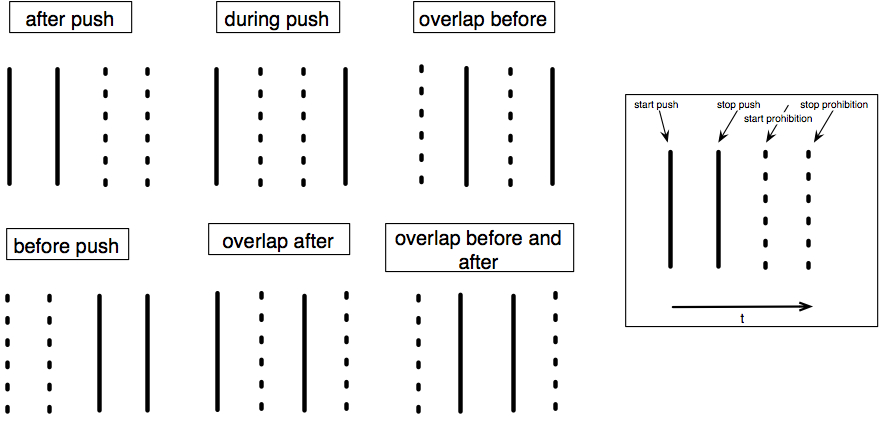}
  \caption{\textbf{Basic temporal relations between corporal constraints and prohibition$^+$}: The depicted temporal relations between
    \emph{prohibition}$^+$ and corporal constraints (``push'') were observed within the prohibition scenario. Additionally
    two complex relations were observed which can be decomposed in the depicted ones (see text).}
  \label{fig:push_rels}
\end{figure*}
For \emph{no push} relations the following holds: \emph{No} is uttered while the robot is and continues to be in a `positive mood', for its agency 
is not impeded. Instead of restraining the robot's arm, as they were taught to do, participants often just held the object out of the robot's reach, 
which has no impact on its motivational state. Such interaction will lead to at least one exemplar of \emph{no} in the robot's lexicon which is 
associated to a \emph{smm} vector which has a positive motivational entry. This is \textbf{detrimental} to the learning outcome.

In contrast, Deechee will already be `in a negative mood' in case of participants starting to restrain its arm before uttering a $prohibition^+$
(\emph{during push}). In this case one embodied word will enter the lexicon: \emph{no} associated with a \emph{smm} vector containing a negative 
motivational value. This is how we imagined the interaction to unfold motivated by assumptions of simultaneity in ostensive theories of meaning. 
This is \textbf{beneficial}.

In case of a participant starting to produce an utterance followed by him or her constraining the robot's arm movement during that production (\emph{overlap before}), 
two lexical entries will be created: a \emph{no}, associated with a \emph{smm} vector with a positive motivational entry, and additionally a \emph{no}, which is 
associated with an otherwise identical \emph{smm} vector but with a negative motivational entry. This is \textbf{in-between}.

If the onset of utterance production happens during a \emph{push} but extends to after the end of the push (\emph{overlap after}), the result will be one
additional \emph{no} in Deechee's lexicon associated to a \emph{smm} vector with negative motivational entry \emph{as long as the utterance is not overly long}.
The robot's motivational system is implemented such, that its motivational state has a certain time lag. The only exception to this rule are restrictions of
Deechee's freedom of movement which will make it grumpy immediately. Therefore the presence of said \emph{overlap after} relation between the mentioned actions
is most probably \textbf{beneficial}.
\begin{table*}[h]
  \caption{\textbf{Counts of temporal relationships between physical constraints and prohibitive utterances}. Given are the counts of observed
    temporal relationships. Both \emph{prohibitions} as well as \emph{disallowances} were taken into consideration in the given count. Counts 
    are given for all participants and sessions in the prohibition scenario in which participants were told to physically restrain the robot 
    in case of it approaching a forbidden object. Furthermore a total count per participants is given in the last section of the table. A missing
    relationship type in a session indicates that all counts were 0. Temporal relationships of the listed types set in bold are very likely to be
    detrimental for an association of the salient word with negative affect in our architecture. Relationships of a type set in italic are less likely to be
    detrimental for said association depending on the length of the gap between push(es) and utterance and the hypothesized duration of the 
    motivational state triggered by physical restraint.}
  \label{tbl:prohibition_rel}
  \begin{tabular*}{\hsize}{@{\extracolsep{\fill}}llllllllllll}
    \toprule
    &  & P13 & P14 & P15 & P16 & P17 & P18 & P19 & P20 & P21 & P22\\
    \midrule
    \multirow{8}{*}{s1} & \textbf{no\_push} & 0 & 0 & 15 & 14 & 1 & 4 & 6 & 0 & 1 & 4\\
    & \textbf{before\_push} & 1 & 0 & 0 & 0 & 0 & 1 & 3 & 2 & 5 & 0\\
    & \textbf{overlap\_before} & 1 & 0 & 0 & 0 & 1 & 0 & 0 & 0 & 1 & 0\\
    & \textbf{overlap\_before\_and\_after} & 1 & 0 & 0 & 0 & 2 & 0 & 0 & 0 & 0 & 0\\
    & \textsl{after\_push} & 0 & 0 & 1 & 0 & 1 & 0 & 1 & 2 & 3 & 0\\
    & overlap\_after & 1 & 0 & 1 & 0 & 1 & 1 & 0 & 0 & 0 & 0\\
    & \textsl{between\_pushes} & 0 & 0 & 0 & 0 & 1 & 0 & 0 & 2 & 1 & 0\\
    & during\_push & 4 & 0 & 1 & 0 & 2 & 0 & 1 & 4 & 3 & 0\\
    \midrule
    \multirow{8}{*}{s2} & \textbf{no\_push} & 0 & 3 & 5 & 10 & 0 & 0 & 0 & 3 & 30 & 2\\
    & \textbf{before\_push} & 3 & 4 & 1 & 0 & 0 & 2 & 0 & 3 & 0 & 0\\
    & \textbf{overlap\_before} & 2 & 0 & 1 & 0 & 1 & 1 & 0 & 1 & 0 & 0\\
    & \textbf{overlap\_before\_and\_after} & 0 & 0 & 1 & 0 & 0 & 0 & 0 & 1 & 0 & 0\\
    & \textsl{after\_push} & 0 & 1 & 0 & 0 & 0 & 0 & 0 & 1 & 0 & 0\\
    & overlap\_after & 1 & 0 & 0 & 0 & 0 & 0 & 0 & 0 & 0 & 0\\
    & \textsl{between\_pushes} & 0 & 0 & 0 & 0 & 0 & 0 & 0 & 1 & 0 & 0\\
    & \textsl{during\_several\_pushes} & 0 & 0 & 1 & 0 & 0 & 0 & 0 & 2 & 0 & 0\\
    & during\_push & 5 & 0 & 2 & 0 & 4 & 2 & 0 & 2 & 0 & 0\\
    \midrule
    \multirow{8}{*}{s3} & \textbf{no\_push} & 0 & 2 & 1 & 3 & 0 & 1 & 0 & 3 & 30 & 5\\
    & \textbf{before\_push} & 0 & 3 & 2 & 0 & 0 & 1 & 3 & 2 & 0 & 0\\
    & \textbf{overlap\_before} & 1 & 1 & 6 & 3 & 3 & 0 & 1 & 2 & 0 & 0\\
    & \textbf{overlap\_before\_and\_after} & 0 & 1 & 0 & 0 & 0 & 1 & 0 & 1 & 0 & 0\\
    & \textsl{after\_push} & 0 & 1 & 0 & 1 & 1 & 1 & 0 & 0 & 0 & 0\\
    & overlap\_after & 0 & 0 & 2 & 1 & 1 & 0 & 0 & 1 & 0 & 0\\
    & \textsl{between\_pushes} & 0 & 0 & 5 & 1 & 4 & 0 & 0 & 0 & 0 & 0\\
    & \textsl{during\_several\_pushes} & 0 & 0 & 0 & 0 & 0 & 0 & 0 & 1 & 0 & 0\\
    & during\_push & 2 & 0 & 8 & 1 & 8 & 0 & 2 & 5 & 0 & 0\\
    \midrule
    \multirow{8}{*}{total} & \textbf{no\_push} & 0 & 5 & 21 & 27 & 1 & 5 & 6 & 6 & 61 & 11\\
    & \textbf{before\_push} & 4 & 7 & 3 & 0 & 0 & 4 & 6 & 7 & 5 & 0\\
    & \textbf{overlap\_before} & 4 & 1 & 7 & 3 & 5 & 1 & 1 & 3 & 1 & 0\\
    & \textbf{overlap\_before\_and\_after} & 1 & 1 & 1 & 0 & 2 & 1 & 0 & 2 & 0 & 0\\
    & \textsl{after\_push} & 0 & 2 & 1 & 1 & 2 & 1 & 1 & 3 & 3 & 0\\
    & overlap\_after & 2 & 0 & 3 & 1 & 2 & 1 & 0 & 1 & 0 & 0\\
    & \textsl{between\_pushes} & 0 & 0 & 5 & 1 & 5 & 0 & 0 & 3 & 1 & 0\\
    & \textsl{during\_several\_pushes} & 0 & 0 & 1 & 0 & 0 & 0 & 0 & 3 & 0 & 0\\
    & during\_push & 11 & 0 & 11 & 1 & 14 & 2 & 3 & 11 & 3 & 0\\
    \bottomrule
  \end{tabular*}
\end{table*}
\noindent Table \ref{tbl:prohibition_mot} shows the counts of the various temporal relations between corporal restraint and prohibition$^+$ as well as the motivational
states in which the robot was in when the respective form of prohibition was performed.
Table \ref{tbl:neg_int_int_mot1} shows the motivational states of the robot during the other highly frequent negation types \emph{negative intent interpretations}
and \emph{negative motivational questions} in the rejection experiment. Table \ref{tbl:neg_int_int_mot2} shows the same for our participants from the prohibition
experiment.
\begin{table*}[h]
  \caption{\textbf{Motivational states during utterances of prohibition and disallowance}. 
    Given are the counts of the robot's motivational states for each temporal relationship between prohibition/disallowance and physical restraint.
    The counts are listed per participant within the prohibition experiment (see table \ref{tbl:prohibition_rel} for the frequencies of these relationships). 
    The counts are accumulated over the first three sessions, i.e. the sessions in which physical restraint could possibly occur.
    Note that one occurrence of such a temporal relationship can yield more than 1 to the count as the robot's motivational state can change while the respective
    utterance is being produced. The entries for \emph{P13} for \emph{overlap\_before\_and\_after}
    are so large due to a glitch in the motivational and/or behavioral system.
    Symbols used: \emph{-}: negative motivation, \emph{+}: positive motivation, \emph{O}: neutral motivation}
  \label{tbl:prohibition_mot}
  \begin{tabular*}{\hsize}{@{\extracolsep{\fill}}llllllllllllllll}
    \toprule
    &  & P13 &  &  & P14 &  &  & P15 &  &  & P16 &  &  & P17\\
    \cmidrule(lr){2-4}\cmidrule(lr){5-7}\cmidrule(lr){8-10}\cmidrule(lr){11-13}\cmidrule(lr){14-16}
    & - & O & + & - & O & + & - & O & + & - & O & + & - & O & +\\
    \cmidrule(lr){2-4}\cmidrule(lr){5-7}\cmidrule(lr){8-10}\cmidrule(lr){11-13}\cmidrule(lr){14-16}
    no\_push & 0 & 0 & 0 & 0 & 0 & 5 & 4 & 4 & 17 & 0 & 1 & 14 & 0 & 0 & 1\\
    before\_push & 0 & 1 & 4 & 0 & 0 & 7 & 0 & 0 & 3 & 0 & 0 & 0 & 0 & 0 & 0\\
    overlap\_before & 4 & 0 & 4 & 1 & 0 & 1 & 6 & 0 & 6 & 3 & 0 & 3 & 5 & 0 & 3\\
    overlap\_before\_and\_after & 11 & 12 & 0 & 1 & 0 & 1 & 1 & 0 & 1 & 0 & 0 & 0 & 2 & 0 & 2\\
    after\_push & 0 & 0 & 0 & 2 & 0 & 1 & 1 & 0 & 0 & 0 & 0 & 1 & 2 & 0 & 0\\
    overlap\_after & 2 & 0 & 0 & 0 & 0 & 0 & 3 & 1 & 0 & 1 & 0 & 0 & 2 & 0 & 0\\
    between\_pushes & 0 & 0 & 0 & 0 & 0 & 0 & 2 & 0 & 3 & 1 & 1 & 1 & 4 & 0 & 2\\
    during\_several\_pushes & 0 & 0 & 0 & 0 & 0 & 0 & 1 & 0 & 0 & 0 & 0 & 0 & 0 & 0 & 0\\
    during\_push & 11 & 0 & 0 & 0 & 0 & 0 & 11 & 0 & 0 & 1 & 0 & 0 & 14 & 0 & 0\\
  \end{tabular*}
  \begin{tabular*}{\hsize}{@{\extracolsep{\fill}}llllllllllllllll}
    \toprule
    & & P18 &  &  & P19 &  &  & P20 &  &  & P21 &  &  & P22\\
    \cmidrule(lr){2-4}\cmidrule(lr){5-7}\cmidrule(lr){8-10}\cmidrule(lr){11-13}\cmidrule(lr){14-16}
    & - & O & + & - & O & + & - & O & + & - & O & + & - & O & +\\
    \cmidrule(lr){2-4}\cmidrule(lr){5-7}\cmidrule(lr){8-10}\cmidrule(lr){11-13}\cmidrule(lr){14-16}
    no\_push & 0 & 0 & 5 & 0 & 2 & 4 & 2 & 4 & 1 & 3 & 14 & 16 & 0 & 1 & 7\\
    before\_push & 0 & 0 & 4 & 0 & 0 & 6 & 0 & 0 & 7 & 0 & 0 & 5 & 0 & 0 & 0\\
    overlap\_before & 1 & 0 & 1 & 1 & 0 & 1 & 3 & 0 & 3 & 1 & 0 & 1 & 0 & 0 & 0\\
    overlap\_before\_and\_after & 1 & 0 & 1 & 0 & 0 & 0 & 2 & 0 & 2 & 0 & 0 & 0 & 0 & 0 & 0\\
    after\_push & 1 & 0 & 0 & 1 & 0 & 1 & 3 & 0 & 0 & 3 & 0 & 0 & 0 & 0 & 0\\
    overlap\_after & 1 & 0 & 0 & 0 & 0 & 0 & 1 & 0 & 0 & 0 & 0 & 0 & 0 & 0 & 0\\
    between\_pushes & 0 & 0 & 0 & 0 & 0 & 0 & 2 & 0 & 1 & 0 & 0 & 1 & 0 & 0 & 0\\
    during\_several\_pushes & 0 & 0 & 0 & 0 & 0 & 0 & 3 & 0 & 0 & 0 & 0 & 0 & 0 & 0 & 0\\
    during\_push & 2 & 0 & 0 & 3 & 0 & 0 & 11 & 0 & 0 & 3 & 0 & 0 & 0 & 0 & 0\\
    \bottomrule
  \end{tabular*}
\end{table*}
\begin{table*}[h]
  \setlength{\tabcolsep}{0.95ex}
  \caption{\textbf{Motivational states during negative intent interpretations and neg. mot. questions within prohibition experiment}. 
    Given are the counts/number of associations of the robot's motivational states per stated utterance type. These frequencies are
    listed per session and accumulated across sessions. Symbols used: \emph{\#}: number of occurrences of the stated utterance type,
    \emph{-}: frequency of negative motivational state, \emph{+}: frequency of positive motivational state, \emph{O}: frequency of
    neutral motivational state.}
  \label{tbl:neg_int_int_mot2}
  \begin{small}
    \begin{tabular*}{\hsize}{@{\extracolsep{\fill}}llllllllllllllllllllll}
      \toprule
      &  & \multicolumn{4}{c}{P13} & \multicolumn{4}{c}{P14} & \multicolumn{4}{c}{P15} & \multicolumn{4}{c}{P16} & \multicolumn{4}{c}{P17}\\
      \cmidrule(lr){3-6}\cmidrule(lr){7-10}\cmidrule(lr){11-14}\cmidrule(lr){15-18}\cmidrule(lr){19-22}
      &  & \# & - & O & + & \# & - & O & + & \# & - & O & + & \# & - & O & + & \# & - & O & +\\
      \cmidrule(lr){3-6}\cmidrule(lr){7-10}\cmidrule(lr){11-14}\cmidrule(lr){15-18}\cmidrule(lr){19-22}
      \multirow{2}{*}{s1} & neg. int. int. & 6 & 5 & 3 & 1 & 0 & 0 & 0 & 0 & 15 & 9 & 7 & 0 & 7 & 2 & 4 & 1 & 4 & 3 & 1 & 0\\
      & neg. mot. question & 4 & 2 & 1 & 1 & 0 & 0 & 0 & 0 & 11 & 5 & 6 & 1 & 3 & 2 & 2 & 0 & 1 & 1 & 0 & 0\\
      \multirow{2}{*}{s2} & neg. int. int. & 6 & 6 & 0 & 0 & 0 & 0 & 0 & 0 & 4 & 4 & 0 & 0 & 4 & 3 & 2 & 0 & 4 & 4 & 1 & 0\\
      & neg. mot. question & 1 & 0 & 0 & 1 & 0 & 0 & 0 & 0 & 8 & 2 & 4 & 3 & 3 & 1 & 2 & 0 & 1 & 1 & 0 & 0\\
      \multirow{2}{*}{s3} & neg. int. int. & 0 & 0 & 0 & 0 & 0 & 0 & 0 & 0 & 4 & 3 & 1 & 0 & 2 & 1 & 1 & 0 & 1 & 1 & 0 & 0\\
      & neg. mot. question & 1 & 1 & 0 & 0 & 0 & 0 & 0 & 0 & 6 & 4 & 3 & 1 & 1 & 0 & 1 & 0 & 1 & 0 & 1 & 0\\
      \multirow{2}{*}{s4} & neg. int. int. & 1 & 1 & 0 & 0 & 0 & 0 & 0 & 0 & 8 & 5 & 4 & 1 & 5 & 2 & 4 & 1 & 1 & 1 & 0 & 0\\
      & neg. mot. question & 4 & 4 & 0 & 0 & 0 & 0 & 0 & 0 & 9 & 4 & 4 & 2 & 4 & 1 & 3 & 1 & 0 & 0 & 0 & 0\\
      \multirow{2}{*}{s5} & neg. int. int. & 2 & 2 & 0 & 0 & 0 & 0 & 0 & 0 & 7 & 5 & 1 & 1 & 13 & 4 & 9 & 0 & 3 & 2 & 1 & 0\\
      & neg. mot. question & 2 & 2 & 0 & 0 & 0 & 0 & 0 & 0 & 18 & 9 & 8 & 2 & 3 & 3 & 0 & 0 & 4 & 2 & 2 & 0\\
      \multirow{2}{*}{total} & neg. int. int. & 15 & 14 & 3 & 1 & 0 & 0 & 0 & 0 & 38 & 26 & 13 & 2 & 31 & 12 & 20 & 2 & 13 & 11 & 3 & 0\\
      & neg. mot. question & 12 & 9 & 1 & 2 & 0 & 0 & 0 & 0 & 52 & 24 & 25 & 9 & 14 & 7 & 8 & 1 & 7 & 4 & 3 & 0\\
    \end{tabular*}
    \setlength{\tabcolsep}{1.1ex}
    \begin{tabular*}{\hsize}{@{\extracolsep{\fill}}llllllllllllllllllllll}
      \toprule
      &  & \multicolumn{4}{c}{P18} & \multicolumn{4}{c}{P19} & \multicolumn{4}{c}{P20} & \multicolumn{4}{c}{P21} & \multicolumn{4}{c}{P22}\\
      \cmidrule(lr){3-6}\cmidrule(lr){7-10}\cmidrule(lr){11-14}\cmidrule(lr){15-18}\cmidrule(lr){19-22}
      &  & \# & - & O & + & \# & - & O & + & \# & - & O & + & \# & - & O & + & \# & - & O & +\\
      \cmidrule(lr){3-6}\cmidrule(lr){7-10}\cmidrule(lr){11-14}\cmidrule(lr){15-18}\cmidrule(lr){19-22}
      \multirow{2}{*}{s1} & neg. int. int. & 15 & 9 & 4 & 3 & 4 & 3 & 1 & 0 & 8 & 4 & 5 & 0 & 1 & 1 & 0 & 0 & 2 & 0 & 1 & 1\\
      & neg. mot. question & 8 & 7 & 2 & 1 & 4 & 4 & 1 & 0 & 2 & 1 & 1 & 1 & 3 & 1 & 3 & 0 & 0 & 0 & 0 & 0\\
      \multirow{2}{*}{s2} & neg. int. int. & 4 & 4 & 0 & 0 & 4 & 4 & 0 & 0 & 1 & 1 & 0 & 0 & 0 & 0 & 0 & 0 & 0 & 0 & 0 & 0\\
      & neg. mot. question & 2 & 1 & 1 & 0 & 3 & 3 & 1 & 0 & 10 & 7 & 4 & 0 & 3 & 3 & 0 & 0 & 0 & 0 & 0 & 0\\
      \multirow{2}{*}{s3} & neg. int. int. & 6 & 4 & 3 & 0 & 5 & 5 & 0 & 0 & 5 & 4 & 2 & 1 & 0 & 0 & 0 & 0 & 0 & 0 & 0 & 0\\
      & neg. mot. question & 2 & 0 & 2 & 0 & 0 & 0 & 0 & 0 & 7 & 5 & 3 & 0 & 3 & 2 & 0 & 1 & 0 & 0 & 0 & 0\\
      \multirow{2}{*}{s4} & neg. int. int. & 2 & 2 & 0 & 0 & 1 & 1 & 0 & 0 & 6 & 4 & 2 & 0 & 0 & 0 & 0 & 0 & 0 & 0 & 0 & 0\\
      & neg. mot. question & 1 & 1 & 0 & 0 & 3 & 3 & 0 & 0 & 6 & 5 & 2 & 1 & 3 & 2 & 1 & 0 & 0 & 0 & 0 & 0\\
      \multirow{2}{*}{s5} & neg. int. int. & 3 & 2 & 0 & 1 & 4 & 4 & 1 & 0 & 2 & 2 & 1 & 0 & 3 & 3 & 0 & 0 & 0 & 0 & 0 & 0\\
      & neg. mot. question & 1 & 1 & 0 & 0 & 2 & 2 & 0 & 0 & 13 & 10 & 3 & 1 & 8 & 7 & 1 & 0 & 0 & 0 & 0 & 0\\
      \multirow{2}{*}{total} & neg. int. int. & 30 & 21 & 7 & 4 & 18 & 17 & 2 & 0 & 22 & 15 & 10 & 1 & 4 & 4 & 0 & 0 & 2 & 0 & 1 & 1\\
      & neg. mot. question & 14 & 10 & 5 & 1 & 12 & 12 & 2 & 0 & 38 & 28 & 13 & 3 & 20 & 15 & 5 & 1 & 0 & 0 & 0 & 0\\
      \bottomrule
    \end{tabular*}
  \end{small}
\end{table*}
\begin{table*}[h]
  \setlength{\tabcolsep}{0.93ex}
  \caption{\textbf{Motivational states during negative intent interpretations and neg. mot. questions within rejection experiment}. 
    Given are the counts/number of associations of the robot's motivational states per stated utterances type. These frequencies are
    listed per session and accumulated across sessions. Note that one utterance of these types can co-occur with more than one motivational state such that
    the sum of motivational states in the table  may be larger than the total number of utterances. Symbols used: \emph{\#}: number of occurrences of the
    stated utterance type, \emph{-}: frequency of negative motivational state, \emph{+}: frequency of positive motivational state, \emph{O}: frequency of
    neutral motivational state.}
  \label{tbl:neg_int_int_mot1}
  \begin{small}
    \begin{tabular*}{\hsize}{@{\extracolsep{\fill}}llllllllllllllllllllll}
      \toprule
      &  & \multicolumn{4}{c}{P01}  & \multicolumn{4}{c}{P04} & \multicolumn{4}{c}{P05} & \multicolumn{4}{c}{P06} & \multicolumn{4}{c}{P07}\\
      \cmidrule(lr){3-6}\cmidrule(lr){7-10}\cmidrule(lr){11-14}\cmidrule(lr){15-18}\cmidrule(lr){19-22}
      &  & \# & - & O & + & \# & - & O & + & \# & - & O & + & \# & - & O & + & \# & - & O & +\\
      \cmidrule(lr){3-6}\cmidrule(lr){7-10}\cmidrule(lr){11-14}\cmidrule(lr){15-18}\cmidrule(lr){19-22}
      \multirow{2}{*}{s1} & neg. int. int. & 1 & 1 & 0 & 0 & 6 & 6 & 0 & 0 & 17 & 11 & 0 & 7 & 5 & 4 & 1 & 1 & 12 & 10 & 2 & 0\\
      & neg. mot. question & 0 & 0 & 0 & 0 & 3 & 3 & 0 & 0 & 6 & 1 & 3 & 2 & 3 & 2 & 0 & 1 & 13 & 5 & 7 & 2\\
      \multirow{2}{*}{s2} & neg. int. int. & 1 & 1 & 0 & 0 & 6 & 6 & 1 & 0 & 7 & 2 & 3 & 2 & 8 & 4 & 3 & 1 & 16 & 8 & 8 & 0\\
      & neg. mot. question & 0 & 0 & 0 & 0 & 7 & 7 & 0 & 0 & 11 & 7 & 4 & 1 & 1 & 1 & 1 & 0 & 7 & 1 & 5 & 2\\
      \multirow{2}{*}{s3} & neg. int. int. & 0 & 0 & 0 & 0 & 3 & 3 & 0 & 0 & 6 & 4 & 3 & 0 & 12 & 12 & 2 & 0 & 7 & 5 & 2 & 0\\
      & neg. mot. question & 0 & 0 & 0 & 0 & 5 & 3 & 2 & 0 & 2 & 0 & 2 & 1 & 2 & 2 & 0 & 0 & 18 & 2 & 9 & 9\\
      \multirow{2}{*}{s4} & neg. int. int. & 0 & 0 & 0 & 0 & 0 & 0 & 0 & 0 & 3 & 2 & 0 & 1 & 13 & 12 & 3 & 1 & 7 & 6 & 0 & 2\\
      & neg. mot. question & 0 & 0 & 0 & 0 & 1 & 0 & 1 & 0 & 5 & 4 & 1 & 1 & 3 & 2 & 1 & 0 & 10 & 0 & 6 & 5\\
      \multirow{2}{*}{s5} & neg. int. int. & 0 & 0 & 0 & 0 & 8 & 7 & 0 & 1 & 3 & 1 & 2 & 0 & 11 & 8 & 5 & 0 & 7 & 6 & 2 & 0\\
      & neg. mot. question & 0 & 0 & 0 & 0 & 8 & 8 & 0 & 0 & 6 & 4 & 3 & 1 & 1 & 0 & 1 & 0 & 24 & 5 & 14 & 7\\
      \multirow{2}{*}{total} & neg. int. int. & 2 & 2 & 0 & 0 & 23 & 22 & 1 & 1 & 36 & 20 & 8 & 10 & 49 & 40 & 14 & 3 & 49 & 35 & 14 & 2\\
      & neg. mot. question & 0 & 0 & 0 & 0 & 24 & 21 & 3 & 0 & 30 & 16 & 13 & 6 & 10 & 7 & 3 & 1 & 72 & 13 & 41 & 25\\
    \end{tabular*}
    \setlength{\tabcolsep}{1.12ex}
    \begin{tabular*}{\hsize}{@{\extracolsep{\fill}}llllllllllllllllllllllll}
      \toprule
      &  & \multicolumn{4}{c}{P08} & \multicolumn{4}{c}{P09} & \multicolumn{4}{c}{P10} & \multicolumn{4}{c}{P11} & \multicolumn{4}{c}{P12}\\
      \cmidrule(lr){3-6}\cmidrule(lr){7-10}\cmidrule(lr){11-14}\cmidrule(lr){15-18}\cmidrule(lr){19-22}
      &  & \# & - & O & + & \# & - & O & + & \# & - & O & + & \# & - & O & + & \# & - & O & + &\\
      \cmidrule(lr){3-6}\cmidrule(lr){7-10}\cmidrule(lr){11-14}\cmidrule(lr){15-18}\cmidrule(lr){19-22}
      \multirow{2}{*}{s1} & neg. int. int. & 6 & 3 & 3 & 0 & 6 & 6 & 1 & 1 & 0 & 0 & 0 & 0 & 9 & 6 & 4 & 1 & 7 & 4 & 2 & 3\\
      & neg. mot. question & 10 & 1 & 9 & 1 & 4 & 3 & 2 & 0 & 0 & 0 & 0 & 0 & 3 & 2 & 0 & 1 & 2 & 0 & 2 & 0\\
      \multirow{2}{*}{s2} & neg. int. int. & 3 & 2 & 1 & 0 & 6 & 4 & 4 & 0 & 0 & 0 & 0 & 0 & 4 & 3 & 1 & 0 & 2 & 1 & 1 & 0\\
      & neg. mot. question & 7 & 1 & 6 & 0 & 5 & 2 & 1 & 2 & 2 & 2 & 0 & 0 & 0 & 0 & 0 & 0 & 2 & 1 & 1 & 0\\
      \multirow{2}{*}{s3} & neg. int. int. & 1 & 1 & 0 & 0 & 6 & 4 & 1 & 1 & 1 & 1 & 1 & 1 & 6 & 6 & 0 & 0 & 0 & 0 & 0 & 0\\
      & neg. mot. question & 10 & 6 & 1 & 3 & 7 & 3 & 6 & 1 & 1 & 1 & 0 & 0 & 2 & 2 & 1 & 0 & 0 & 0 & 0 & 0\\
      \multirow{2}{*}{s4} & neg. int. int. & 4 & 3 & 1 & 0 & 5 & 2 & 5 & 0 & 0 & 0 & 0 & 0 & 0 & 0 & 0 & 0 & 0 & 0 & 0 & 0\\
      & neg. mot. question & 10 & 6 & 4 & 0 & 0 & 0 & 0 & 0 & 3 & 3 & 3 & 2 & 2 & 1 & 0 & 1 & 0 & 0 & 0 & 0\\
      \multirow{2}{*}{s5} & neg. int. int. & 4 & 3 & 2 & 0 & 2 & 2 & 0 & 0 & 0 & 0 & 0 & 0 & 0 & 0 & 0 & 0 & 0 & 0 & 0 & 0\\
      & neg. mot. question & 6 & 3 & 2 & 1 & 4 & 1 & 1 & 3 & 3 & 4 & 2 & 0 & 0 & 0 & 0 & 0 & 0 & 0 & 0 & 0\\
      \multirow{2}{*}{total} & neg. int. int. & 18 & 12 & 7 & 0 & 25 & 18 & 11 & 2 & 1 & 1 & 1 & 1 & 19 & 15 & 5 & 1 & 9 & 5 & 3 & 3\\
      & neg. mot. question & 43 & 17 & 22 & 5 & 20 & 9 & 10 & 6 & 9 & 10 & 5 & 2 & 7 & 5 & 1 & 2 & 4 & 1 & 3 & 0\\
      \bottomrule
    \end{tabular*}
  \end{small}
\end{table*}

\end{screenonly}

\clearpage

\bibliographystyle{ACM-Reference-Format}
\bibliography{arXiv-prohibition}

\end{document}